\documentclass[10pt]{article}
\usepackage[preprint]{tmlr}

\usepackage{microtype}
\usepackage{amsmath,amssymb}
\usepackage{booktabs}
\usepackage{graphicx}
\usepackage{xcolor}
\usepackage{hyperref}
\usepackage{url}
\hypersetup{pdftitle={Minibatch persistency, eight years later: what batch
  reuse costs in steps and joules, and what it saves in data},
  pdfauthor={Matteo Fischetti}}

\newcommand{\Kpers}{4}                  
\newcommand{\nseeds}{8}                 
\newcommand{\nseedsprov}{3}             
\newcommand{\freshbudget}{100}          
\newcommand{\nrunstotal}{144}           
\newcommand{\nrunscoarse}{36}           
\newcommand{\nevals}{16}                
\newcommand{\evalstride}{6.25}          
\newcommand{\alphalevel}{0.05}

\newcommand{\nlayer}{6}
\newcommand{\nhead}{6}
\newcommand{\nembd}{384}
\newcommand{\nparams}{49\,M}            
\newcommand{\seqlen}{1024}
\newcommand{\warmupfrac}{2\%}
\newcommand{\microbatch}{8}
\newcommand{\nlrcoarse}{four}
\newcommand{\nlrfinal}{two}
\newcommand{\lrlow}{$3 \times 10^{-4}$}
\newcommand{\lrhigh}{$8 \times 10^{-3}$}
\newcommand{\ntests}{294}               

\newcommand{\taua}{5.3400}              
\newcommand{\taub}{6.0772}              
\newcommand{\tauc}{7.0681}              
\newcommand{\tauadja}{5.3400}           
\newcommand{\tauadjb}{5.7372}           
\newcommand{\tauadjc}{6.6681}           

\newcommand{\tauEtwoa}{4.380}
\newcommand{\tauEtwob}{4.914}
\newcommand{\tauEtwoc}{6.087}
\newcommand{\Etwofraction}{0.625}

\newcommand{\ga}{2.844}                     
\newcommand{\gc}{0.4097}                    

\newcommand{\pvalue}{0.000566}

\newcommand{\sigmaD}{0.1133}                
\newcommand{\powerrule}{561}                

\newcommand{\gaFINAL}{2.886}                
\newcommand{\gcFINAL}{0.4190}               
\newcommand{\halvingFINAL}{1.443}           
\newcommand{\tstatFINAL}{38.82}
\newcommand{\dfstatFINAL}{7}
\newcommand{\pvalueFINAL}{9.8 \times 10^{-10}}     
\newcommand{\ciupperFINAL}{2.543}           
\newcommand{\aaratiosFINAL}{0.9979 / 0.9823 / 1.0043}
\newcommand{\aasigmaFINAL}{0.0266 / 0.0552 / 0.0594}
\newcommand{\poolendFINAL}{3.893 / 1.579 / 1.514}  
\newcommand{\tauadjnb}{17} \newcommand{\tauadjnc}{20}   
\newcommand{\gaadjFINAL}{2.888}
\newcommand{\gcadjFINAL}{-0.00297}
\newcommand{\pvalueadjFINAL}{1.6 \times 10^{-11}}  
\newcommand{\sigmaDFINAL}{0.1411}
\newcommand{\seedsneededFINAL}{12}          
\newcommand{\underpoweredthreshold}{0.12}   

\newcommand{\bridgen}{24}                   
  
\newcommand{\bridgera}{0.9968} \newcommand{\bridgerb}{1.0235} \newcommand{\bridgerc}{1.0091}
\newcommand{\bridgecia}{[-0.018, +0.011]}
\newcommand{\bridgecib}{[+0.009, +0.038]}
\newcommand{\bridgecic}{[-0.022, +0.040]}
\newcommand{\bridgepairs}{8 / 8 / 7}
\newcommand{\bridgegatemean}{0.05} \newcommand{\bridgegatemargin}{0.10}
\newcommand{\bridgeprimera}{0.9983} \newcommand{\bridgeprimerb}{1.0178} \newcommand{\bridgeprimerc}{0.9726}
\newcommand{\bridgeprimecia}{[-0.009, +0.005]}
\newcommand{\bridgeprimecib}{[+0.001, +0.035]}
\newcommand{\bridgeprimecic}{[-0.081, +0.026]}
\newcommand{\primaryontfour}{138}           
\newcommand{\primaryonupscale}{6}           
\newcommand{\bridgemicro}{32}               

\newcommand{\tenergyperMtok}{2.10}      
\newcommand{\lenergyperMtok}{1.21}      

\newcommand{\ipminodekJ}{210.9}         
\newcommand{\nvmlgpukJ}{54.5}           
\newcommand{\ipmiratio}{3.9}            
\newcommand{\ipmidelta}{349}            
\newcommand{\nvmldelta}{270}            
\newcommand{\hostgap}{80}               
\newcommand{\upscaleover}{11}           

\newcommand{\nrunsnow}{144}             
\newcommand{\nseedsnow}{8}              
\newcommand{\npairs}{8}                 

\newcommand{\Rstepa}{3.886}   \newcommand{\Rstepalo}{3.773}   \newcommand{\Rstepahi}{4.002}
\newcommand{\Rstepb}{1.554}   \newcommand{\Rstepblo}{1.483}   \newcommand{\Rstepbhi}{1.63}
\newcommand{\Rstepc}{1.419}   \newcommand{\Rstepclo}{1.367}   \newcommand{\Rstepchi}{1.474}
\newcommand{\Rfresha}{0.9732} \newcommand{\Rfreshalo}{0.9449} \newcommand{\Rfreshahi}{1.002}
\newcommand{\Rfreshb}{0.3948} \newcommand{\Rfreshblo}{0.3766} \newcommand{\Rfreshbhi}{0.4139}
\newcommand{\Rfreshc}{0.3784} \newcommand{\Rfreshclo}{0.3644} \newcommand{\Rfreshchi}{0.393}
\newcommand{\Rseca}{3.852}    \newcommand{\Rsecalo}{3.739}    \newcommand{\Rsecahi}{3.968}
\newcommand{\Rsecb}{1.501}    \newcommand{\Rsecblo}{1.433}    \newcommand{\Rsecbhi}{1.572}
\newcommand{\Rsecc}{1.402}    \newcommand{\Rsecclo}{1.296}    \newcommand{\Rsecchi}{1.516}
\newcommand{\Rjoulea}{3.866}  \newcommand{\Rjoulealo}{3.76}  \newcommand{\Rjouleahi}{3.975}
\newcommand{\Rjouleb}{1.515}  \newcommand{\Rjouleblo}{1.448}  \newcommand{\Rjoulebhi}{1.584}
\newcommand{\Rjoulec}{1.442}  \newcommand{\Rjouleclo}{1.408}  \newcommand{\Rjoulechi}{1.476}

\newcommand{\aastepa}{0.9979}   \newcommand{\aastepalo}{0.9759}  \newcommand{\aastepahi}{1.02}
\newcommand{\aastepb}{0.9823}  \newcommand{\aastepblo}{0.938}  \newcommand{\aastepbhi}{1.029}
\newcommand{\aastepc}{1.004}   \newcommand{\aastepclo}{0.9557}  \newcommand{\aastepchi}{1.055}
\newcommand{\aafresha}{1.005}  \newcommand{\aafreshalo}{0.9829} \newcommand{\aafreshahi}{1.028}
\newcommand{\aafreshb}{1.007}  \newcommand{\aafreshblo}{0.9618} \newcommand{\aafreshbhi}{1.054}
\newcommand{\aafreshc}{1.099}  \newcommand{\aafreshclo}{1.049}  \newcommand{\aafreshchi}{1.152}  
\newcommand{\aaseca}{0.994}   \newcommand{\aasecalo}{0.9635}   \newcommand{\aasecahi}{1.025}
\newcommand{\aasecb}{0.9605}   \newcommand{\aasecblo}{0.8944}   \newcommand{\aasecbhi}{1.032}
\newcommand{\aasecc}{0.9808}    \newcommand{\aasecclo}{0.8725}   \newcommand{\aasecchi}{1.103}
\newcommand{\aajoulea}{0.9959}  \newcommand{\aajoulealo}{0.9712} \newcommand{\aajouleahi}{1.021}
\newcommand{\aajouleb}{0.9678} \newcommand{\aajouleblo}{0.9076} \newcommand{\aajoulebhi}{1.032}
\newcommand{\aajoulec}{1.009}  \newcommand{\aajouleclo}{0.9515} \newcommand{\aajoulechi}{1.069}

\newcommand{\freshdevtol}{2}
\newcommand{\freshdeva}{0.19}
\newcommand{\freshdevb}{1.6}
\newcommand{\freshdevc}{6.7}
\newcommand{\stepspersalo}{1570}        
\newcommand{\stepspersahi}{1689}        
\newcommand{\stepspersb}{189}           
\newcommand{\stepspersc}{45}            
\newcommand{\aaoffsetc}{1.094}          
\newcommand{\aastepscmin}{28.7}         
\newcommand{\aastepscmax}{34.0}         

\newcommand{\gpupowerlimit}{70}
\newcommand{\nseedstfourc}{7}           
\newcommand{\basesecsa}{429.0}  \newcommand{\basesecsasd}{11.0}  \newcommand{\basekJa}{28.34}  \newcommand{\basekJasd}{0.62}
\newcommand{\basesecsb}{495.4}  \newcommand{\basesecsbsd}{26.1}  \newcommand{\basekJb}{32.58}  \newcommand{\basekJbsd}{1.59}
\newcommand{\basesecsc}{499.4}  \newcommand{\basesecscsd}{29.6}  \newcommand{\basekJc}{32.88}  \newcommand{\basekJcsd}{1.60}
\newcommand{\perssecsa}{1653}   \newcommand{\perssecsasd}{78.8}  \newcommand{\perskJa}{109.6}  \newcommand{\perskJasd}{4.18}
\newcommand{\perssecsb}{742.7}  \newcommand{\perssecsbsd}{20.8}  \newcommand{\perskJb}{49.30}  \newcommand{\perskJbsd}{0.97}
\newcommand{\perssecsc}{721.5}  \newcommand{\perssecscsd}{24.9}  \newcommand{\perskJc}{47.46}  \newcommand{\perskJcsd}{1.40}

\newcommand{\etaone}{8.963 \times 10^{-4}}   
\newcommand{\lrmn}{3}
\newcommand{\lrmnruns}{9}
\newcommand{\lrmtunedstepa}{3.846}  \newcommand{\lrmtunedstepasem}{0.097}
\newcommand{\lrmmatchstepa}{3.816}  \newcommand{\lrmmatchstepasem}{0.107}
\newcommand{\lrmdeltastepa}{-0.031} \newcommand{\lrmdeltastepasem}{0.022}

\newcommand{\lrmdeltafresha}{-0.0077} \newcommand{\lrmdeltafreshasem}{0.0054}
\newcommand{\lrmtunedstepb}{1.561}  \newcommand{\lrmtunedstepbsem}{0.063}   
\newcommand{\lrmtunedstepc}{1.411}  \newcommand{\lrmtunedstepcsem}{0.036}   
\newcommand{\lrmtunedfreshb}{0.3964} \newcommand{\lrmtunedfreshbsem}{0.0159}
\newcommand{\lrmtunedfreshc}{0.3762} \newcommand{\lrmtunedfreshcsem}{0.0096}
\newcommand{\lrmstepsfinala}{1686 / 1570 / 1665}    
\newcommand{\lrmstepsmatcheda}{1657 / 1566 / 1669}  

\newcommand{\crosslossa}{5.07}     
\newcommand{\crosstoka}{20.2}      
\newcommand{\persfirstevalb}{5.86} 
\newcommand{\persfirstevalc}{6.73} 
\newcommand{\jlosshi}{6.5}         
\newcommand{\jlosslo}{6.0}         
\newcommand{\jratiosixfivec}{1.16} 
\newcommand{\jratiosixc}{0.97}     
\newcommand{\finalbaseb}{4.62} \newcommand{\finalpersb}{4.06}   
\newcommand{\finalbasec}{5.79} \newcommand{\finalpersc}{4.37}   

\newcommand{\etwonruns}{144}
\newcommand{\etwoseeds}{8}
\newcommand{\etwoseedrange}{8--15}
\newcommand{\lastseed}{7}                
\newcommand{\etwocadence}{47 / 11 / 2}       
\newcommand{\etwohorizon}{191 / 48 / 12}     
\newcommand{\etwonevalbase}{64 / 69 / 95}    
\newcommand{\etwonevalpers}{259 / 277 / 380} 
\newcommand{\etwonevalaa}{64 / 69 / 94}      
\newcommand{\etwofresh}{0.625}               
\newcommand{\etwotaua}{4.3801} \newcommand{\etwotaub}{4.9137} \newcommand{\etwotauc}{6.0866}

\newcommand{\etwoga}{4.638}                  
\newcommand{\etwogc}{-0.1181}                
\newcommand{\etwohalving}{2.319}             
\newcommand{\etwot}{148.8} \newcommand{\etwodf}{7}
\newcommand{\etwop}{8.17 \times 10^{-14}}    
\newcommand{\etwociupper}{4.817}             
\newcommand{\etwogctwoeval}{-0.1134}         
\newcommand{\etwoptwoeval}{6.90 \times 10^{-14}}
\newcommand{\etwoaasigma}{0.0172 / 0.0401 / 0.0792}  
\newcommand{\etwosigmaD}{0.0904}             
\newcommand{\etwoseedsneeded}{5}             
\newcommand{\etwopoolend}{5.639 / 1.431 / 0.9032}    

  \newcommand{\etwoRstepc}{0.878}
 
\newcommand{\etwoRfresha}{1.410}  \newcommand{\etwoRfreshc}{0.221}
  \newcommand{\etwoRsecc}{0.875}
 
  \newcommand{\etwoRjoulec}{0.879}

\newcommand{\eonenruns}{96}
\newcommand{\eonencoarse}{24}
\newcommand{\eonebudget}{25}                 
\newcommand{\eonesteps}{3048 / 760 / 188}    
\newcommand{\eoneitems}{24{,}384 / 24{,}320 / 24{,}064}   
\newcommand{\eonefwdtok}{99.6}               
\newcommand{\eonepassa}{4.7710} \newcommand{\eoneepocha}{4.2243}   
\newcommand{\eonepassb}{4.8880} \newcommand{\eoneepochb}{4.7066}   
\newcommand{\eonepassc}{5.8529} \newcommand{\eoneepochc}{5.8724}   
\newcommand{\eonedeltaa}{+0.5468} \newcommand{\eonedeltaasem}{0.0055}
\newcommand{\eonedeltab}{+0.1814} \newcommand{\eonedeltabsem}{0.0078}
\newcommand{\eonedeltac}{-0.0195} \newcommand{\eonedeltacsem}{0.0178}
\newcommand{\eonetstata}{99.1}  \newcommand{\eonepa}{2.8 \times 10^{-12}}
\newcommand{\eonetstatb}{23.3}  \newcommand{\eonepb}{6.7 \times 10^{-8}}
\newcommand{\eonetstatc}{-1.09} \newcommand{\eonepc}{0.31}
  \newcommand{\eonewinsc}{5}

\newcommand{\preregvonehash}{0c3163f}      
\newcommand{\preregvtwohash}{7ca286b}      
\newcommand{\preregvthreehash}{553bc69}    
\newcommand{\preregvfourhash}{801135f}     
\newcommand{\preregvfivehash}{c9ae507}     
\newcommand{\preregvsixhash}{244a26a}      
\newcommand{\preregvsixerratumhash}{6166ef8} 
\newcommand{\preregvsixsha}{44871071fb5d8948d0e8e175b21382dabfc52fc04691c9dc76623bbea060b1d1}
\newcommand{\osfembargo}{2027-08-10}       

\newcommand{\betaone}{0.9}              
\newcommand{\betatwo}{0.95}             
\newcommand{\weightdecay}{0.1}          
\newcommand{\gradclip}{1.0}             
\newcommand{\lrfloorfrac}{0.1}          
\newcommand{\gridratio}{2.99}
\newcommand{\evalseqs}{160}
\newcommand{\evalseed}{1234}
\newcommand{\torchtfour}{2.10.0+cu128}
\newcommand{\torchblackwell}{2.13.0+cu130}
\newcommand{\driverblackwell}{610.57.04}
\newcommand{\cudatfour}{12.8}
\newcommand{\cudablackwell}{13.0}
\newcommand{\dtwelveparams}{162\,M}     
\newcommand{\dtwelvemflop}{742}         
\newcommand{\dsixmflop}{180}            
\newcommand{\rescalefactor}{4.1}        
\newcommand{\budgetbefore}{400}         
\newcommand{\lrcoarsebefore}{six}       
\newcommand{\armsbefore}{four}          
\newcommand{\preflightgpuh}{73}         
\newcommand{\originalgpuh}{1{,}725}     
\newcommand{\preflighttokps}{48}        
\newcommand{\budgetfactor}{4}           
\newcommand{\lrcoarsefactor}{1.5}       
\newcommand{\armsfactor}{1.33}          
\newcommand{\bridgegatepct}{10}
\newcommand{\aahorizonoffset}{0.1 / 0.3 / 1.0}
\newcommand{\rrsplit}{5:4}              
\newcommand{\lfourspeedratio}{1.22}     

\newcommand{\lntwo}{0.693}                
\newcommand{\nseedsfloor}{3}              
\newcommand{\rhogridfactor}{2.99}         
\newcommand{\nrunssecondlook}{106}        
\newcommand{\nrunsaftersecondlook}{38}    
\newcommand{\nseedssecondlook}{6}         
\newcommand{\gasecondlook}{2.89}          
\newcommand{\gcsecondlook}{0.44}          
\newcommand{\sigmaDsecondlook}{0.113}     
\newcommand{\etwopredRa}{5.61}            
\newcommand{\etwopredRb}{1.42}            
\newcommand{\etwopredRc}{0.88}            
\newcommand{\etwopredJc}{0.89}            
\newcommand{\etwontauruns}{36}            
\newcommand{\fmapfa}{0.25}                
\newcommand{\fmapfb}{0.375}               
\newcommand{\fmapfc}{0.5}                 
\newcommand{\fmapfd}{0.625}               
\newcommand{\fmapfe}{0.75}                
\newcommand{\fmaptaua}{6.909}             
\newcommand{\fmaptaub}{6.498}             
\newcommand{\fmaptauc}{6.266}             
\newcommand{\fmaptaud}{6.087}             
\newcommand{\fmaptaue}{5.960}             
\newcommand{\fmappreda}{1.19}             
\newcommand{\fmappredb}{1.01}             
\newcommand{\fmappredc}{0.93}             
\newcommand{\fmappredd}{0.88}             
\newcommand{\fmapprede}{0.82}             
\newcommand{\etwotauband}{0.2}            
\newcommand{\eonedeltastar}{0.02}         
\newcommand{\eonepower}{0.8}              
\newcommand{\blackwellcap}{600}           
\newcommand{\etwogres}{\texttt{--gres=gpu:rtx6000:1}}   
\newcommand{\ethreenruns}{48}             
\newcommand{\ethreelrnruns}{12}           
\newcommand{\ethreesteps}{762}            
\newcommand{\ethreefresh}{399.5}          
\newcommand{\ethreewarmup}{15}            
\newcommand{\ethreeseed}{8}               

\newcommand{\zfourquanta}{5.96}   
\newcommand{\zfourquantb}{6.90}   
\newcommand{\zfourquantc}{8.22}   
\newcommand{\zfourjpairedhi}{1.02} \newcommand{\zfourjpairedhilo}{0.98} \newcommand{\zfourjpairedhihi}{1.06}   
\newcommand{\zfourjpairedlo}{0.85} \newcommand{\zfourjpairedlolo}{0.80} \newcommand{\zfourjpairedlohi}{0.90}   
\newcommand{\zfourEtwoseedfirst}{8} \newcommand{\zfourEtwoseedlast}{15}
\newcommand{\zfouraaspreada}{0.08} \newcommand{\zfouraaspreadb}{0.20} \newcommand{\zfouraaspreadc}{0.05}

\newcommand{\etwofmapRa}{1.018}
\newcommand{\etwofmapRloa}{0.962}
\newcommand{\etwofmapRhia}{1.078}
\newcommand{\etwofmapna}{8}
\newcommand{\etwofmapJa}{1.023}
\newcommand{\etwofmapRb}{0.952}
\newcommand{\etwofmapRlob}{0.879}
\newcommand{\etwofmapRhib}{1.031}
\newcommand{\etwofmapnb}{8}
\newcommand{\etwofmapJb}{0.953}
\newcommand{\etwofmapRc}{0.923}
\newcommand{\etwofmapRloc}{0.854}
\newcommand{\etwofmapRhic}{0.997}
\newcommand{\etwofmapnc}{8}
\newcommand{\etwofmapJc}{0.926}
\newcommand{\etwofmapRd}{0.878}
\newcommand{\etwofmapRlod}{0.805}
\newcommand{\etwofmapRhid}{0.957}
\newcommand{\etwofmapnd}{8}
\newcommand{\etwofmapJd}{0.879}
\newcommand{\etwofmapRe}{0.812}
\newcommand{\etwofmapRloe}{0.720}
\newcommand{\etwofmapRhie}{0.915}
\newcommand{\etwofmapne}{8}
\newcommand{\etwofmapJe}{0.812}

\newcommand{\etwoRpmmb}{1.347}
\newcommand{\etwoRpmmlob}{1.295}
\newcommand{\etwoRpmmhib}{1.402}

\newcommand{\etwoRpmmc}{0.797}
\newcommand{\etwoRpmmloc}{0.703}
\newcommand{\etwoRpmmhic}{0.904}
\newcommand{\etwoRpmmnc}{6}
\newcommand{\etwogpmmc}{-0.2028}

\newcommand{\etwoRpmpa}{5.516}
\newcommand{\etwoRpmploa}{5.427}
\newcommand{\etwoRpmphia}{5.607}
\newcommand{\etwoRpmpna}{8}
\newcommand{\etwoRpmpb}{1.407}
\newcommand{\etwoRpmplob}{1.369}
\newcommand{\etwoRpmphib}{1.447}
\newcommand{\etwoRpmpnb}{8}
\newcommand{\etwoRpmpc}{0.922}
\newcommand{\etwoRpmploc}{0.855}
\newcommand{\etwoRpmphic}{0.994}
\newcommand{\etwoRpmpnc}{8}
\newcommand{\etwogpmpc}{-0.0779}
\newcommand{\etwogpmpa}{4.516}
\newcommand{\etwoppmp}{9.76e-13}
\newcommand{\etwoTstepa}{5.638}
\newcommand{\etwoTsteploa}{5.582}
\newcommand{\etwoTstephia}{5.693}
\newcommand{\etwoAAstepa}{1.004}
\newcommand{\etwoAAsteploa}{0.990}
\newcommand{\etwoAAstephia}{1.019}
\newcommand{\etwoTfresha}{1.410}
\newcommand{\etwoTfreshloa}{1.396}
\newcommand{\etwoTfreshhia}{1.424}
\newcommand{\etwoAAfresha}{1.005}
\newcommand{\etwoAAfreshloa}{0.990}
\newcommand{\etwoAAfreshhia}{1.020}
\newcommand{\etwoTseca}{5.589}
\newcommand{\etwoTsecloa}{5.454}
\newcommand{\etwoTsechia}{5.727}
\newcommand{\etwoAAseca}{1.004}
\newcommand{\etwoAAsecloa}{0.985}
\newcommand{\etwoAAsechia}{1.023}
\newcommand{\etwoTjoulea}{5.647}
\newcommand{\etwoTjouleloa}{5.509}
\newcommand{\etwoTjoulehia}{5.788}
\newcommand{\etwoAAjoulea}{1.007}
\newcommand{\etwoAAjouleloa}{0.988}
\newcommand{\etwoAAjoulehia}{1.026}

\newcommand{\primPSpersseca}{0.991}
\newcommand{\primPSperssecloa}{0.973}
\newcommand{\primPSperssechia}{1.010}
\newcommand{\primPSpersjoulea}{0.995}
\newcommand{\primPSpersjouleloa}{0.979}
\newcommand{\primPSpersjoulehia}{1.011}
\newcommand{\primPSaaseca}{0.996}
\newcommand{\primPSaasecloa}{0.958}
\newcommand{\primPSaasechia}{1.036}
\newcommand{\primPSaajoulea}{0.998}
\newcommand{\primPSaajouleloa}{0.969}
\newcommand{\primPSaajoulehia}{1.028}
\newcommand{\etwoTstepb}{1.425}
\newcommand{\etwoTsteplob}{1.386}
\newcommand{\etwoTstephib}{1.464}
\newcommand{\etwoAAstepb}{1.005}
\newcommand{\etwoAAsteplob}{0.972}
\newcommand{\etwoAAstephib}{1.039}
\newcommand{\etwoTfreshb}{0.357}
\newcommand{\etwoTfreshlob}{0.347}
\newcommand{\etwoTfreshhib}{0.367}
\newcommand{\etwoAAfreshb}{1.007}
\newcommand{\etwoAAfreshlob}{0.974}
\newcommand{\etwoAAfreshhib}{1.042}
\newcommand{\etwoTsecb}{1.417}
\newcommand{\etwoTseclob}{1.383}
\newcommand{\etwoTsechib}{1.452}
\newcommand{\etwoAAsecb}{1.001}
\newcommand{\etwoAAseclob}{0.968}
\newcommand{\etwoAAsechib}{1.035}
\newcommand{\etwoTjouleb}{1.423}
\newcommand{\etwoTjoulelob}{1.383}
\newcommand{\etwoTjoulehib}{1.465}
\newcommand{\etwoAAjouleb}{1.005}
\newcommand{\etwoAAjoulelob}{0.975}
\newcommand{\etwoAAjoulehib}{1.035}

\newcommand{\primPSperssecb}{0.965}
\newcommand{\primPSpersseclob}{0.928}
\newcommand{\primPSperssechib}{1.004}
\newcommand{\primPSpersjouleb}{0.974}
\newcommand{\primPSpersjoulelob}{0.946}
\newcommand{\primPSpersjoulehib}{1.003}
\newcommand{\primPSaasecb}{0.978}
\newcommand{\primPSaaseclob}{0.938}
\newcommand{\primPSaasechib}{1.020}
\newcommand{\primPSaajouleb}{0.985}
\newcommand{\primPSaajoulelob}{0.953}
\newcommand{\primPSaajoulehib}{1.019}
\newcommand{\etwoTstepc}{0.878}
\newcommand{\etwoTsteploc}{0.805}
\newcommand{\etwoTstephic}{0.957}
\newcommand{\etwoAAstepc}{0.981}
\newcommand{\etwoAAsteploc}{0.918}
\newcommand{\etwoAAstephic}{1.048}
\newcommand{\etwoTfreshc}{0.221}
\newcommand{\etwoTfreshloc}{0.204}
\newcommand{\etwoTfreshhic}{0.241}
\newcommand{\etwoAAfreshc}{0.986}
\newcommand{\etwoAAfreshloc}{0.921}
\newcommand{\etwoAAfreshhic}{1.056}
\newcommand{\etwoTsecc}{0.875}
\newcommand{\etwoTsecloc}{0.803}
\newcommand{\etwoTsechic}{0.955}
\newcommand{\etwoAAsecc}{0.979}
\newcommand{\etwoAAsecloc}{0.916}
\newcommand{\etwoAAsechic}{1.046}
\newcommand{\etwoTjoulec}{0.879}
\newcommand{\etwoTjouleloc}{0.799}
\newcommand{\etwoTjoulehic}{0.967}
\newcommand{\etwoAAjoulec}{0.984}
\newcommand{\etwoAAjouleloc}{0.914}
\newcommand{\etwoAAjoulehic}{1.060}

\newcommand{\etwoPSperssecc}{0.997}
\newcommand{\etwoPSperssecloc}{0.995}
\newcommand{\etwoPSperssechic}{0.999}
\newcommand{\etwoPSpersjoulec}{1.001}
\newcommand{\etwoPSpersjouleloc}{0.984}
\newcommand{\etwoPSpersjoulehic}{1.019}
\newcommand{\etwoPSaasecc}{0.998}
\newcommand{\etwoPSaasecloc}{0.997}
\newcommand{\etwoPSaasechic}{1.000}
\newcommand{\etwoPSaajoulec}{1.004}
\newcommand{\etwoPSaajouleloc}{0.984}
\newcommand{\etwoPSaajoulehic}{1.024}
\newcommand{\primPSperssecc}{0.988}
\newcommand{\primPSperssecloc}{0.905}
\newcommand{\primPSperssechic}{1.079}
\newcommand{\primPSpersjoulec}{1.016}
\newcommand{\primPSpersjouleloc}{0.989}
\newcommand{\primPSpersjoulehic}{1.044}
\newcommand{\primPSaasecc}{0.977}
\newcommand{\primPSaasecloc}{0.899}
\newcommand{\primPSaasechic}{1.061}
\newcommand{\primPSaajoulec}{1.004}
\newcommand{\primPSaajouleloc}{0.984}
\newcommand{\primPSaajoulehic}{1.025}
\newcommand{\etwoAbsSecbasea}{174}
\newcommand{\etwoAbsKJbasea}{72.0}
\newcommand{\etwoAbsWbasea}{413}

\newcommand{\etwoAbsSecpersa}{974}
\newcommand{\etwoAbsKJpersa}{406.9}
\newcommand{\etwoAbsWpersa}{418}

\newcommand{\etwoAbsSecaaa}{175}
\newcommand{\etwoAbsKJaaa}{72.5}
\newcommand{\etwoAbsWaaa}{415}

\newcommand{\etwoAbsSecbaseb}{177}
\newcommand{\etwoAbsKJbaseb}{74.0}
\newcommand{\etwoAbsWbaseb}{417}

\newcommand{\etwoAbsSecpersb}{251}
\newcommand{\etwoAbsKJpersb}{105.4}
\newcommand{\etwoAbsWpersb}{419}

\newcommand{\etwoAbsSecaab}{178}
\newcommand{\etwoAbsKJaab}{74.4}
\newcommand{\etwoAbsWaab}{419}

\newcommand{\etwoAbsSecbasec}{186}
\newcommand{\etwoAbsKJbasec}{77.8}
\newcommand{\etwoAbsWbasec}{418}
\newcommand{\etwoAbsNbasec}{8}
\newcommand{\etwoAbsSecpersc}{162}
\newcommand{\etwoAbsKJpersc}{68.1}
\newcommand{\etwoAbsWpersc}{420}

\newcommand{\etwoAbsSecaac}{182}
\newcommand{\etwoAbsKJaac}{76.4}
\newcommand{\etwoAbsWaac}{420}

\newcommand{\etwoXfracbaseloa}{56}
\newcommand{\etwoXfracbasehia}{58}

\newcommand{\etwoXfracpersloa}{79}
\newcommand{\etwoXfracpershia}{83}

\newcommand{\etwoXfracbaselob}{56}
\newcommand{\etwoXfracbasehib}{62}

\newcommand{\etwoXfracperslob}{20}
\newcommand{\etwoXfracpershib}{22}

\newcommand{\etwoXstepbaseloc}{104}
\newcommand{\etwoXstepbasehic}{138}
\newcommand{\etwoXfracbaseloc}{55}
\newcommand{\etwoXfracbasehic}{72}
\newcommand{\etwoXlrbaseloc}{0.26}
\newcommand{\etwoXlrbasehic}{0.49}
\newcommand{\etwoXplannedbasec}{190}
\newcommand{\etwoXsteppersloc}{101}
\newcommand{\etwoXsteppershic}{109}
\newcommand{\etwoXfracpersloc}{13}
\newcommand{\etwoXfracpershic}{14}
\newcommand{\etwoXlrpersloc}{0.97}

\newcommand{\etwoXplannedpersc}{762}

\newcommand{\etwoXfracaaloc}{55}
\newcommand{\etwoXfracaahic}{66}

\newcommand{\primXplannedbasea}{3051}

\newcommand{\primXplannedpersa}{12207}

\newcommand{\primXplannedbaseb}{762}

\newcommand{\primXplannedpersb}{3051}

\newcommand{\primXstepbaseloc}{30}
\newcommand{\primXstepbasehic}{34}
\newcommand{\primXfracbaseloc}{16}
\newcommand{\primXfracbasehic}{18}

\newcommand{\primXplannedbasec}{190}

\newcommand{\primXfracpersloc}{6}

\newcommand{\primXplannedpersc}{762}

\newcommand{\etwoWinEtabaseb}{8}
\newcommand{\etwoWinEtapersb}{7}

\newcommand{\etwoWinEtabasec}{8}
\newcommand{\etwoWinEtapersc}{8}

\newcommand{\rhogrida}{2.99}

\newcommand{\coarseBaseAtEtaa}{5.186}
\newcommand{\coarseBaseAtThreeEtaa}{5.213}
\newcommand{\rhogridb}{1.00}

\newcommand{\coarseBaseAtEtab}{5.996}
\newcommand{\coarseBaseAtThreeEtab}{6.369}
\newcommand{\rhogridc}{1.00}

\newcommand{\coarseBaseAtEtac}{6.996}
\newcommand{\coarseBaseAtThreeEtac}{7.262}
\newcommand{\primDmean}{1.937}
\newcommand{\primDlb}{1.842}
\newcommand{\primSignP}{0.0039}
\newcommand{\primMDE}{0.124}
\newcommand{\primMDEreg}{0.105}
\newcommand{\primSigmaDraw}{0.1132}
\newcommand{\primSeedsRaw}{7.2}
\newcommand{\etwoSigmaDlog}{0.0999}
\newcommand{\adjT}{70.1}
\newcommand{\adjSigmaD}{0.1166}
\newcommand{\adjNbelowOne}{3}
\newcommand{\adjgmfive}{0.0048}
\newcommand{\adjgmtwo}{0.0005}
\newcommand{\adjgptwo}{-0.0082}
\newcommand{\adjgpfive}{-0.0120}

\newcommand{\firstEvalMinbaseb}{6.6932}
\newcommand{\firstEvalMinpersb}{5.7606}
\newcommand{\firstEvalMinaab}{6.7631}
\newcommand{\firstEvalMinbasec}{7.6964}
\newcommand{\firstEvalMinpersc}{6.6956}
\newcommand{\firstEvalMinaac}{7.7232}
\newcommand{\eoneCIloa}{0.5337}
\newcommand{\eoneCIhia}{0.5598}

\newcommand{\eoneCIlob}{0.1630}
\newcommand{\eoneCIhib}{0.1998}

\newcommand{\eoneCIloc}{-0.0615}
\newcommand{\eoneCIhic}{0.0226}
\newcommand{\eoneSDc}{0.0503}
\newcommand{\eoneSignPc}{0.73}
\newcommand{\eoneNneededc}{50}
\newcommand{\hoursTfour}{279}
\newcommand{\nrunsTfour}{178}
\newcommand{\hoursLfour}{1}
\newcommand{\nrunsLfour}{5}
\newcommand{\hoursBlackwell}{37}
\newcommand{\nrunsBlackwell}{294}

\newcommand{\nrunsall}{477}
\newcommand{\looga}{2.868}
\newcommand{\loogc}{0.4136}
\newcommand{\loot}{34.14}
\newcommand{\loopval}{2.10e-08}
\newcommand{\loosigmaD}{0.1507}
\newcommand{\loon}{7}
\newcommand{\loostepc}{1.414}
\newcommand{\loosteploc}{1.353}
\newcommand{\loostephic}{1.477}

\newcommand{\loosecc}{1.446}
\newcommand{\loosecloc}{1.398}
\newcommand{\loosechic}{1.496}
\newcommand{\looAAsecc}{1.024}
\newcommand{\looAAsecloc}{0.959}
\newcommand{\looAAsechic}{1.095}
\newcommand{\loojoulec}{1.444}
\newcommand{\loojouleloc}{1.405}
\newcommand{\loojoulehic}{1.485}
\newcommand{\looAAjoulec}{1.024}
\newcommand{\looAAjouleloc}{0.969}
\newcommand{\looAAjoulehic}{1.082}
\newcommand{\loofreshc}{0.377}
\newcommand{\loofreshloc}{0.361}
\newcommand{\loofreshhic}{0.394}

\newcommand{\adjRstepa}{3.886}
\newcommand{\adjRsteploa}{3.773}
\newcommand{\adjRstephia}{4.002}
\newcommand{\adjAAstepa}{0.998}
\newcommand{\adjAAsteploa}{0.976}
\newcommand{\adjAAstephia}{1.020}
\newcommand{\adjRfresha}{0.973}
\newcommand{\adjRfreshloa}{0.945}
\newcommand{\adjRfreshhia}{1.002}
\newcommand{\adjAAfresha}{1.005}
\newcommand{\adjAAfreshloa}{0.983}
\newcommand{\adjAAfreshhia}{1.028}
\newcommand{\adjRseca}{3.852}
\newcommand{\adjRsecloa}{3.739}
\newcommand{\adjRsechia}{3.968}
\newcommand{\adjAAseca}{0.994}
\newcommand{\adjAAsecloa}{0.964}
\newcommand{\adjAAsechia}{1.025}
\newcommand{\adjRjoulea}{3.866}
\newcommand{\adjRjouleloa}{3.760}
\newcommand{\adjRjoulehia}{3.975}
\newcommand{\adjAAjoulea}{0.996}
\newcommand{\adjAAjouleloa}{0.971}
\newcommand{\adjAAjoulehia}{1.021}
\newcommand{\adjRstepb}{1.157}
\newcommand{\adjRsteplob}{1.107}
\newcommand{\adjRstephib}{1.209}
\newcommand{\adjAAstepb}{0.987}
\newcommand{\adjAAsteplob}{0.948}
\newcommand{\adjAAstephib}{1.027}
\newcommand{\adjRfreshb}{0.293}
\newcommand{\adjRfreshlob}{0.281}
\newcommand{\adjRfreshhib}{0.307}
\newcommand{\adjAAfreshb}{1.004}
\newcommand{\adjAAfreshlob}{0.964}
\newcommand{\adjAAfreshhib}{1.045}
\newcommand{\adjRsecb}{1.138}
\newcommand{\adjRseclob}{1.070}
\newcommand{\adjRsechib}{1.211}
\newcommand{\adjAAsecb}{0.965}
\newcommand{\adjAAseclob}{0.906}
\newcommand{\adjAAsechib}{1.028}
\newcommand{\adjRjouleb}{1.146}
\newcommand{\adjRjoulelob}{1.084}
\newcommand{\adjRjoulehib}{1.212}
\newcommand{\adjAAjouleb}{0.972}
\newcommand{\adjAAjoulelob}{0.918}
\newcommand{\adjAAjoulehib}{1.029}
\newcommand{\adjRstepc}{0.995}
\newcommand{\adjRsteploc}{0.934}
\newcommand{\adjRstephic}{1.059}
\newcommand{\adjAAstepc}{1.002}
\newcommand{\adjAAsteploc}{0.925}
\newcommand{\adjAAstephic}{1.086}
\newcommand{\adjRfreshc}{0.263}
\newcommand{\adjRfreshloc}{0.247}
\newcommand{\adjRfreshhic}{0.280}
\newcommand{\adjAAfreshc}{1.059}
\newcommand{\adjAAfreshloc}{0.980}
\newcommand{\adjAAfreshhic}{1.145}
\newcommand{\adjRsecc}{0.993}
\newcommand{\adjRsecloc}{0.900}
\newcommand{\adjRsechic}{1.096}
\newcommand{\adjAAsecc}{0.987}
\newcommand{\adjAAsecloc}{0.877}
\newcommand{\adjAAsechic}{1.111}
\newcommand{\adjRjoulec}{1.014}
\newcommand{\adjRjouleloc}{0.960}
\newcommand{\adjRjoulehic}{1.071}
\newcommand{\adjAAjoulec}{1.010}
\newcommand{\adjAAjouleloc}{0.932}
\newcommand{\adjAAjoulehic}{1.094}

\newcommand{\primTauPrimeRstep}{0.865}
\newcommand{\primTauPrimeRsteplo}{0.812}
\newcommand{\primTauPrimeRstephi}{0.922}
\newcommand{\primTauPrimeRnstep}{8}
\newcommand{\primTauPrimeRjoule}{0.885}
\newcommand{\primTauPrimeRjoulelo}{0.846}
\newcommand{\primTauPrimeRjoulehi}{0.927}

\newcommand{\etwoDmean}{4.756}
\newcommand{\etwoDlb}{4.696}
\newcommand{\etwoSignP}{0.0039}
\newcommand{\etwoNpositive}{8}
\newcommand{\etwoSeedsLog}{5.6}
\newcommand{\etwoSeedsLogCeil}{6}
\newcommand{\etwoSeedsRaw}{4.6}
\newcommand{\cardspeedratio}{8.0}
\newcommand{\cardspeedTfourSec}{499}
\newcommand{\cardspeedBlackwellSec}{62}
\newcommand{\etworhoprimea}{0.33}
\newcommand{\etworhoprimeb}{1.00}
\newcommand{\etworhoprimec}{1.00}
\newcommand{\hourscoarseTfour}{14.3}
\newcommand{\nrunscoarseTfour}{31}
\newcommand{\hourscoarseLfour}{1.0}
\newcommand{\nrunscoarseLfour}{5}
\newcommand{\hoursfinalTfour}{234.7}
\newcommand{\nrunsfinalTfour}{138}
\newcommand{\hoursfinalBlackwell}{1.0}
\newcommand{\nrunsfinalBlackwell}{6}
\newcommand{\hourslrmTfour}{30.4}
\newcommand{\nrunslrmTfour}{9}
\newcommand{\hoursbridgeBlackwell}{2.0}
\newcommand{\nrunsbridgeBlackwell}{24}
\newcommand{\hourseonecoarseBlackwell}{0.5}
\newcommand{\nrunseonecoarseBlackwell}{24}
\newcommand{\hourseoneBlackwell}{8.0}
\newcommand{\nrunseoneBlackwell}{96}
\newcommand{\hoursetwoBlackwell}{25.0}
\newcommand{\nrunsetwoBlackwell}{144}
\newcommand{\eoneanchorBhundreda}{4.2072}
\newcommand{\eoneanchorBhundredna}{8}
\newcommand{\eoneanchorBhundredb}{4.6668}

\newcommand{\eoneanchorBhundredc}{5.8467}

\newcommand{\eonebestdeltaa}{0.5487}
\newcommand{\eonebestloa}{0.5352}
\newcommand{\eonebesthia}{0.5621}

\newcommand{\eonebestdeltab}{0.1826}
\newcommand{\eonebestlob}{0.1642}
\newcommand{\eonebesthib}{0.2009}

\newcommand{\eonebestdeltac}{-0.0193}
\newcommand{\eonebestloc}{-0.0613}
\newcommand{\eonebesthic}{0.0226}

\newcommand{\ratepairbaseLoa}{8.96e-04}
\newcommand{\ratepairbaseHia}{2.68e-03}
\newcommand{\ratepairpersLoa}{8.96e-04}
\newcommand{\ratepairpersHia}{2.68e-03}
\newcommand{\ratepairaaLoa}{8.96e-04}
\newcommand{\ratepairaaHia}{2.68e-03}
\newcommand{\ratepairbaseLob}{3.00e-04}
\newcommand{\ratepairbaseHib}{8.96e-04}
\newcommand{\ratepairpersLob}{8.96e-04}
\newcommand{\ratepairpersHib}{2.68e-03}
\newcommand{\ratepairaaLob}{8.96e-04}
\newcommand{\ratepairaaHib}{2.68e-03}
\newcommand{\ratepairbaseLoc}{8.96e-04}
\newcommand{\ratepairbaseHic}{2.68e-03}
\newcommand{\ratepairpersLoc}{3.00e-04}
\newcommand{\ratepairpersHic}{8.96e-04}
\newcommand{\ratepairaaLoc}{8.96e-04}
\newcommand{\ratepairaaHic}{2.68e-03}
\newcommand{\coarseseed}{0}
\newcommand{\eoneseedrange}{0--7}

\newcommand{\ethreehours}{16.8}
\newcommand{\ethreeRstep}{1.036}
\newcommand{\ethreeRsteplo}{0.983}
\newcommand{\ethreeRstephi}{1.091}
\newcommand{\ethreeRstepn}{8}

\newcommand{\ethreeAAstep}{0.995}
\newcommand{\ethreeAAsteplo}{0.941}
\newcommand{\ethreeAAstephi}{1.053}

\newcommand{\ethreeRsec}{1.028}
\newcommand{\ethreeRseclo}{0.983}
\newcommand{\ethreeRsechi}{1.076}

\newcommand{\ethreeRjoule}{1.046}
\newcommand{\ethreeRjoulelo}{0.992}
\newcommand{\ethreeRjoulehi}{1.103}

\newcommand{\ethreeXbaselo}{94}
\newcommand{\ethreeXbasehi}{109}
\newcommand{\ethreeXfracbaselo}{12}

\newcommand{\ethreeXlrbasehi}{0.98}
\newcommand{\ethreeWinEtabase}{8}

\newcommand{\ethreeXperslo}{98}
\newcommand{\ethreeXpershi}{110}

\newcommand{\ethreeXfracpershi}{14}
\newcommand{\ethreeXlrperslo}{0.96}

\newcommand{\ethreeXaalo}{93}
\newcommand{\ethreeXaahi}{106}

\newcommand{\ethreelrnrunsarchive}{12}
\newcommand{\ethreelrhours}{2.1}

\newcommand{\ethreelrStepsbaseA}{168}

\newcommand{\ethreelrStepsbaseB}{117}

\newcommand{\ethreelrBestbaseC}{6.264}

\newcommand{\ethreelrBestbaseD}{6.641}

\newcommand{\ethreelrStepspersA}{130}

\newcommand{\ethreelrStepspersB}{104}

\newcommand{\ethreelrStepspersC}{149}

\newcommand{\ethreelrStepspersD}{233}

\newcommand{\ethreelrStepsaaB}{123}

\newcommand{\ethreelrRho}{1.00}
\newcommand{\fwconfig}{sample-10BT}
\newcommand{\fwtrainshards}{12}
\newcommand{\fwshardMtok}{100}
\newcommand{\fwtrainGtok}{1.2}
\newcommand{\fwvalMtok}{20}
\newcommand{\primXstepsbasec}{191}
\newcommand{\primXstepspersc}{761}
\newcommand{\ngroupsprov}{9}
\newcommand{\ngroupsfinal}{24}
\newcommand{\rhotwoa}{1.21}   
\newcommand{\rhotwob}{1.72}   
\newcommand{\rhotwoc}{0.68}   
\newcommand{\fwerlevel}{0.10} 
\newcommand{\primPSmaxdevpct}{4}  
\newcommand{\lrdispEoneCoarse}{8.96 \times 10^{-4}}  
\newcommand{\lrdispCoarseBase}{8.96 \times 10^{-4}}  
\newcommand{\lrdispCoarsePersa}{2.68 \times 10^{-3}} 

\title{Minibatch persistency, eight years later:\\
what batch reuse costs in steps and joules,\\
and what it saves in data}

\author{\name Matteo Fischetti \email matteo.fischetti@unipd.it \\
      \addr Department of Information Engineering\\
      University of Padova, Italy}

\def\month{MM}  
\def\year{YYYY} 
\def\openreview{\url{https://openreview.net/forum?id=XXXX}} 

\begin{document}
\maketitle

\begin{abstract}
Minibatch persistency reuses data instead of reading it: rather than drawing a
fresh minibatch at every optimizer step, it takes $K$ consecutive steps
on the same one. Absorbed into \emph{data echoing} in 2019, it has carried one
objection --- that reuse merely imitates a larger learning rate --- and no
baseline tuned as carefully as the method itself. This paper runs the missing
test. A pre-registered study trains a
\nparams{}-parameter Transformer on FineWeb-Edu at minibatch size
$B\in\{32,128,512\}$,
\nseeds{} seeds per cell, tuning the learning rate separately for every batch
size and every arm, against a reuse-free control that changes the sampling and
nothing else. Each headline claim is a cost to reach a fixed loss, read on four
axes: optimizer steps, fresh tokens, seconds, and joules at the socket. We then
replicate on new seeds and a newer GPU generation, and put the three arms on one
schedule in steps.

The outcome of our study is that what minibatch reuse buys is neither speed nor
energy but data, and only at large minibatch size: at $B=32$ it reads more fresh
tokens than the baseline, not fewer. On steps, seconds and joules it is at best
free; and at $B=512$, where it looks best, a registered control cannot separate
the effect of reuse from the position on the learning-rate schedule at
$n=\nseeds{}$ seeds. The technique is therefore worth using where fresh data
rather than
compute is the binding cost: a corpus that runs out, a pipeline
that pays per sample, a stream that cannot be rewound. Where the data can simply
be read again, spaced epochs do as well or better.

\end{abstract}

\section{Introduction}
\label{sec:intro}

Training a modern neural network is an exercise in paying for gradients. Every
optimizer step consumes a minibatch, and producing that minibatch is not free:
it must be read from storage, decoded, augmented, collated and moved to the
accelerator, and in a large fraction of real pipelines this work---not the
matrix multiplications---is what sets the pace. The natural question is
whether each minibatch has to be spent after a single use. If a batch of $B$
examples is a good enough sample of the data distribution, one could take $K$
consecutive optimizer steps on it before paying for the next one, thus dividing
the cost of the input pipeline by $K$. We call this \emph{minibatch
persistency} in what follows, following the name introduced
by~\citet{fischetti2018}.

The idea is eight years old, and its history is instructive. It was proposed in
2018 with encouraging but small-scale evidence, and thirteen months later it
was cited---from the first preprint version---by \citet{choi2019}, who
generalized it into a family of techniques called \emph{data echoing} and, in
the same paragraph, stated the objection that has shadowed it ever since:
\begin{quote}
``Fischetti et al.\ (2018) describe a special case of data echoing they call
`minibatch persistency' that reuses minibatches for multiple consecutive SGD
updates. They run experiments on CIFAR-10, but do not tune hyperparameters for
the baseline or for their method. Neither their method nor their baseline reach
competitive test numbers in their experiments, leaving open the question of
whether minibatch persistency has an advantage over a well-tuned baseline.''
\end{quote}
The objection is correct, and it is a methodological one, not a conceptual one:
the criticism is not that the idea is wrong but that the 2018 evidence could not
tell. It is also, by now, the standard criticism of the whole efficient-training
literature, codified by \citet{shallue2019} and sharpened by \citet{kaddour2023}
into the observation that most claimed training speed-ups evaporate once the
baseline is tuned with the same care as the method. Note that the objection
bites with particular force here. At first order, taking $K$ steps on the same
batch at learning rate $\eta$ resembles a single step at a learning rate of
about $K\eta$; hence an untuned comparison between $K>1$ and $K=1$ may be
measuring nothing but a learning rate, and any honest revival of the idea has to
rule that out before it says anything else.

The conjecture underneath the method has since been examined once on tuned
baselines. \citet{choi2019} report, for batch echoing at $e = 2$, that repeated
batches approximate fresh batches better as the batch size approaches the
training set size---the 2018 conjecture in their words
(Section~\ref{sec:related}). Two things, however, were left undone. First, the
larger reuse factors in that paper belong to \emph{example} echoing,
in which duplicated examples are reshuffled into fresh batches; \emph{batch}
echoing, which is minibatch persistency in the strict sense, was swept only up
to $e=2$ and is the weakest member of the family. Second, and more surprising
for a technique whose entire rationale is saving work, nobody has ever reported
what it saves in joules.

This paper takes up both. Its aim is not to propose the idea, which
\citet{fischetti2018} did, but to measure it under the conditions the 2019
objection demands. The main contributions are as follows.
\begin{itemize}
  \item \textbf{A tuned answer to the tuned-baseline objection.} We tune the
    learning rate independently inside every (batch size, arm) cell, with the
    same number of learning rates per arm within a batch size, and we report
    $\rho(B) = \eta^\star(K{=}4)/\eta^\star(K{=}1)$ as a first-class result,
    where $\eta^\star(K)$ is the cost-minimizing learning rate of the arm with
    reuse factor $K$. It is a registered descriptive diagnostic, read from a
    single-seed coarse sweep with no interval, of whether persistency is
    anything more than a re-parametrized learning rate. At $B = 32$ and
    $B = 128$ it is inconsistent with that reading; at $B = 512$ it is
    reported as indeterminate.
  \item \textbf{A pre-registered confirmatory design.} The hypothesis, the
    endpoint, the target definition, the estimator, the exclusion rules and the
    stopping rule for the seed count were registered on the OSF Registries
    before the confirmatory runs, and the analysis decisions taken afterwards
    were themselves timestamped. Section~\ref{sec:setup} describes the
    apparatus and Section~\ref{sec:availability} lists what a reader receives
    to check it, so that the order in which things happened need not be taken
    on our word.
  \item \textbf{Cost measured on four axes, including energy.} Every evaluation
    logs optimizer steps, fresh tokens, wall-clock seconds and joules, and every
    claim except the epoch contrast of Section~\ref{sec:results:e1} is a
    cost-to-target claim. To our knowledge this is the first study of
    batch reuse that reports joules at all, let alone joules with a stated
    provenance and a per-device attribution argument
    (Section~\ref{sec:energy}).
  \item \textbf{A calibrated noise floor.} Alongside the baseline and the
    persistency arm we run an A/A arm that is distributionally identical to the
    baseline by construction. Its measured ratio to the baseline has true value
    one, so its dispersion across seeds tells the reader---and told us---how
    small a ratio this apparatus can honestly resolve.
\end{itemize}

Two claims run through the paper, and we keep them separable on purpose. The
first is an \emph{optimization} claim: at a fixed target validation loss, how
many optimizer steps and how many fresh tokens does reuse cost, and how does
that penalty behave as the batch size grows. The second is an \emph{energy}
claim: on a pipeline where producing data costs real work, how many joules does
reuse save at the same target. They are supported by different evidence and can
fail independently---in particular, an energy saving that turned out to be
entirely explained by the input pipeline would say nothing about optimization,
and we would say so in those words rather than let the two be read as one.

\paragraph{Who the data reading is for.} The optimization claim has a
practical reader. Where unique data is the scarce resource---low-resource
languages, domain corpora, licensed or annotated text, the data-constrained
regime of \citet{muennighoff2023}---reuse at large batch reaches the same loss
in the same number of steps on \etwoRfreshc{} of the fresh tokens. Where every
fresh example costs work before it reaches the accelerator---decoding and
augmentation, simulation, environment rollouts and policy generations in
reinforcement learning, synthetic data---the tokens saved are the pipeline's
cost, which is the setting data echoing was built for \citep{choi2019}. And
where the stream cannot be rewound, reuse is the only way to take more than
one step per arriving batch. Two limits travel with the offer. It holds at
large batch: at $B = 32$ reuse costs data rather than saving it. And where
the data can be revisited, four epochs spaced out are at least as good as
immediate reuse (Section~\ref{sec:results:e1}), so persistency earns its place
only where a second pass is impossible or a fresh sample has a price.

The paper is organized as follows. Section~\ref{sec:related} places the idea in
the literature, including a recent theoretical line that studies batch reuse
without reference to either the 2018 or the 2019 work. Section~\ref{sec:method}
defines the two knobs of the design and the cost-to-target estimator.
Section~\ref{sec:setup} describes the experimental apparatus and the
pre-registration. Section~\ref{sec:energy} is devoted to the measurement of
energy, which turned out to be the hardest part of the study and deserves a
section of its own. Section~\ref{sec:results} reports the results. Finally,
conclusions and directions for future research are drawn in
Section~\ref{sec:conclusions}.

\section{Related work}
\label{sec:related}

\paragraph{Minibatch persistency.}
The technique was proposed by \citet{fischetti2018} on small-scale evidence,
with no baseline tuned to the same effort and no reading of how the effect moves
with the batch size; supplying that test is the present study. We next review
the four lines of work that bear on batch reuse, and say plainly where our study
sits with respect to each.

\paragraph{Data echoing and the fate of the idea.}
\citet{choi2019} introduced data echoing as a family of methods that insert a
repetition point somewhere in the input pipeline and reuse the output of every
upstream stage $e$ times. Minibatch persistency is the member of that family in
which the repetition point sits after batching, so the reused object is the
assembled minibatch itself. Their empirical study covers Transformer on LM1B,
ResNet-50 on ImageNet and SSD on COCO, with quasi-random hyperparameter search
(of the order of one hundred trials per configuration) for both the baseline
and the treatment. We adopt their requirement that baseline and treatment be
tuned with the same effort, at a far smaller search budget: \nlrcoarse{} rates
per cell against their hundred trials. Two of their findings frame our own.
Their Section~3.4 sweeps the batch size for batch echoing at $e = 2$, on
Transformer/LM1B and ResNet-50/ImageNet, and reports that ``as the batch size
increases, the performance of batch echoing relative to the baseline either
stays the same or improves'' (their Figure~6). That is the 2018 conjecture,
tested once on tuned baselines and at the smallest reuse factor. The larger
reuse factors in that paper belong to \emph{example} echoing, where duplicated
examples are reshuffled into new batches. What that leaves to the present
study is the reuse factor $K = \Kpers{}$, the cost-to-target reading on four
axes including joules, and the pre-registration. The paper itself was not
accepted at ICLR 2020 and remains an unrefereed preprint, with a substantial
citation record.

\paragraph{Reuse elsewhere in systems and optimization.}
The mechanism has been rediscovered wherever the input pipeline is expensive.
\citet{ramezani2020} recycle a sampled minibatch across several steps in graph
neural network training, with convergence guarantees, and cite the 2018 work by
name; \citet{agarwal2020} analyze stochastic optimization when the data pipeline
lags the accelerator, and show that reuse accelerates the curvature-dominated
phase of convergence while leaving the statistical rate intact, a prediction
this study does not test. We do not claim novelty of the mechanism itself.

\paragraph{The theory of batch reuse, 2024--2025.}
A recent theoretical line studies what changes when the same batch is used more
than once, and reaches a conclusion stronger than a throughput argument:
repetition changes \emph{what is learnable}. \citet{dandi2024} show that several
gradient steps on the same batch allow two-layer networks to learn functions
that single-pass SGD provably cannot, and related work extends the picture to
multi-index targets \citep{repetita2024} and to scaling laws under data reuse
\citep{linwu2025}. At the time of writing, none of these three works cites
either the 2018 or the
2019 paper---the bridge between the theory of repetition and the practice of
echoing has not been built. We do not build it here either; our contribution is
empirical. But it is the reason we consider the question worth reopening rather
than closed by \citet{choi2019}, and we return to it in
Section~\ref{sec:conclusions}.

\paragraph{Repetition at the epoch level.}
A separate and by now settled question is whether repeating \emph{data} pays at
all: \citet{muennighoff2023} show that, in the data-constrained regime, up to
roughly four epochs of repeated tokens are almost as good as fresh ones. That
result concerns repetition reshuffled at epoch scale, not $K$ consecutive steps
on an intact batch, and it does not transfer automatically---both the reshuffling
theory and the echoing experiments indicate that reshuffled repeats beat
identical ones. We flag this asymmetry here because it is the honest weakness of
the pure form we test: the strict variant, with the batch left intact and no
augmentation between repeats, sits at the unfavorable end of the reuse spectrum.

\paragraph{Methodology and measurement.}
Our protocol follows the training-benchmark tradition of \citet{shallue2019},
\citet{kaddour2023} and the AlgoPerf benchmark of \citet{dahl2023}: tuned
baselines, cost-to-target curves rather than accuracy at a fixed number of
epochs, and speed-ups reported against a budget that both arms actually spend.
On the energy side we follow the measurement practice established by
\citet{you2023zeus} and \citet{chung2024perseus}, who read GPU energy from the
device itself rather than from a model of it; Section~\ref{sec:energy} explains
why we adopt per-device counters as the energy axis, and why a single
node-level reading serves only to size what the device counter cannot see,
not as a reported quantity. The regulatory context, finally, is
no longer hypothetical: Annex XI of the EU AI Act \citep{euaiact} requires the
energy consumption of general-purpose model training to be documented, which
makes the joule a reporting unit and not merely a scientific curiosity.

\section{What is measured, and how}
\label{sec:method}

In this section we define the design under test, the cost axes on which it is
evaluated, and the estimator that turns a training curve into a number. The
mechanism itself is trivial---this is part of the appeal of the idea---so the
seriousness of the study has to live in the apparatus around it.

\subsection{Two knobs, not one}
\label{sec:method:knobs}

Reuse is usually described by a single number, the number of times a batch is
consumed. We found that one number conflates two decisions, and we therefore
parametrize the design by two.

\begin{itemize}
  \item $K$ := the number of consecutive optimizer steps taken between two
    refills of the data pipeline.
  \item $M$ := the size of the \emph{pool} refilled at each such point, measured
    in batches; the pool therefore holds $M \cdot B$ fresh examples.
\end{itemize}

The ratio $K/M$ is the \emph{echo factor}: the number of optimizer steps the
pipeline is asked to feed per batch of fresh data. Given a pool, steps are drawn
from it in one of two modes. In \texttt{repeat} the pool is a single batch,
passed to the optimizer $K$ times unchanged. In \texttt{resample} the pool is
cut into $K$ disjoint batches by a random permutation when $M = K$, and one is
consumed per step---drawing is without replacement. When $K > M$ the
permutation is redrawn as needed, so with $M = 1$ every step sees a
permutation of the same batch. Three arms follow, and they are the whole
experiment:

\begin{center}
\begin{tabular}{llll}
\toprule
arm & $K$ & $M$, mode & role \\
\midrule
baseline    & $1$ & $1$, --- & control, tuned independently \\
persistency & $\Kpers$ & $1$, \texttt{repeat} & the 2018 treatment (= batch echoing) \\
A/A null    & $\Kpers$ & $\Kpers$, \texttt{resample} & calibration, echo factor $1$ \\
\bottomrule
\end{tabular}
\end{center}

$K = \Kpers$ is the reuse factor of the whole study, and no other is run.
\citet{choi2019} swept batch echoing only to $e = 2$
(Section~\ref{sec:related}), so four is the first factor beyond the range
they tested; every fresh-token saving reported below is $R/K$ at this $K$.

The training loop is unremarkable and reads as follows:
\begin{quote}\ttfamily
for step in range(total\_steps):\\
\hspace*{1em}if step \% K == 0: pool = next(loader)\hfill\# fresh data, once every K steps\\
\hspace*{1em}batch = draw(pool, mode)\\
\hspace*{1em}loss = model(batch).loss; loss.backward()\\
\hspace*{1em}opt.step(); opt.zero\_grad(set\_to\_none=True); sched.step()
\end{quote}
Only the peak learning rate is tuned, per $(B, K)$ cell. The schedule shape is
fixed: cosine with a warmup of \warmupfrac{} of the run's own planned steps and
a floor of \lrfloorfrac{} of peak, anchored to each run's token budget
(Section~\ref{sec:setup:workload}). Because the budget is in fresh tokens, the
$K = \Kpers$ arm anneals over $\Kpers\times$ the optimizer steps of the
baseline, so every cost-to-target reads the two arms at different points of
their schedules. Section~\ref{sec:results:e3} reports the control that puts
the arms on one schedule at $B = 512$. The rate is \emph{not} scaled by $k$:
the adaptive variant of the 2018 paper is deliberately outside the
confirmatory design, for the reason given in Section~\ref{sec:related}.

\paragraph{Why the third arm is a calibration arm, and not a control for locality.}
The $K=M$ \texttt{resample} arm was originally conceived as a locality control:
same amount of reuse, but drawn from a wider pool. An adversarial review of our
own harness argued, from the epoch-shuffled stream of
Section~\ref{sec:setup:workload}, that the arm is distributionally identical
to the baseline. The pool is a permutation of what the baseline would have
seen, so the sequence of gradients has the same law. The argument assumes
that the stream is exchangeable at the sequence level; we do not verify that
assumption, and the noise floor below is the arm's observed dispersion, not a
consequence of the identity. We re-designated it, before any confirmatory
run, as an \textbf{A/A null}: its ratio to the baseline has true value $1$
under that assumption, and its observed dispersion across seeds is the
empirical noise floor of every ratio we report. Even then the identity holds
only up to one quantization. An arm with $K > 1$ takes each evaluation, and
its last step, up to $K - 1$ optimizer steps away from the baseline's:
\aahorizonoffset{}\,\% of the run at $B = 32/128/512$
(Table~\ref{tab:crossing}), about one percent at $B = 512$. That is a factor
of five or more below the A/A dispersion of $\sigma_{\log} = \aasigmaFINAL$
measured in Section~\ref{sec:results:aa}.
As a consequence, stated here so that it cannot be reinterpreted later,
\textbf{this study makes no claim about locality}; the repetition-versus-shuffling
decomposition announced in the 2018-era design is named as future work in
Section~\ref{sec:conclusions}. In the code, the figures and the tables the arm
is labelled \emph{A/A calibration}, never \emph{locality control}.

A fourth, partial-echo arm at $M = K/2$ was part of the original design and was
dropped by construction: with $M=1$ a resampled pool is a permutation of the
repeated batch and the gradient is a mean over it, so at $K=2$ the arm would
duplicate the persistency arm at a quarter of the budget. It survives only where
$\lfloor K/2 \rfloor \ge 2$, and is reported---if at all---as secondary.

\subsection{Cost-to-target on four axes}
\label{sec:method:cost}

We compare quality at a fixed budget once only, in the epoch contrast of
Section~\ref{sec:results:e1}, where no arm reaches $\tau(B)$ at \eonebudget{}
Mtok of distinct data. Every other claim in this paper is of the form \emph{how much does it
cost to reach a fixed validation loss $\tau$}. Such a comparison does not
depend on where the budget ends; it does still read the
two arms at different points of their own cosine schedules, as
Section~\ref{sec:method:knobs} states. Four costs are logged at every
evaluation, and each of them defines a cost-to-target:
\begin{enumerate}
  \item[(i)] \textbf{optimizer steps} --- the scale-free axis, and the primary one;
  \item[(ii)] \textbf{fresh tokens} --- the data the pipeline had to produce;
  \item[(iii)] \textbf{wall-clock seconds} --- excluding evaluation time;
  \item[(iv)] \textbf{joules} --- GPU energy, excluding evaluation, with the
    provenance recorded alongside the number (Section~\ref{sec:energy}).
\end{enumerate}
Note that on the fresh-token axis the accounting is not a matter of taste: when
$K$ does not divide the step budget, the last pool is only partially consumed,
and a naive counter credits the persistency arm with data it never used. Our
harness counts fresh examples at the moment the pool is filled, and the
regression tests of the accounting module exist precisely to keep that honest.
Evaluation is excluded from the time and energy counters---an evaluation is
identical across arms by design, so including it would dilute every ratio toward
one.

\paragraph{The crossing rule.}
Reading a cost-to-target off a noisy curve requires a rule, and the rule must be
fixed in advance because it is exactly where a favorable reading can be
manufactured. Ours is: cost-to-target is obtained by linear interpolation at the
first evaluation index $i$ such that the loss is at or below $\tau$ at $i$
\emph{and} at $i+1$---two consecutive evaluations below target, with a run's
final evaluation counting as self-confirming. A single noisy dip therefore does
not end a run.

\paragraph{Learning rate, and how it enters.}
Within each (cell, seed), the reported cost is the one at the learning rate that
minimizes it, with the same number of learning rates swept for every arm in a
given $B$---enforced by the sweep expansion, not assumed. Cells whose winning
learning rate sits at the edge of the swept grid are flagged, and the treatment
of those flags is a pre-registered decision that we discuss in
Section~\ref{sec:setup}.

\subsection{The quantity under test}
\label{sec:method:hypothesis}

For each seed $s$ and batch size $B$ we form the paired step-cost ratio
\begin{equation}
  R(B, s) \;=\;
  \frac{\text{steps\_to\_target}(K{=}\Kpers, M{=}1;\, B, s)}
       {\text{steps\_to\_target}(K{=}1;\, B, s)},
  \label{eq:ratio}
\end{equation}
each arm at its own best learning rate, both at the pre-registered target
$\tau(B)$. The \emph{step penalty} is $g(B,s) = R(B,s) - 1 \ge 0$ up to noise.
Pairing is legitimate because the two arms of a seed share the data seed---hence
the identical order of fresh examples---the initialization seed, and the same
fixed evaluation set (Section~\ref{sec:setup:workload}).

Note that the fresh-token ratio is a deterministic transform of
\eqref{eq:ratio}, namely, $\text{fresh\_ratio} = R/K$ at $M=1$; the two axes
carry one test, not two, and we register $R$ as the primary scale-free quantity.
This also neutralizes a mechanical artifact: a target so easy that both arms
reach it at the first evaluation would produce $R \equiv 1$ and a spurious
$1/K$ data saving, which is why the target definition carries a floor check
(Section~\ref{sec:setup}).

The 2018 conjecture---repeated batches approximate fresh ones as the batch size
grows---becomes a statement about how $g$ moves with $B$:
\begin{itemize}
  \item[] \textbf{H1:} $g(B)$ decreases in $B$; specifically $g(512) < g(32)$.
  \item[] \textbf{H0:} $g(512) \ge g(32)$.
\end{itemize}
The test is a one-sided paired $t$-test at $\alpha = \alphalevel$ on the
per-seed difference $D(s) = \log g(32,s) - \log g(512,s)$, with an
untransformed fallback $D(s) = g(32,s) - g(512,s)$ applied to \emph{all} seeds
if any penalty is non-positive---the choice being made once, by that rule, and
not per seed. Seeds are the unit of replication. The decision rule, registered
verbatim, admits three outcomes: the conjecture is \emph{supported} if
$p < \alphalevel$ and the point estimate satisfies
$\hat g(512) \le \hat g(32)/2$; it is \emph{refuted} if the one-sided confidence
interval for $g(32) - g(512)$ excludes any halving; and anything else is
\emph{inconclusive}, reported as such with estimates and intervals and with no
claim in either direction. An inconclusive or refuting outcome is published with
the same prominence as a supporting one.

\section{Experimental apparatus}
\label{sec:setup}

We next describe the workload, the two-stage sweep, the pre-registration, and
the hardware---in that order, and with the level of detail that would allow the
study to be repeated on a free-tier account. That is where the confirmatory
study was in fact run, with the one exception Section~\ref{sec:setup:hardware}
records.

\subsection{Workload}
\label{sec:setup:workload}

The headline experiment, named \texttt{gap\_vs\_batch} in our code, trains a
decoder-only Transformer of the GPT family: \nlayer{} layers, \nhead{} heads,
width \nembd{} (about \nparams{} parameters), sequence length \seqlen{}, mixed
precision, AdamW with a cosine schedule. The harness requests bf16, but the T4
has no bf16 units, so \primaryontfour{} of the \nrunstotal{} runs of the
confirmatory study computed in fp16 autocast with dynamic loss scaling, at a
micro-batch of \microbatch{} sequences. Each run header records the
substitution. The \primaryonupscale{} runs of the group $(B = 512,
\text{seed } 7)$ ran on the Blackwell in bf16 at a micro-batch of
\bridgemicro{}, under the hardware rule of Section~\ref{sec:setup:hardware}.
So did the extensions, and the difference is declared wherever the two are
compared. The data are FineWeb-Edu \citep{penedo2024fineweb}, pre-tokenized
once and stored as \texttt{uint16} shards, as described below;
pre-tokenization is not a convenience but a requirement of the design, since
streaming from a remote hub would inject exactly the input-pipeline overhead
whose effect we are trying to measure separately.

\paragraph{Optimizer and schedule.}
AdamW runs with $\beta_1 = \betaone$, $\beta_2 = \betatwo$, weight decay
\weightdecay{} and gradient clipping at \gradclip{}. Only the peak learning
rate is tuned (Section~\ref{sec:setup:stages}); the schedule is a cosine over
the run's planned optimizer steps, with a linear warmup of \warmupfrac{} of
those steps and a floor of \lrfloorfrac{} of peak. The planned steps follow
from the fresh-token budget: $S_{\mathrm{base}}(B)$, the baseline's, is
\primXplannedbasea{}/\primXplannedbaseb{}/\primXplannedbasec{} at
$B = 32/128/512$, the A/A arm shares it, and the persistency arm plans
\primXplannedpersa{}/\primXplannedpersb{}/\primXplannedpersc{}, $\Kpers\times$
as many. The two arms of a pair therefore anneal over different horizons, and a
cost-to-target reads them at different fractions of their cosines
(Section~\ref{sec:results:e3}). The schedule is anchored to the planned count.
The steps executed, which Table~\ref{tab:crossing} prints, differ from it by
one or two where the fresh-token budget cuts the last pool: \primXstepsbasec{}
and \primXstepspersc{} at $B = 512$ against the planned \primXplannedbasec{}
and \primXplannedpersc{}. A fraction of a cosine is always a fraction of the
planned count.

\paragraph{Data and evaluation.}
The corpus is the \texttt{\fwconfig{}} configuration of FineWeb-Edu, tokenized
with the GPT-2 BPE by \texttt{scripts/prepare\_fineweb.py} into shards in the
\texttt{llm.c} format. There are \fwtrainshards{} training shards of
\fwshardMtok{} Mtok, \fwtrainGtok{} Gtok in all, and one validation shard
holding the corpus's first \fwvalMtok{} Mtok. The training shard set is fixed for the
whole study and recorded in every run's configuration; sequences are
contiguous windows of \seqlen{} tokens. For data
seed $s$ and epoch $e$ the stream is the permutation
\texttt{default\_rng([s, e])} of all sequences, consumed without replacement,
so both arms of a seed see the identical order of fresh examples. The
evaluation set is \evalseqs{} sequences of \seqlen{} tokens, drawn once without
replacement with seed \evalseed{} from the validation shard, which the
training stream never reads. It is the same set for every batch size, arm,
card and extension. A target, a floor check and every paired difference in this
paper are read on it, so its sampling error is common-mode and cancels to
first order in every ratio.

Each run has a fixed budget of \freshbudget{} million fresh tokens---identical
across arms within a cell---and is evaluated \nevals{} times, at fixed
fresh-token boundaries \evalstride{} Mtok apart. Note that the boundaries are
the same for all arms in a cell: an evaluation cadence tied to optimizer steps
would give the arms different numbers of evaluations and, with the crossing rule
of Section~\ref{sec:method:cost}, different effective resolutions. Batch sizes
are $B \in \{32, 128, 512\}$, the range over which the conjecture makes a
prediction. The micro-batch size is fixed per card---\microbatch{} sequences
on the T4, \bridgemicro{} on the Blackwell---with gradient accumulation
supplying the rest, so that the optimizer update does not depend on $B$.

The size of the model deserves a word, since it is smaller than we would have
liked. The original design used a 12-layer model of \dtwelveparams{}
parameters (\dtwelvemflop{} MFLOP per token) and a budget of \budgetbefore{}
Mtok per run. The pre-flight measurement put it at roughly \originalgpuh{}
GPU-hours against a free tier of 30 hours a week---not a study, a wish.
Amendment~1 rescaled it by four factors. The model went to \nlayer{} layers at
width \nembd{}, \dsixmflop{} MFLOP per token and \rescalefactor$\times$
cheaper. The budget went to \freshbudget{} Mtok (\budgetfactor$\times$), the
coarse grid from \lrcoarsebefore{} to \nlrcoarse{} rates
(\lrcoarsefactor$\times$), and the arms per cell from \armsbefore{} to three
by dropping the secondary partial-echo arm (\armsfactor$\times$). The
learning-rate factor acts on the coarse stage only---the final stage carried
\nlrfinal{} rates forward before and after the amendment---and the other
three on both stages. That is how \originalgpuh{} GPU-hours became
\preflightgpuh{}, the pre-flight estimate for $n = \nseedsprov$ seeds at about
\preflighttokps{} ktok/s. The executed campaign is counted in card-hours: the
wall time of training plus evaluation, summed over the runs of each card and
never across cards. It came to \hoursTfour{} on the T4 (\nrunsTfour{} runs),
\hoursLfour{} on the L4 (\nrunsLfour{} runs) and \hoursBlackwell{} on the
Blackwell (\nrunsBlackwell{} runs), \nrunsall{} runs in all.
Table~\ref{tab:platforms} gives the hours per stage and per card; the
confirmatory stage alone took \hoursfinalTfour{} on the T4
(\nrunsfinalTfour{} runs) and \hoursfinalBlackwell{} on the Blackwell
(\nrunsfinalBlackwell{} runs). We declare the consequence rather
than hide it: the study is budget-limited, it runs in a regime where a single
epoch of fresh data is never exhausted, and it does not speak to the scale at
which the reuse factors of \citet{choi2019} were measured.

\subsection{Two stages, and where the learning rate is decided}
\label{sec:setup:stages}

The sweep runs in two stages, both resumable---a practical necessity, as
Section~\ref{sec:setup:hardware} explains.

\begin{itemize}
  \item \textbf{Coarse:} \nlrcoarse{} learning rates per cell on a geometric
    grid from \lrlow{} to \lrhigh{}, one seed (seed~\coarseseed), a quarter of the token budget.
    Its purpose is twofold: to locate the learning rate, and to define the
    targets.
  \item \textbf{Final:} the best \nlrfinal{} learning rates per cell, carried
    forward from the coarse stage, at the full budget and with \nseeds{} seeds.
    This is the confirmatory stage; \nrunstotal{} runs in total.
\end{itemize}

The learning rate is tuned inside every $(B, \text{arm})$ cell, with the same
number of learning rates for every arm within a $B$. This is the non-negotiable
part of the design: it is what the 2019 objection asks for, and everything else
in the paper is conditional on it. Its resolution is stated as plainly: the
grid spacing is a factor \gridratio{}, and the final stage does not refine
it---it carries two grid points forward and reads the better one. We write
$\eta^\star(\text{arm})$ for the rate at which an arm reaches its target in
the fewest steps, and $\eta_1 := \eta^\star(K{=}1)$ for the baseline's. The
baseline's winner is the same grid point, $\eta_1 = \etaone$, at all three
batch sizes. Over a sixteenfold change of $B$ the tuned optimum does not move
by one grid step, so the grid locates it only to within that factor. A
by-product of the tuning is the ratio
\begin{equation}
  \rho(B) \;=\; \frac{\eta^\star(K{=}\Kpers)}{\eta^\star(K{=}1)},
  \label{eq:rho}
\end{equation}
which we report as a result in its own right (Section~\ref{sec:results}): if
reuse were nothing but a larger learning rate in disguise, $\rho$ would sit near
$1/K$; thresholds for reading it were fixed before the data were seen.

\subsection{Targets}
\label{sec:setup:targets}

The target $\tau(B)$ is computed \emph{from the coarse stage only, and from
baseline ($K=1$) runs only}---the frozen $\tau(B)$ is therefore exogenous to
every treatment arm. The re-adjusted target of
Section~\ref{sec:setup:prereg} is not, since its floor check reads the first
evaluation of all arms. Concretely, $\tau(B)$ is the running-minimum
validation loss reached by the best coarse baseline run at $75\%$ of the
coarse budget. A floor check is mandatory: every final-stage run in the cell
must still be above $\tau(B)$ at its first evaluation, with a margin
(registered as: first-eval running-min $> \tau(B) + 0.01$ nats). Where the
check fails, $\tau(B)$ is lowered by $0.02$ nats and the check is repeated;
every adjustment is reported. Table~\ref{tab:targets} gives the frozen values
beside the re-adjusted ones, with the steps the re-adjustment took. The
targets were written into the repository before the final stage started and
were never edited afterwards.

Two sensitivities are mandatory and are reported in
Section~\ref{sec:results}: all primary quantities are recomputed at
$\tau(B) \pm 0.02$ and $\pm 0.05$ nats, and the confirmatory claim stands only
if the decision is unchanged at $\pm 0.02$.

\subsection{Pre-registration and the paper trail}
\label{sec:setup:prereg}

The analysis plan---hypothesis, endpoint, target rule, estimators, exclusion
rules, censoring, multiplicity, and the stopping rule for the number of
seeds---was written before the confirmatory runs and registered on the OSF
Registries \citep{osf}. It is frozen in the source repository under a signed
tag, \texttt{prereg-v1} at commit \texttt{\preregvonehash{}}, and the three
amendments it carries are dated files rather than edits, each under a tag of
its own. The extensions have theirs: \texttt{prereg-v5} at
\texttt{\preregvfivehash{}} for E1 and E2, \texttt{prereg-v6} at
\texttt{\preregvsixhash{}} for E3; Section~\ref{sec:availability} lists every
tag with its hash.
Amendment~1 is the resizing described in Section~\ref{sec:setup:workload},
with the budget-limited regime declared. It also fixed the power contingency
in advance: seeds are added up to the cap of \nseeds{}, and ``if $\sigma_D >
\underpoweredthreshold$ the study is reported as inconclusive, not as a
negative result''. Amendment~2 adds $\rho(B)$ and a learning-rate-matched
arm, to answer the ``is it just a learning rate?'' objection with evidence
rather than with an argument. Amendment~3 is a rule for heterogeneous
hardware, described below. The registration declares our
foreknowledge honestly---at the time of registering, pilot data had been
observed, though none of the proposed analyses had been run---and the
registration, together with the two post-registration decision documents, is
anchored by OpenTimestamps, so that the order of events can be verified offline
by a referee who trusts neither us nor the OSF (Section~\ref{sec:availability}).

Two interim looks at the confirmatory data are on record, and we disclose them
here rather than in a footnote. The primary was read at $n = \nseedsprov{}$
seeds on 16 August 2026, before the seed count was raised to \nseeds{} under
the registered power rule; the addendum that raised it records the reading.
Then, on 4 September 2026, with \nrunssecondlook{} of the \nrunstotal{} runs
on disk---\nseedssecondlook{} seeds of \nseeds{}---the four cost axes and the
per-seed penalties were read in the course of writing this draft. That
reading is what motivated the two extensions recorded as dated additions to
the registration document: a multi-epoch arm, and a replication of the primary
with a step-cadence instrument and deeper targets, on new seeds. The remaining
\nrunsaftersecondlook{} runs executed after that reading. Neither look changed
a rule of the primary analysis, whose statistic is computed once, on the
closed set of \nrunstotal{} runs. Every figure and table of this paper is
drawn on closed run sets: the primary's on that one, each extension's on its
own. What the looks did change is disclosed where it
happened: the extensions were designed after, and because of, what the second
look showed, and their registration says so in as many words.

Two decisions were taken after registration, and both were written down and
timestamped \emph{before} the runs that they would affect were analyzed.

\paragraph{Learning rates at the edge of the grid.}
The plan excludes from the confirmatory set any cell whose winning learning rate
sits at the edge of the swept grid. Applied literally to the final stage---where
only \nlrfinal{} learning rates survive from the coarse stage, so that the
argmin is \emph{always} at an edge---the rule would exclude everything, while
flagging nothing about the tuning. We therefore read the rule as applying to the
coarse grid, where it is informative and where it in fact fires zero times out
of nine. The reading is recorded with its justification, and it is conditional
on that audit: if the coarse stage ever stops passing the check, the decision is
void.

\paragraph{Target re-adjustment.}
The plan prescribes a mechanical re-adjustment of $\tau$ once the final stage is
complete, since the floor check depends on the first-evaluation minimum over all
runs. The re-adjustment is applied by a command that writes a new file rather
than editing the frozen one. Which arm forces it is a matter of record. At
$B = 32$ every arm's first-evaluation minimum lies above the frozen target and
nothing moves. At $B = 128$ the minima are \firstEvalMinbaseb{},
\firstEvalMinpersb{} and \firstEvalMinaab{} nats for the baseline, the
persistency and the A/A arm; at $B = 512$ they are \firstEvalMinbasec{},
\firstEvalMinpersc{} and \firstEvalMinaac{}. Against the frozen targets of
Table~\ref{tab:targets} only the $K = \Kpers$ arm lies below, so the
adjustment at both batch sizes is forced by the treated arm alone. On the
baseline and A/A minima it would be zero, and $\hat g(512)$ would keep its
frozen-target value of \gcFINAL{}. We note explicitly that the adjustment
moves the ratio \emph{toward} the hypothesis. The rule of priority, stated
once: the primary test reads the frozen target, because it is the exogenous
one; the re-adjusted target is reported beside it, never in its place
(Section~\ref{sec:results:primary}, Table~\ref{tab:adjaxes}).

\paragraph{An anomaly, documented before it mattered.}
The evaluation of an arm with $K>1$ does not land on a pool boundary but $K-1$
steps inside the pool---at $K=\Kpers$, three steps in. Attributing the
evaluation to the end of the pool instead is a defensible alternative, and it
moves $R$ \emph{against} our hypothesis. It is registered as a sensitivity, never
as the primary reading, in a document dated before the run that would decide the
matter; the document has since been vindicated, the apparent anomaly having
turned out to be noise at two seeds.

\paragraph{Two extensions, registered before their runs.}
The plan was amended once more, in a version frozen before either extension
executed, to add the two experiments of
Sections~\ref{sec:results:e2} and~\ref{sec:results:e1}. The first is the
replication the ceiling argument calls for: the same three arms and three batch
sizes on \etwoseeds{} new seeds (\etwoseedrange{}, disjoint from the primary's),
with targets set on the full-length baselines, an evaluation grid in optimizer
steps rather than in fresh tokens, and a confirmation horizon in place of the
next-evaluation rule---the amendment states the horizon, the cadence, the
resulting evaluation counts as an identity, and the reason for each. The second
holds the data fixed and varies only the spacing of the repetitions, to
separate reuse from an early second epoch. Both were registered with their
endpoints, their tests and their seed counts; neither borrows alpha from the
primary, and the primary's own reading is not amended by either. The same
version fixed the hardware bridge of Section~\ref{sec:results:bridge}---its
runs, its statistic and its equivalence margin---before any bridge run existed.

\subsection{Hardware, and what it forces}
\label{sec:setup:hardware}

The confirmatory study runs on Kaggle notebooks, two NVIDIA T4 per session,
under a free quota of 30 GPU-hours per week with a session cap of about eleven
hours. This has three consequences worth stating, because they shaped the
design more than any scientific preference did.

First, \emph{everything must be resumable}. A run that cannot survive the death
of its session cannot be completed at all here, so the harness checkpoints and
restores optimizer, scheduler, data-iterator and RNG state exactly, and the
restore path is covered by tests that compare a resumed trajectory against an
uninterrupted one.

Second, \emph{the assignment unit is the $(B, \text{seed})$ group}.
Amendment~3 to the pre-registration, dated 8 August 2026 while the coarse stage
was running, added a second free tier---an NVIDIA L4 on Modal, measured at
\lenergyperMtok{} kJ per Mtok against the T4's \tenergyperMtok{}---and fixed
the rule for two chips. Every run that enters a paired ratio, both arms and
every rate compared within the cell, executes on the same chip. Absolute
seconds and joules are never pooled across machines and are reported per
card. A group found to straddle two cards is dropped from the time and energy
axes as censored. If the cards' ratios were to disagree beyond the A/A noise,
the secondary axes would be reported per machine with no pooled claim. The
allocation was deterministic and was written for the \ngroupsprov{} groups of
the provisional primary at $n = \nseedsprov$ seeds: dealt round-robin
\rrsplit{} over the L4 (machine~0) and the T4 (machine~1), matching the
measured throughput ratio of \lfourspeedratio{}. The campaign executed has
\ngroupsfinal{} groups, three batch sizes by \nseeds{} seeds. Two deviations
from that allocation are on record. (i)~No final-stage run executed on
machine~0: Modal's credit was spent before the final stage began, so the L4
carries \nrunsLfour{} coarse runs and nothing else. Both partitions of the
\ngroupsprov{} registered groups ran on the T4, as the addendum of 16 August
2026 states. (ii)~The university HPC facility became available on 5 September
2026 and gave the study the Blackwell. On 6 September 2026 the final stage
stood at \primaryontfour{} of \nrunstotal{} runs, all on the T4, and the
second interim look of 4 September had read \nrunssecondlook{} of them. The
one group not yet started, $(B = 512, \text{seed } 7)$, was assigned whole to
that facility, outside the registered partition, by a decision committed
before any of its runs. So \primaryontfour{} primary runs executed on the T4
and \primaryonupscale{} on an RTX~PRO~6000 (Blackwell). Every paired ratio of
that group has both legs on the same card and its absolute seconds and joules
are kept apart; the rule is what allows the time and energy axes to be read
at all. The two extensions ran entirely on the Blackwell. The
bridge of Section~\ref{sec:results:bridge} establishes that the baseline
arm's steps-to-target transfers between the two cards within
$\pm\bridgegatemargin{}$ in log, about $\pm\bridgegatepct\%$. The transfer of
the between-arm ratio itself is not measured, and every reading of an
extension beside the primary carries that qualification.

Third, \emph{the sweep is sharded rather than parallelized}. The two T4 of a
session are billed as one, so a session runs two shards of the same sweep, and
the campaign advances in sessions of a few hours over several weeks.

Table~\ref{tab:platforms} collects the three platforms, what ran on each, and
the software each ran under; Section~\ref{sec:energy} explains the energy
column.

\begin{table}[t]
\centering
\footnotesize
\setlength{\tabcolsep}{4pt}
\begin{tabular}{p{2.2cm}p{3.7cm}p{1.2cm}p{1.9cm}p{1.8cm}p{1.0cm}p{1.6cm}}
\toprule
card & what ran on it: stage, runs, card-hours & driver & torch (CUDA) & precision & micro-batch & energy method \\
\midrule
Kaggle T4 & coarse sweep, \nrunscoarseTfour{} runs, \hourscoarseTfour{}~h; confirmatory stage, \nrunsfinalTfour{} runs, \hoursfinalTfour{}~h; learning-rate-matched arm, \nrunslrmTfour{} runs, \hourslrmTfour{}~h & not recorded & \torchtfour{} (\cudatfour) & fp16 autocast, dynamic loss scaling & \microbatch{} & NVML total-energy counter \\
Modal L4 & coarse sweep, \nrunscoarseLfour{} runs, \hourscoarseLfour{}~h & not recorded & not recorded & bf16 & \microbatch{} & NVML total-energy counter \\
RTX~PRO~6000 Blackwell (university HPC) & confirmatory group $(B = 512, \text{seed } 7)$, \nrunsfinalBlackwell{} runs, \hoursfinalBlackwell{}~h; bridge, \nrunsbridgeBlackwell{} runs, \hoursbridgeBlackwell{}~h; E1 coarse pass, \nrunseonecoarseBlackwell{} runs, \hourseonecoarseBlackwell{}~h; E1, \nrunseoneBlackwell{} runs, \hourseoneBlackwell{}~h; E2, \nrunsetwoBlackwell{} runs, \hoursetwoBlackwell{}~h; E3, \ethreenruns{} runs, \ethreehours{}~h; E3-lr, \ethreelrnrunsarchive{} runs, \ethreelrhours{}~h & \driverblackwell{} & \torchblackwell{} (\cudablackwell) & bf16 & \bridgemicro{} & NVML total-energy counter \\
\bottomrule
\end{tabular}
\caption{Platforms and software. Card-hours are the wall time of training
plus evaluation, summed over the runs of one card within a stage and never
across cards; the per-card totals are in Section~\ref{sec:setup:workload}.
The CUDA version is the one of the torch build; the driver of the free-tier
cards is not recorded in the run headers. Micro-batch is the number of
sequences per forward/backward pass, gradient accumulation supplying the rest
of $B$.}
\label{tab:platforms}
\end{table}

The harness itself is a single Python package with \ntests{} passing tests; it
was subjected to an adversarial review whose brief was to find what would make
the numbers wrong without making anything crash, and several of the rules stated
in Section~\ref{sec:method}---the crossing rule, the equal-cadence evaluation,
the fresh-token accounting, the A/A re-designation---are its findings, applied
before any confirmatory run.

\section{Measuring the joules}
\label{sec:energy}

An energy claim is only as good as the attribution behind it, and attribution
turned out to be the hardest part of this study. In this section we state where
a saving could come from at all, how we measure energy, what we found when we
compared the available instruments, and which axis we consequently designate as
primary. We report the negative findings in full: two of the three measurement
routes we tried do not support the claim we wanted to make with them.

\subsection{Three channels, and only three}
\label{sec:energy:channels}

A persistent step costs the same FLOPs as a fresh one---the model does not know
that it has seen this batch before. Joules can therefore go down through three
channels, and it is worth naming them separately because they are defensible to
very different degrees.

\begin{enumerate}
  \item \textbf{A data-starved accelerator (input-bound training).} Where the
    pipeline cannot keep up, reuse converts stall time into gradient steps. This
    is real and documented for vision, video and recommendation workloads
    \citep{zhao2022ingestion, audibert2023tfdata}, and it is \emph{absent} in
    language-model pre-training on pre-tokenized tokens, where the loader is not
    the bottleneck. We say this plainly because the opposite claim would be the
    easiest one to make and the first one a referee would reject.
  \item \textbf{CPU and storage work avoided ($\div K$).} Reading, decoding and
    augmenting are performed once per pool instead of once per step. This is the
    most defensible energy channel and, to our knowledge, the one never
    quantified in joules in the literature: the host side of training is
    measured to consume a share
    of total infrastructure power comparable to the trainers themselves, and
    hyperscalers solve input-bound training by \emph{spending}---disaggregated
    preprocessing fleets---where reuse would solve it algorithmically.
  \item \textbf{Fewer total steps to target (an optimization effect).} This
    would hold in any regime, language-model pre-training included, but it has
    never been demonstrated, and the evidence of \citet{choi2019} points the
    other way: reuse buys fewer fresh examples at the price of \emph{more}
    accelerator steps. It is the scientific bet of the study, and the reason the
    step axis, not the joule axis, carries the primary's confirmatory test.
\end{enumerate}

Note that only channel~3 is a statement about optimization. Channels~1 and~2 are
statements about a pipeline, and a joule saving that came entirely from them
would be a systems result---worth reporting, but not evidence that reuse trains
better. We keep the two readings separate throughout
Section~\ref{sec:results}.

\subsection{The instrument, and its provenance rule}
\label{sec:energy:instrument}

Energy is read from the device, not modeled. Our harness records, for every run
and every evaluation boundary, the GPU energy consumed since the previous
boundary, together with a \emph{provenance} field naming the mechanism that
produced it. Three mechanisms are tried, in order: the NVML total-energy
counter, which integrates in hardware and is exact for the purpose; a fallback
that integrates sampled instantaneous power over the interval; and, failing
both, an explicit \texttt{None}. The rule the harness enforces is that
\textbf{there is no joule without a provenance}, and that a zero is never a
measurement---a regression test exists for precisely that failure mode, since a
silently zeroed energy counter is the one bug that would make our headline
number look excellent.

It is worth noting which cards have the hardware counter, because it is not the
distinction one expects. The total-energy counter is present on data-center
parts and absent on the workstation and consumer silicon that most academic
groups actually own, including the Ampere cards available to us; the Blackwell
server-edition cards of the university HPC turned out to have it, which we
learned only by asking the NVML API directly---a query through the
\texttt{nvidia-smi} field list had said otherwise. Where it is absent we fall
back to power sampling, which is less precise but remains \emph{per
device}---and per device is the property that matters, as the next subsection
shows.

\subsection{Why node-level power does not answer the question}
\label{sec:energy:ipmi}

The obvious alternative to a device counter is the node's own power supply,
readable over IPMI, which has the appeal of being a wall-plug measurement: it
sees the CPU, the memory, the fans and the losses that the GPU counter cannot.
We tested it on two clusters, and the outcome was negative in one case and
sobering in the other.

On a shared node---the normal condition on a busy cluster---IPMI measures the
\emph{node}, not the job. In our measurements it overstated the job's energy by
about $\upscaleover\times$, and even the differential reading (power under load
minus idle power) failed to recover the truth while other users' jobs occupied
the remaining cards. Exclusive allocation is the only escape, and on that
machine it is not grantable.

On a cluster where an exclusive allocation \emph{is} grantable, the reading
becomes attributable to the job---and is still a node reading. Over one such
run the node consumed $\ipminodekJ$~kJ while per-device NVML power sampling
reported $\nvmlgpukJ$~kJ for the same interval (these cards have no
counter), a factor of $\ipmiratio\times$ accounted for by seven idle cards,
the host CPU and the fans. The differential is more informative:
$+\ipmidelta$~W at the wall against $+\nvmldelta$~W at the device, leaving
roughly $\hostgap$~W of host-side power that is genuinely ours and that NVML
cannot see. Table~\ref{tab:instruments} collects the instruments, where each
was available to us, and what each turned out to measure.

\begin{table}[t]
\centering
\small
\begin{tabular}{p{3.1cm}p{3.3cm}p{2.6cm}p{5.2cm}}
\toprule
instrument & where & scope & what we found \\
\midrule
NVML total-energy counter & Kaggle T4 (the confirmatory study); university HPC (RTX~PRO~6000 Blackwell, server edition: one confirmatory group, the bridge, the extensions) & one card, integrated in hardware & present; the source of every joule in Section~\ref{sec:results}; no sampling, no coverage clause \\
NVML power sampling & departmental cluster (A40, RTX~3090) & one card, sampled & the only per-device instrument on these cards; used once, on the exclusive-node run of the last row, and no run of the confirmatory study or of an extension contributed a joule from it \\
IPMI, shared node & university HPC & whole node & overstates the job by about $\upscaleover\times$; the differential reading does not recover it while other jobs hold the remaining cards; exclusive allocation not grantable \\
IPMI, exclusive node & departmental cluster & whole node, attributable & $\ipminodekJ$~kJ at the node against $\nvmlgpukJ$~kJ sampled on the card over one run ($\ipmiratio\times$); differential $+\ipmidelta$~W against $+\nvmldelta$~W, i.e.\ about $\hostgap$~W of host-side power NVML cannot see; about sixteen hours of queueing per data point \\
\bottomrule
\end{tabular}
\caption{The energy instruments available to this study and what each
measures. The per-device counter is the only energy axis of the results; the
exclusive-node reading is one run on other hardware, used to size what the
counter cannot see; the shared-node reading is not an instrument for this
question at all.}
\label{tab:instruments}
\end{table}

\subsection{What we report}
\label{sec:energy:decision}

We therefore designate the \textbf{per-device counter as the primary energy
axis}. It is attributable by construction, it carries the same definition on
every platform we use, and it is the axis on which a paired ratio between two
arms of the same seed on the same card is meaningful. The node-level
measurement under exclusive allocation is \textbf{not an axis of any result}.
It exists as one run on a departmental cluster, on cards without the counter,
and it sizes what the device reading cannot see: about $\hostgap$~W of
host-side power on that node. No arm was compared to another on it, and no
wall-plug number appears in Section~\ref{sec:results}.

The limitation this leaves is stated without softening: every joule ratio in
this paper is a ratio of device energies, and host-side energy enters none of
them. What the omission can do to a ratio is bounded, not guessed. A run's
host energy is its host power times its seconds, so under any constant host
power the host-inclusive ratio lies between the device-joule ratio and the
seconds ratio. At the frozen targets both are above one at every batch size
($\Rjoulec$ and $\Rsecc$ at $B = 512$); at the replication's deeper target
$\tau'(512)$ both are below one ($\etwoRjoulec$ and $\etwoRsecc$). The sign of
every energy reading in Sections~\ref{sec:results:axes}
and~\ref{sec:results:secondary} therefore survives any constant host power;
only its size moves, within those bounds. Combining device and host into a
single wall-plug number would require a host-power model we are not in a
position to defend.

\section{Results}
\label{sec:results}

Every stage of the study is complete. The run census is \nrunsall{} runs: the
coarse stage (\nrunscoarse{}), the confirmatory stage (\nrunstotal{}), the
learning-rate-matched arm (\lrmnruns{}), the bridge across cards (\bridgen{}),
the replication (\etwonruns{}) and the epoch contrast (\eonenruns{}, with a
coarse stage of \eonencoarse{} runs of its own). Its cost in card-hours, per
card and per stage, is given once, in Section~\ref{sec:setup:workload}.

Three confirmatory tests are registered, and we count them here once. The
primary test of Section~\ref{sec:results:primary} is one-sided, on the step
axis, on seeds $0$--$7$. The replication of Section~\ref{sec:results:e2} is
the same one-sided test of the same hypothesis on seeds \etwoseedrange{}, with
the instrument corrected. The epoch contrast of Section~\ref{sec:results:e1}
is two-sided, on a different hypothesis, on runs of its own. Each was computed
once, on its closed run set, and each spends $\alpha = \alphalevel{}$. The
registration fixed in advance how the first two are read together: the primary
is the registered protocol's \emph{verdict}, the replication supplies the
paper's \emph{estimate} of $g(B)$, and neither is corrected against the other.
They are two chances at $\alpha = \alphalevel{}$ on one hypothesis, so the
verdict on the 2018 conjecture is to be read at a family-wise error rate below
$\fwerlevel$ over those two tests. That guarantee is carried by the
replication, whose pairs are uncensored; the primary's $p$ is read as an index
of separation, for the reason Section~\ref{sec:results:primary} gives. The
epoch contrast tests a different hypothesis
on disjoint runs and is not corrected against them, as the registration fixes
in advance. Everything else in this section is estimation with intervals or
descriptive, and spends no $\alpha$. Extension~E3, registered on 7~September
2026 as a sensitivity analysis of the replication's estimate at $B = 512$, has
a placeholder in Section~\ref{sec:results:e3}.

\subsection{Is it just a larger learning rate?}
\label{sec:results:rho}

We begin with the objection that has to be cleared before any other result can
be read. Table~\ref{tab:rho} reports $\rho(B)$ of \eqref{eq:rho}, the ratio of
the best learning rate under reuse to the best learning rate of the baseline,
measured on the completed coarse stage. Two symbols recur below.
$\eta^\star(K)$ is the coarse-stage optimum of a cell: the minimizer of best
validation loss over $\log\eta$, obtained by three-point parabolic
interpolation around the grid argmin, as Amendment~2 registered it.
$\eta_1(B)$ is the baseline's top-ranked rate, the one the final stage carries
forward (Section~\ref{sec:setup:stages}). The coarse stage is one seed at a
quarter of the budget over \nlrcoarse{} rates whose spacing is a factor
\rhogridfactor{}; the plan classes $\rho$ as descriptive and spends no
$\alpha$ on it, and a single-seed estimate carries no interval.

\begin{table}[t]
\centering
\footnotesize
\begin{tabular}{lccl}
\toprule
$B$ & $\rho(B)$ & grid argmin & reading (thresholds fixed before the data) \\
\midrule
$32$  & $\rhotwoa$ & $\rhogrida$ & $\ge 0.8$: not explained by a larger learning rate \\
$128$ & $\rhotwob$ & $\rhogridb$ & $\ge 0.8$: not explained by a larger learning rate \\
$512$ & $\rhotwoc$ & $\rhogridc$ & between $0.5$ and $0.8$: indeterminate \\
\midrule
$32$/$128$/$512$, E2 & $\etworhoprimea$/$\etworhoprimeb$/$\etworhoprimec$ & --- & two-point grid at $\tau'$, majority of \etwoseeds{} seeds (Section~\ref{sec:results:e2}) \\
$512$, E3-lr & $\ethreelrRho$ & --- & extension card, $\tau'$, one seed, argmins at $\eta_1$ (Section~\ref{sec:results:e3}) \\
\bottomrule
\end{tabular}
\caption{The tuned learning rate under reuse, relative to the tuned baseline---%
the higher the further from a mere re-parametrization. Coarse stage: one seed,
a quarter of the budget, \nlrcoarse{} rates on a geometric grid of spacing
\rhogridfactor{}, read by best validation loss; $\rho$ is the ratio of the
interpolated optima $\eta^\star$, the grid column the ratio of the raw argmins,
a sensitivity. Descriptive, single seed, no interval. Were persistency nothing
but a larger step size, $\rho$ would sit near $1/K = 0.25$. The E2 row is
$\rho'(B)$ on the replication's inherited two-point grid: the majority best
rate of the persistency arm over that of the baseline, resolved to a factor
\rhogridfactor{}, a different estimator on a different card and depth
(Section~\ref{sec:results:e2}). The last row is the reading of Extension~E3
(Section~\ref{sec:results:e3}) on the replication's card at its deeper
target.}
\label{tab:rho}
\end{table}

It turns out that at $B = 32$ and $B = 128$ the optimal learning rate
\emph{rises} with $K$, which is the opposite of the $\eta^\star/K$ behavior that
would explain the effect away. At $B = 512$ the value falls in the band we
pre-declared as indeterminate, and we report it as such: the datum neither
supports nor refutes the re-parametrization reading at that batch size. On the
raw grid argmins the reading is the same at $B = 32$ and $B = 128$ and more
favorable at $B = 512$, where both arms share the argmin. The grid itself
deserves a sentence. Its spacing is a factor \rhogridfactor{}, and the final
stage carries forward the best \nlrfinal{} rates without refining between
them. The baseline's argmin is the second grid point, $\eta_1 = \etaone$, at
all three batch sizes: over a sixteen-fold range of $B$ the tuned baseline
rate does not move by one grid point, and a shift smaller than a factor three
is below what the sweep resolves. Figure~\ref{fig:lr} is the evidence at that
resolution.

Two further facts bear on the same objection, and we collect them here so
that the reader does not have to assemble them from
Section~\ref{sec:results:secondary}. The learning-rate-matched arm that
Amendment~2 added to answer it turned out degenerate: the treatment's
cost-minimizing rate among the final-stage rates is the baseline's own
$\eta_1$ at every batch size, so the arm re-executes the tuned configuration
and measures replication noise, not the value of re-tuning. The control that
would close the question at $B = 512$ is the baseline at a larger step size,
$K\eta_1$. Amendment~2 registers it at one seed from the coarse grid, at the
nearest grid point (a ratio of \rhogridfactor{}, not \Kpers{}), as a
descriptive reading; only its upgrade to three seeds is unregistered. That
reading is in Figure~\ref{fig:lr}: at $B = 512$ the baseline's best coarse
validation loss is \coarseBaseAtEtac{} at $\eta_1$ and \coarseBaseAtThreeEtac{}
at the next grid point (\coarseBaseAtEtab{} against \coarseBaseAtThreeEtab{}
at $B = 128$, \coarseBaseAtEtaa{} against \coarseBaseAtThreeEtaa{} at
$B = 32$), one seed, a quarter of the budget. At that resolution the baseline
at the larger step ends worse, not better. Extension~E3 reads the same
four-rate grid at $B = 512$ on the replication's card
(Section~\ref{sec:results:e3}). The replication's inherited two-point grid
already gives a second reading, the E2 row of Table~\ref{tab:rho}:
$\rho'(B) = \etworhoprimeb$ and \etworhoprimec{} at $B = 128$ and $B = 512$,
and \etworhoprimea{} at $B = 32$, where the arms part
(Section~\ref{sec:results:e2}). The coarse $\rho(32) = \rhotwoa$ and that
$\rho'(32)$ are not the same reading: the two estimators differ in card,
depth, criterion and resolution, and Section~\ref{sec:results:e2} says at
what resolution the second is read. The objection is therefore answered at
$B = 32$ and $B = 128$ on the registered coarse reading, with the
replication's coarser reading pointing the other way at $B = 32$. At
$B = 512$ it is left open, $\rho$ falling in the band registered as
indeterminate, and every claim in the paper is to be read with that
qualification attached.

\subsection{The primary endpoint}
\label{sec:results:primary}

\begin{table}[t]
\centering
\begin{tabular}{lccc}
\toprule
$B$ & $\tau(B)$ & adjusted $\tau(B)$ & floor-check adjustments \\
\midrule
$32$  & $\taua$ & $\tauadja$ & none \\
$128$ & $\taub$ & $\tauadjb$ & $\tauadjnb$ \\
$512$ & $\tauc$ & $\tauadjc$ & $\tauadjnc$ \\
\bottomrule
\end{tabular}
\caption{Targets, in nats of validation loss, frozen before the confirmatory
stage; the adjusted column is the mechanical re-adjustment of
Section~\ref{sec:setup:prereg}, computed once on the closed confirmatory stage
onto a new file, the last column counting the adjustment steps it took. At
$B = 128$ and $B = 512$ the adjustment is forced by the persistency arm alone,
whose first-evaluation minima are \firstEvalMinpersb{} and \firstEvalMinpersc{}
nats against \firstEvalMinbaseb{} and \firstEvalMinbasec{} for the baseline and
\firstEvalMinaab{} and \firstEvalMinaac{} for the A/A arm; on the baseline and
A/A arms alone no adjustment would fire. The frozen $\tau(B)$ is exogenous to
every treatment arm and carries the registered verdict; the adjusted target is
not, and is read beside it.}
\label{tab:targets}
\end{table}

\begin{figure}[t]
\centering
\includegraphics[width=0.8\textwidth]{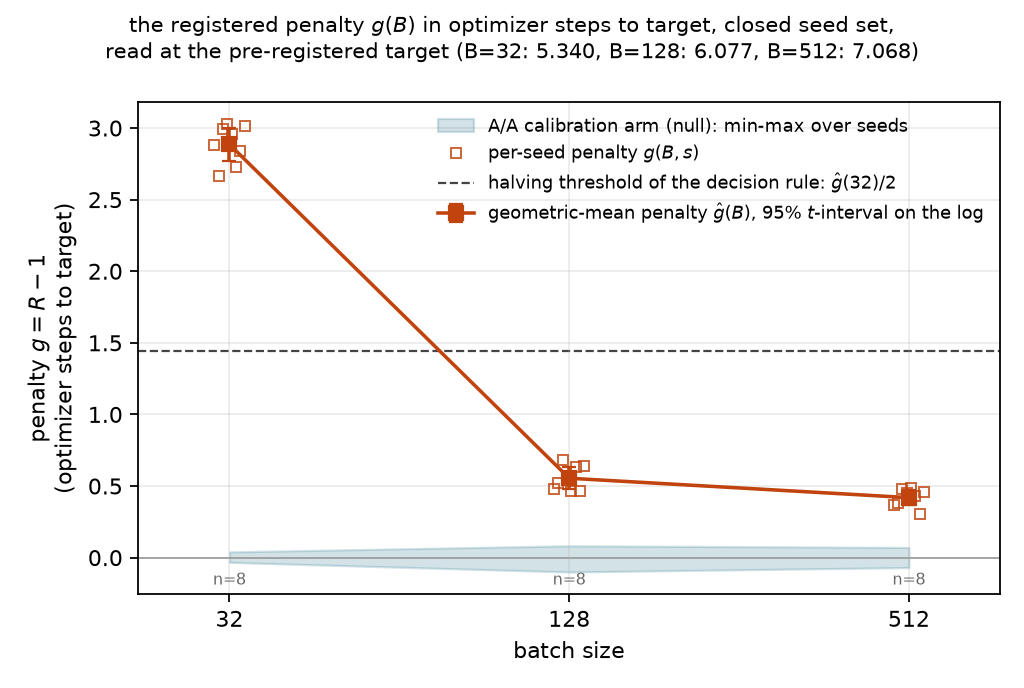}
\caption{The registered quantity: the penalty $\hat g(B)$ against batch size,
per seed, with the geometric mean and its 95\% interval, the halving threshold
of the decision rule, and the min-max band of the A/A arm's per-seed
penalties. Drawn once, on the
closed seed set, as the plan required. At $B = 128$ and $B = 512$ every seed's
persistency cost is that of the arm's first evaluation, so the penalties drawn
there are ceilings (Section~\ref{sec:results:primary}); the replication of
Section~\ref{sec:results:e2} removes the ceiling, and Figure~\ref{fig:rtau}
follows the ratio to deeper targets.}
\label{fig:gap}
\end{figure}

The confirmatory stage closed with all \nrunstotal{} runs on disk, none
excluded under the instrumentation rule, and the test below was computed once,
on that set. At the frozen targets of Table~\ref{tab:targets}, with the edge
policy of Section~\ref{sec:setup:prereg}, the penalties are
$\hat g(32) = \gaFINAL$ against $\hat g(512) = \gcFINAL$, a halving threshold
of $\halvingFINAL$. Every penalty is positive, so the rule's logarithmic branch
applies: the paired one-sided test on $D(s) = \log g(32,s) - \log g(512,s)$
gives $t = \tstatFINAL$ on $\dfstatFINAL$ degrees of freedom,
$p = \pvalueFINAL$, with an upper confidence bound of $\ciupperFINAL$ on
$g(32) - g(512)$, which the registration writes on the untransformed scale
because its refuting branch is stated there. Figure~\ref{fig:gap} draws the
per-seed penalties behind these numbers. The $p$ is to be read as an index of
separation rather than a guaranteed error rate: the persistency leg of the
pair at $B = 512$ is constant across seeds, for the reason given below, and
$\sigma_D$ and $t$ inherit that. One reading does not depend on it: every one
of the \nseeds{} seeds has $D(s) > 0$, a sign test at $p = \primSignP$,
descriptive and not registered. The other uses the same $\sigma_D$: on the
scale of the test the observed $\bar D = \primDmean$ has a one-sided lower
95\% bound of \primDlb{} against a halving threshold of $\ln 2 = \lntwo$. Under the
registered decision rule the outcome is \textbf{conjecture supported}, and it
is unchanged at $\tau \pm 0.02$ and at $\tau \pm 0.05$ nats, all four
perturbations. Under the pool-end attribution of
Section~\ref{sec:setup:prereg} the step ratios are \poolendFINAL{} at the
three batch sizes and the decision is unchanged.

At the re-adjusted targets of Table~\ref{tab:targets} the same rule takes its
other branch. There $\hat g(512) = \gcadjFINAL$ is below zero, so every seed
is read on the untransformed difference $g(32,s) - g(512,s)$, the branch the
replication of Section~\ref{sec:results:e2} also takes:
$\hat g(32) = \gaadjFINAL$, $t = \adjT$, $\sigma_D = \adjSigmaD$ in raw units,
$p = \pvalueadjFINAL$, with \adjNbelowOne{} of \nseeds{} seeds at
$R(512) < 1$. The verdict is unchanged at the re-adjusted targets $\pm 0.02$
and $\pm 0.05$ nats, where $\hat g(512)$ is \adjgmfive{}, \adjgmtwo{},
\adjgptwo{} and \adjgpfive{} at $-0.05$, $-0.02$, $+0.02$ and $+0.05$ nats. Section~\ref{sec:results:secondary} reads the
four cost axes at those targets (Table~\ref{tab:adjaxes}).

One pair of the test straddles the two cards of Section~\ref{sec:setup:hardware}:
seed~$7$ has its $B = 32$ leg on the T4 and its $B = 512$ leg on the
Blackwell, and the bridge of Section~\ref{sec:results:bridge} does not reach
that seed. On the \loon{} seeds whose legs are all on the T4,
$\hat g(32) = \looga$, $\hat g(512) = \loogc$, $t = \loot$, $p = \loopval$ and
$\sigma_D = \loosigmaD$: the verdict is unchanged. The card bias the bridge
measures at $B = 512$, $r_{\mathrm{hw}} = \bridgerc$, acts on the denominator
of that one pair and in the direction of the hypothesis.

The seed count deserves a paragraph of its own, because the registration
anticipated this case and named it. $\sigma_D$ is the standard deviation over
seeds of the paired difference $D(s)$ of the test, not the dispersion of the
A/A arm, which is two to five times smaller
(Section~\ref{sec:results:aa}). On the closed set $\sigma_D = \sigmaDFINAL$ in
log units, and the registered power rule turns that into \seedsneededFINAL{}
seeds, more than the cap of \nseeds{} the plan fixed in advance. Amendment~1
reads: if $\sigma_D > \underpoweredthreshold{}$ the study is reported as
\emph{inconclusive, not as a negative result}. We report the clause as
written, and what it protects against: a false negative. The outcome here is a
rejection at $p = \pvalueFINAL$, and the observed $\bar D$ sits at
\primDmean{} against a minimum detectable $|\Delta\log|$ of \primMDE{} at
$n = \nseeds{}$ (registered: \primMDEreg{}), so the shortfall in power does
not touch the decision. What it touches is the estimate of $g(512)$, which
carries more noise than the plan budgeted for, and more seeds are not
available under the registered cap. Three further things belong here. The
rule and its threshold are written in log units; on the untransformed scale of
the registered confidence bound, at the frozen targets, $\sigma_D = \primSigmaDraw$
and the rule asks for \primSeedsRaw{} seeds, which rounds to the cap (the
\adjSigmaD{} above is the same quantity at the re-adjusted targets). The seed count was raised
from \nseedsprov{} to \nseeds{} after the August reading at
$\sigma_D = \sigmaD$, below the threshold, under the branch of Amendment~1
that adds seeds up to the cap. That decision was taken on the primary pair and
applied at every batch size, where the amendment names the A/A arm and
$B = 512$ only: a deviation from its letter, reported as one. And the $\alpha$ of a test whose $n$
was re-estimated on a variance read from the same data is approximate. The
replication's $\sigma_D$ is \etwoSigmaDlog{} in log units, under the
threshold; the registration makes its estimate the paper's headline for a
reason of the instrument (Section~\ref{sec:results:e2}), not of power.

A property of the instrument has to be read together with these numbers.
$\tau(B)$ was fixed from the coarse stage, whose runs are a quarter of the
length of the confirmatory ones, and on the full-length runs the persistency
arm at $B = 128$ and $B = 512$ is already below $\tau(B)$ at its first
evaluation, in every seed (Appendix~\ref{app:curves} and
Table~\ref{tab:crossing}). Under the crossing rule its cost to target there is
the cost of that first evaluation, uninterpolated, so $R(512)$---and with it
$\hat g(512)$---is a \emph{ceiling}: the penalty at $B = 512$ is at most the
value printed. The registered verdict stands a fortiori, since a smaller
$g(512)$ can only widen the gap the test asks for; the \emph{estimate} does
not, and neither do $\sigma_D$ and the $t$ statistic, both computed with one
leg of the pair nearly constant across seeds. Figure~\ref{fig:rtau} in
Appendix~\ref{app:rtau} shows how the ratio depends on the depth of the target
over the whole range the curves resolve, and the remedy is registered rather
than improvised: a replication of the primary on new seeds, with an evaluation
grid in optimizer steps that samples every arm at the same resolution and with
targets set on the full-length baselines, pre-registered before any of its
runs and reported in its own right. Section~\ref{sec:results:e2} is that
replication, and it removes the ceiling: no seed there crosses at its first
evaluation.

The addendum obliges us to record what was read before this. The same test on
the first $\nseedsprov{}$ seeds, in August, returned
$\hat g(32) = \ga$, $\hat g(512) = \gc$ and
$p = \pvalue$---the interim look that produced the decision to extend
the stage to \nseeds{} seeds, and whose disclosure is the reason the addendum
exists. A second look, on 4~September 2026 with \nrunssecondlook{} of the
\nrunstotal{} runs on disk (\nseedssecondlook{} seeds of \nseeds{}), returned
$\hat g(32) = \gasecondlook$, $\hat g(512) = \gcsecondlook$ and
$\sigma_D = \sigmaDsecondlook$; the two extensions were registered after it
and because of it (Section~\ref{sec:setup:prereg}). The remaining
\nrunsaftersecondlook{} runs, seeds $6$ and $7$, executed after that look, and
the test above was computed once, on the closed set. Every figure and table of
this paper is drawn on closed run sets.

\subsection{The calibration arm, and what it licenses}
\label{sec:results:aa}

The A/A arm returned step ratios of $\aaratiosFINAL$ at the three batch sizes,
each with a confidence interval containing one---the apparatus is not
manufacturing an effect, and the registered harness-integrity gate passes on
the axis we read it on (Section~\ref{sec:results:secondary} records that
reading). Note that this is what licenses the reading of the previous
subsection: a penalty of $\gaFINAL$ at $B = 32$ is not a marginal excursion
above a noise floor whose width we would otherwise be guessing at, but roughly
two orders of magnitude larger than the dispersion the null arm exhibits, whose
logarithmic standard deviations are $\aasigmaFINAL$.

The null arm's dispersions are not the $\sigma_D$ of the power rule. Applied
to the $\sigma_{\log}$ just quoted, the rule $n \approx \powerrule\,\sigma^2$
returns fewer than the registered floor of \nseedsfloor{} seeds at every batch
size. The $\sigma_D$ that drove the seed count is that of the primary
contrast, $D(s)$ of Section~\ref{sec:results:primary}. It stood at \sigmaD{}
at $n = \nseedsprov{}$ in August, from which the rule returned \nseeds{}
seeds, and at \sigmaDFINAL{} on the closed set, where the same rule asks for
\seedsneededFINAL{}. The addendum concedes plainly that the August arithmetic
was done after the primary had been read at $n = \nseedsprov{}$; the estimate
of the dispersion was itself the noisiest thing in that reading, as a sample
of three seeds is apt to be.

\subsection{The bridge across cards}
\label{sec:results:bridge}

The two registered extensions ran on a different card from the primary: the
RTX~PRO~6000 (Blackwell) of a university HPC facility, in bf16 and with a
micro-batch of \bridgemicro{} sequences against the T4's \microbatch{}. So did
one $(B, \text{seed})$ group of the primary itself
(Section~\ref{sec:setup:hardware}). Amendment~3 keeps both legs of every
paired ratio on one card; it does not, by itself, say that a ratio measured on
one card means the same as a ratio measured on the other. The registration
therefore added a bridge: the primary's baseline arm, at its tuned rate and on
the primary's own seeds, re-executed on the new card---\bridgen{} runs,
configuration-identical to the runs they are paired with. For each seed,
$r_{\mathrm{hw}}$ is the ratio of steps to target on the new card to steps to
target on the T4, and the gate is an equivalence test fixed before any bridge
run existed: $|\,\overline{\log r_{\mathrm{hw}}}\,| < \bridgegatemean{}$, with
the 95\% interval inside $\pm\bridgegatemargin{}$ in log steps. A seed whose
primary group ran on the new card is excluded by the registered rule, which is
why the bridge carries \bridgepairs{} pairs at $B = 32$, $128$ and $512$.

The gate passes at every batch size. The binding reading is at the deeper
$\tau'(B)$ of the replication, where the registration requires the gate to
pass before any sentence reads the replication beside the primary:
$r_{\mathrm{hw}} = \bridgeprimera{}$, \bridgeprimerb{} and \bridgeprimerc{}
at $B = 32$, $128$ and $512$, with intervals on the log of \bridgeprimecia{},
\bridgeprimecib{} and \bridgeprimecic{}. At the frozen $\tau(B)$ it passes as
well, $r_{\mathrm{hw}} = \bridgera{}$, \bridgerb{} and \bridgerc{} with
intervals \bridgecia{}, \bridgecib{} and \bridgecic{}, but that reading is
coarse by quantization. At $B = 512$ the baseline crosses $\tau$ at steps
\primXstepbaseloc--\primXstepbasehic{} of \primXplannedbasec{}, and the
\nevals{} evaluations of the primary's grid space those steps by more than a
third of that distance (Table~\ref{tab:crossing}). Two entries deserve a word.
At $B = 128$ the interval excludes zero at both targets: the new card needs
\bridgerb{} times the T4's steps to reach $\tau$ and \bridgeprimerb{} times to
reach $\tau'$, a bias of the size a change of numerical precision can produce,
and one the margin was set to absorb. And the widest interval is the one at
$B = 512$ and $\tau'$: inside the margin, and the sentences of
Section~\ref{sec:results:e2} that read the replication beside the primary rest
on it.

What the bridge establishes is that the \emph{baseline} arm's steps to target
transfer across the two cards within the registered margin of
$\pm\bridgegatemargin{}$ in log steps. The transfer of the between-arm ratio
itself is not measured: no persistency run exists on both cards. The
persistency arm computes in fp16 with loss scaling on one card and in bf16 on
the other, at micro-batch \microbatch{} against \bridgemicro{}, taking four
updates on one batch. An interaction between arm and card of the size the
margin allows is of the order of the replication's effect at $B = 512$, the
logarithm of \etwoRstepc{}. The replication is internally valid on its own
card without the bridge; what the bridge licenses is reading its ratios beside
the primary's, within that margin. The gate is defined on steps only, and a
ratio of seconds is not among the quantities it says transfer; the absolute
seconds and joules of Section~\ref{sec:results:secondary} are reported per
card and never averaged across the two. One pair of the primary test straddles
the cards, seed~$7$ at $B = 32$ on the T4 against $B = 512$ on the Blackwell,
a seed the bridge does not reach; Section~\ref{sec:results:primary} gives the
reading without it.

\subsection{The replication with the corrected instrument}
\label{sec:results:e2}

The remedy promised in Section~\ref{sec:results:primary} was registered before
any of its runs, at commit \texttt{\preregvfivehash{}} of the registration
(prereg-v5), and is reported here in its own right. It repeats the primary
comparison on \etwoseeds{} fresh seeds (\etwoseedrange{}) and changes three
things. Two are properties of the instrument. Evaluation runs on a grid of
optimizer steps, every \etwocadence{} steps at $B = 32$, $128$ and $512$,
identical for all three arms within a batch size, so the persistency arm is no
longer sampled at a quarter of the baseline's resolution and the $K - 1$ step
offset of Section~\ref{sec:setup:prereg} is removed by construction rather
than corrected for. And every run executed on one model of card, the Blackwell
of Section~\ref{sec:results:bridge}, so that no ratio has its two legs on
different cards. The third is the depth of the target, on which the sign at
$B = 512$ depends (Appendix~\ref{app:rtau}). Targets are set by the registered
rule on the primary's own full-length baselines---the \etwontauruns{} final
baseline runs of seeds $0$--$5$, both rates, on the T4, aggregated as the
minimum over them---at a fraction \etwofresh{} of the fresh budget. That puts
$\tau'(B) = \etwotaua{}$, \etwotaub{} and \etwotauc{} nats, deeper everywhere
than the frozen $\tau(B)$ of Table~\ref{tab:targets}. Those runs had been read
when the replication was registered, and the registration says so: on the
\nrunssecondlook{} primary runs then on disk, $f = \etwofresh{}$ predicted
$R(32) = \etwopredRa$, $R(128) = \etwopredRb$, $R(512) = \etwopredRc$ and a
paired joule ratio of \etwopredJc{} at $B = 512$. The fraction was fixed by
reachability: \etwofresh{} is the deepest of the five registered fractions at
which all three arms reach $\tau'$ in every observed seed at every $B$, and at
$f = \fmapfe$ the persistency arm at $B = 32$ reaches $\tau'(32)$ in none of
the observed seeds.
Table~\ref{tab:fmap} prints the registered map from $f$ to $\tau'(512)$ and to
the predicted $R(512)$ beside what the replication measured at each of the
five depths; the estimate at $B = 512$ is conditional on $f$ in exactly that
sense.

\begin{table}[t]
\centering
\footnotesize
\begin{tabular}{lcccccc}
\toprule
$f$ & $\tau'(512)$ (nats) & registered $R(512)$ & replication $R(512)$ & 95\% interval & $n$ & joules \\
\midrule
\fmapfa & \fmaptaua & \fmappreda & \etwofmapRa & $[\etwofmapRloa,\,\etwofmapRhia]$ & \etwofmapna & \etwofmapJa \\
\fmapfb & \fmaptaub & \fmappredb & \etwofmapRb & $[\etwofmapRlob,\,\etwofmapRhib]$ & \etwofmapnb & \etwofmapJb \\
\fmapfc & \fmaptauc & \fmappredc & \etwofmapRc & $[\etwofmapRloc,\,\etwofmapRhic]$ & \etwofmapnc & \etwofmapJc \\
\textbf{\fmapfd} & \textbf{\fmaptaud} & \textbf{\fmappredd} & \textbf{\etwofmapRd} & $[\etwofmapRlod,\,\etwofmapRhid]$ & \etwofmapnd & \textbf{\etwofmapJd} \\
\fmapfe & \fmaptaue & \fmapprede & \etwofmapRe & $[\etwofmapRloe,\,\etwofmapRhie]$ & \etwofmapne & \etwofmapJe \\
\bottomrule
\end{tabular}
\caption{The registered map from the target fraction $f$ to the depth
$\tau'(512)$ and to the mean paired step ratio $R(512)$ that the
\nrunssecondlook{} primary runs then on disk predicted at each depth (Extension
E2 of the registration), beside the replication's own reading at the same five
depths: geometric mean over its seeds of the persistency-to-baseline steps to
target, the 95\% $t$-interval, the pairs entering, and the joule ratio. The
registered fraction is $f = \fmapfd$ (bold), the depth at which every arm
reaches the target in every observed seed; the other four rows are
descriptive.}
\label{tab:fmap}
\end{table}

The evaluation counts were registered as an identity rather than as numbers
with a tolerance, and they hold exactly: \etwonevalbase{} evaluation records
per baseline run at the three batch sizes, \etwonevalpers{} per persistency run
and \etwonevalaa{} per A/A run. On a finer grid the confirmation rule of
Section~\ref{sec:method} loses its meaning---two evaluations are twelve
optimizer steps apart at $B = 512$ on the primary's grid but two on this
one---so the replication confirms a crossing over a \emph{horizon} instead: the
first evaluation at or below $\tau'$ that is still at or below it at the first
evaluation at least $\lceil S_{\mathrm{base}}(B)/16 \rceil$ steps later,
$S_{\mathrm{base}}(B)$ being the baseline's planned optimizer steps at the
fresh budget (the executed counts are in Table~\ref{tab:crossinge2}), which is
\etwohorizon{} steps, the
primary's own spacing. The reading under the
unmodified two-evaluation rule is reported beside it below.

The registered verdict is \textbf{supported}: $\hat g(32) = \etwoga{}$ against
$\hat g(512) = \etwogc{}$, a halving threshold of \etwohalving{},
$t = \etwot{}$ on $\etwodf{}$ degrees of freedom, $p = \etwop{}$, with the
one-sided upper interval on $\hat g(32) - \hat g(512)$ at \etwociupper{}. On
the untransformed scale of the test the observed $\bar D = \etwoDmean$ has a
one-sided lower 95\% bound of \etwoDlb{} against the halving threshold of
\etwohalving{}. \etwoNpositive{} of \etwoseeds{} seeds have $D(s) > 0$, a
sign test at $p = \etwoSignP$, descriptive. The primary's own runs at
$B = 512$, seven pairs on the T4 and one on the Blackwell, corroborate the
sign within the card. Read at this $\tau'(512)$ with the estimator of
Figure~\ref{fig:rtau}, they give \primTauPrimeRstep{}
$[\primTauPrimeRsteplo,\,\primTauPrimeRstephi]$ on steps and
\primTauPrimeRjoule{} $[\primTauPrimeRjoulelo,\,\primTauPrimeRjoulehi]$ on
joules, $n = \primTauPrimeRnstep$. The verdict
is unchanged at $\tau' \pm 0.02$ and $\tau' \pm 0.05$ nats, the minor
sensitivities; the binding one the registration names is
$\tau' \pm \etwotauband$ nats. At $\tau' + \etwotauband$, the shallower side,
the step ratios are \etwoRpmpa{} $[\etwoRpmploa,\,\etwoRpmphia]$, \etwoRpmpb{}
$[\etwoRpmplob,\,\etwoRpmphib]$ and \etwoRpmpc{} $[\etwoRpmploc,\,\etwoRpmphic]$
on \etwoRpmpna{}, \etwoRpmpnb{} and \etwoRpmpnc{} pairs,
$\hat g(32) = \etwogpmpa$, $\hat g(512) = \etwogpmpc$, $p = \etwoppmp$:
supported. At $\tau' - \etwotauband$, the deeper side, the persistency arm at
$B = 32$ reaches the target in no seed, so the registered test is not
evaluable there; $B = 512$ keeps \etwoRpmmnc{} pairs of \etwoseeds{}, with
$R(512) = \etwoRpmmc$ $[\etwoRpmmloc,\,\etwoRpmmhic]$ and
$\hat g(512) = \etwogpmmc$, and $B = 128$ gives \etwoRpmmb{}
$[\etwoRpmmlob,\,\etwoRpmmhib]$. Under the two-evaluation rule the same runs
give $\hat g(512) = \etwogctwoeval{}$ and $p = \etwoptwoeval{}$: the decision
does not turn on the confirmation rule, a point worth making explicitly
because the rule was changed after the primary had been read. The calibration
arm is the registered harness gate on the replication's own runs, read on
steps with the fresh-token ratio beside it, as the registration prescribes. On
steps it returns \etwoAAstepa{} $[\etwoAAsteploa,\,\etwoAAstephia]$,
\etwoAAstepb{} $[\etwoAAsteplob,\,\etwoAAstephib]$ and \etwoAAstepc{}
$[\etwoAAsteploc,\,\etwoAAstephic]$; on fresh tokens \etwoAAfresha{}
$[\etwoAAfreshloa,\,\etwoAAfreshhia]$, \etwoAAfreshb{}
$[\etwoAAfreshlob,\,\etwoAAfreshhib]$ and \etwoAAfreshc{}
$[\etwoAAfreshloc,\,\etwoAAfreshhic]$. Every interval covers one on both axes:
the gate passes, and the fresh-token excursion of the primary's null arm at
$B = 512$ (Section~\ref{sec:results:secondary}) is absent on this grid. The
null arm's dispersions are $\sigma_{\log} = \etwoaasigma{}$. The primary
contrast's $\sigma_D$ is \etwosigmaD{} in raw units and \etwoSigmaDlog{} in
log units. The registered power rule, written in log units, returns
\etwoSeedsLog{}, so \etwoSeedsLogCeil{} seeds would have sufficed; on the raw
scale of this test's branch it returns \etwoSeedsRaw{}, which rounds up to
\etwoseedsneeded{}. Under the pool-end attribution of Section~\ref{sec:setup:prereg} the
ratios move to \etwopoolend{}, against the hypothesis at $B = 512$ and still
short of overturning it.

\begin{figure}[t]
\centering
\includegraphics[width=\textwidth]{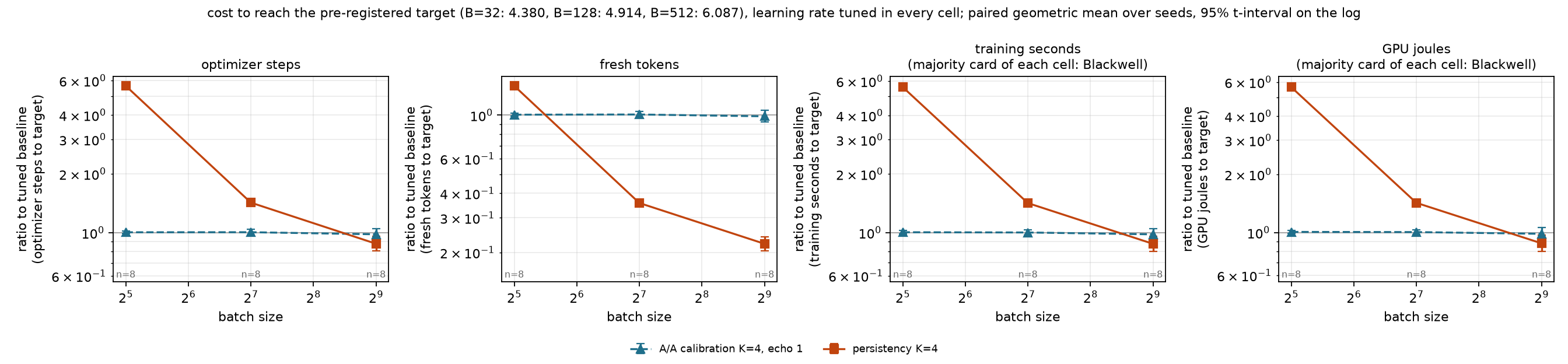}
\caption{The replication's four cost axes: cost to reach the deeper target
$\tau'(B)$, relative to the tuned baseline, against batch size---below one
means reuse is cheaper. Each panel is one cost axis
(optimizer steps, fresh tokens, training seconds, GPU joules). Each point is
the estimator of Table~\ref{tab:e2axes}: the paired geometric mean over seeds
of the per-seed ratio, with its 95\% $t$-interval on the logarithm,
$n = \etwoseeds$ pairs in every cell. Seconds and joules are all on the
Blackwell, so no pair is left out on any axis. No cell is a ceiling: no seed
crosses at its first evaluation (Table~\ref{tab:crossinge2}), so every marker
is filled. The A/A calibration arm (blue, dashed) sits on the line of no
effect on every axis, fresh tokens included, since the evaluation grid of this
study removes the offset of Section~\ref{sec:results:secondary}.}
\label{fig:ratioe2}
\end{figure}

What the corrected grid changes is the \emph{estimate}, and it changes it in
the direction the ceiling argument predicted. No seed now crosses its target at
its first evaluation, in any of the nine cells
(Table~\ref{tab:crossinge2}), against every one of \nseeds{} seeds at $B = 128$
and at $B = 512$ in the primary: the ceiling was an artifact of a coarse grid and a
target fixed on quarter-length runs, and it was hiding a penalty at $B = 512$
that is not merely small but slightly \emph{negative}. Table~\ref{tab:e2axes}
gives every cell on the four axes with the null arm beside it, the twin of
Table~\ref{tab:wins} at $\tau'$, and Figure~\ref{fig:ratioe2} draws it. At
$B = 512$ the persistency arm reaches the
target with \etwoTstepc{} of the baseline's steps and compute
($[\etwoTsteploc,\,\etwoTstephic]$), \etwoTsecc{} of its seconds
($[\etwoTsecloc,\,\etwoTsechic]$) and \etwoTjoulec{} of its joules
($[\etwoTjouleloc,\,\etwoTjoulehic]$), on \etwoTfreshc{} of its fresh tokens.
That is below one on all four axes, each interval entirely below one, and the
null arm's intervals cover one on each of them (\etwoAAjoulec{}
$[\etwoAAjouleloc,\,\etwoAAjoulehic]$ on joules). The objection this answers is
the one Appendix~\ref{app:rtau} raises, that the sign of the energy ratio at
$B = 512$ depends on the depth of the target: at $\tau'(512)$ it is below one.
At $B = 32$ and $B = 128$ it is not, at any depth the curves resolve. There
the persistency arm needs \etwoTstepb{} and \etwoTstepa{} times the steps,
\etwoTjouleb{} and \etwoTjoulea{} times the joules and \etwoTsecb{} and
\etwoTseca{} times the seconds, the three axes agreeing within one percent in
every cell (the per-step residuals of Section~\ref{sec:results:secondary}).
Two cells move against reuse between the two readings. At $B = 32$ the step
ratio grows from \Rstepa{} at $\tau$ to \etwoTstepa{} at $\tau'$, a cell
without a ceiling in either study (Appendix~\ref{app:rtau} follows it between
the two depths). And the fresh-token ratio at $B = 32$,
\Rfresha{} at $\tau$ with an interval covering one, is \etwoTfresha{}
$[\etwoTfreshloa,\,\etwoTfreshhia]$ at $\tau'$ (Appendix~\ref{app:rtau}): at
the deeper target reuse consumes more fresh data than the baseline there, not
less. Reuse buys
data at $B = 128$ and $B = 512$ and not at $B = 32$; it buys energy at
$B = 512$ only, and at the deeper target only.

\begin{table}[t]
\centering
\footnotesize
\begin{tabular}{llcccc}
\toprule
arm & $B$ & fresh tokens & steps (= compute) & seconds & joules \\
\midrule
persistency & $32$
  & $\etwoTfresha^{\,\blacktriangle}$ & $\etwoTstepa^{\,\blacktriangle}$ & $\etwoTseca^{\,\blacktriangle}$ & $\etwoTjoulea^{\,\blacktriangle}$ \\
 & & {\scriptsize$[\etwoTfreshloa,\,\etwoTfreshhia]$} & {\scriptsize$[\etwoTsteploa,\,\etwoTstephia]$}
   & {\scriptsize$[\etwoTsecloa,\,\etwoTsechia]$} & {\scriptsize$[\etwoTjouleloa,\,\etwoTjoulehia]$} \\
persistency & $128$
  & $\etwoTfreshb^{\,\blacktriangledown}$ & $\etwoTstepb^{\,\blacktriangle}$ & $\etwoTsecb^{\,\blacktriangle}$ & $\etwoTjouleb^{\,\blacktriangle}$ \\
 & & {\scriptsize$[\etwoTfreshlob,\,\etwoTfreshhib]$} & {\scriptsize$[\etwoTsteplob,\,\etwoTstephib]$}
   & {\scriptsize$[\etwoTseclob,\,\etwoTsechib]$} & {\scriptsize$[\etwoTjoulelob,\,\etwoTjoulehib]$} \\
persistency & $512$
  & $\etwoTfreshc^{\,\blacktriangledown}$ & $\etwoTstepc^{\,\blacktriangledown}$ & $\etwoTsecc^{\,\blacktriangledown}$ & $\etwoTjoulec^{\,\blacktriangledown}$ \\
 & & {\scriptsize$[\etwoTfreshloc,\,\etwoTfreshhic]$} & {\scriptsize$[\etwoTsteploc,\,\etwoTstephic]$}
   & {\scriptsize$[\etwoTsecloc,\,\etwoTsechic]$} & {\scriptsize$[\etwoTjouleloc,\,\etwoTjoulehic]$} \\
\midrule
A/A null & $32$
  & $\etwoAAfresha$ & $\etwoAAstepa$ & $\etwoAAseca$ & $\etwoAAjoulea$ \\
 & & {\scriptsize$[\etwoAAfreshloa,\,\etwoAAfreshhia]$} & {\scriptsize$[\etwoAAsteploa,\,\etwoAAstephia]$}
   & {\scriptsize$[\etwoAAsecloa,\,\etwoAAsechia]$} & {\scriptsize$[\etwoAAjouleloa,\,\etwoAAjoulehia]$} \\
A/A null & $128$
  & $\etwoAAfreshb$ & $\etwoAAstepb$ & $\etwoAAsecb$ & $\etwoAAjouleb$ \\
 & & {\scriptsize$[\etwoAAfreshlob,\,\etwoAAfreshhib]$} & {\scriptsize$[\etwoAAsteplob,\,\etwoAAstephib]$}
   & {\scriptsize$[\etwoAAseclob,\,\etwoAAsechib]$} & {\scriptsize$[\etwoAAjoulelob,\,\etwoAAjoulehib]$} \\
A/A null & $512$
  & $\etwoAAfreshc$ & $\etwoAAstepc$ & $\etwoAAsecc$ & $\etwoAAjoulec$ \\
 & & {\scriptsize$[\etwoAAfreshloc,\,\etwoAAfreshhic]$} & {\scriptsize$[\etwoAAsteploc,\,\etwoAAstephic]$}
   & {\scriptsize$[\etwoAAsecloc,\,\etwoAAsechic]$} & {\scriptsize$[\etwoAAjouleloc,\,\etwoAAjoulehic]$} \\
\bottomrule
\end{tabular}
\caption{The replication's twin of Table~\ref{tab:wins}: cost-to-target
ratios of the persistency arm and of the A/A calibration arm to the tuned
baseline at $\tau'(B)$, on the Blackwell, seeds \etwoseedrange{}: geometric
mean over \etwoseeds{} pairs with the 95\% $t$-interval beneath. Below one
favors reuse; $\blacktriangledown$ marks an outright win (interval entirely
below one), $\blacktriangle$ an outright loss (interval entirely above one),
no mark an interval covering one. No cell is a ceiling: no seed crosses at its
first evaluation (Table~\ref{tab:crossinge2}). Descriptive; the replication's
one confirmatory test is on the step axis.}
\label{tab:e2axes}
\end{table}

\paragraph{The learning rates, inherited.}
The replication ran no coarse stage of its own. Its two rates per
$(B, \text{arm})$ are the primary's final-stage pair, selected by the coarse
stage on the T4---one seed, a quarter of the budget, fp16 at micro-batch
\microbatch{}, read by best validation loss at a shallower depth---and carried
forward without re-location. The choice between the two is made per (cell,
seed) as in the primary, with the same number of candidates per arm, so the
parity of tuning effort of Section~\ref{sec:setup:stages} holds. Which rate
wins is the replication's own reading of the re-parametrization question,
quantized to a two-point grid. At $B = 512$ the cost-minimizing rate is
$\eta_1$ in \etwoWinEtabasec{} of \etwoseeds{} baseline seeds and
\etwoWinEtapersc{} of \etwoseeds{} persistency seeds, so the headline is
measured at equal rates, as in the primary
(Section~\ref{sec:results:secondary}); at $B = 128$ in \etwoWinEtabaseb{} and
\etwoWinEtapersb{}. At $B = 32$ the arms part: the baseline and the null arm
take the grid point above $\eta_1$ (\rhogridfactor{} times it) in every seed,
and the persistency arm takes $\eta_1$ in every seed---the smaller step at the
deeper target, in the cell where reuse loses on every axis. Written as the
ratio of Table~\ref{tab:rho}, the majority best rate of the persistency arm
over that of the baseline, $\rho'(B) = \etworhoprimea$, \etworhoprimeb{} and
\etworhoprimec{} at $B = 32$, $128$ and $512$, resolved to a factor
\rhogridfactor{} and descriptive. At $B = 32$ that value lies in the band the
registration reads as a re-parametrization, where the coarse
$\rho(32) = \rhotwoa$ reads the opposite. The two are not the same reading:
the estimators differ in card, depth, criterion and resolution, and a
two-point grid cannot place an optimum between its points. Whether the
inherited pair brackets the optimum on this card is not checked by the bridge,
which reads the baseline at fixed $\eta_1$ only; Extension~E3 reads the full
coarse grid at $B = 512$ on this card (Section~\ref{sec:results:e3}), and the
estimate at $B = 512$ is read with that condition attached: on that grid
both argmins fall on $\eta_1$, inside the pair, at one seed, so the condition
holds on the reading available.

\paragraph{Where on its schedule each arm is read.}
Every arm's cosine spans its own run, so at a common fresh budget the $K = 4$
arm anneals over four times the baseline's optimizer steps:
\etwoXplannedpersc{} planned steps against \etwoXplannedbasec{} at $B = 512$.
The executed counts of Table~\ref{tab:crossinge2} differ from the planned
ones by one or two steps, the cut at the fresh-token budget; the cosine
fractions below are on the planned denominator. The two legs of $R$ are
therefore read at different points of their schedules. At $B = 512$ the baseline crosses
$\tau'$ at steps \etwoXstepbaseloc--\etwoXstepbasehic{} of
\etwoXplannedbasec{}, \etwoXfracbaseloc--\etwoXfracbasehic\% of the way
through its cosine at \etwoXlrbaseloc--\etwoXlrbasehic{} of the peak rate,
while the persistency arm crosses at steps
\etwoXsteppersloc--\etwoXsteppershic{} of \etwoXplannedpersc{},
\etwoXfracpersloc--\etwoXfracpershic\% through at \etwoXlrpersloc{} of the
peak. The position reverses with $B$: at $B = 32$ the persistency arm crosses
at \etwoXfracpersloa--\etwoXfracpershia\% of its cosine and the baseline at
\etwoXfracbaseloa--\etwoXfracbasehia\%, and at $B = 128$ at
\etwoXfracperslob--\etwoXfracpershib\% against
\etwoXfracbaselob--\etwoXfracbasehib\%. The A/A arm shares the baseline's
horizon (\etwoXfracaaloc--\etwoXfracaahic\% at $B = 512$) and does not
calibrate this. In the primary, at the frozen targets, both arms cross early
(\primXfracbaseloc--\primXfracbasehic\% and \primXfracpersloc\% at
$B = 512$). The sign at $B = 512$ thus admits a second reading, position on
the schedule rather than reuse, and nothing in the replication's own data
separates the two. The registered control is Extension~E3
(Section~\ref{sec:results:e3}): on one schedule in steps for all three arms,
$R_{\mathrm{sched}}(512) = \ethreeRstep{}$ $[\ethreeRsteplo, \ethreeRstephi]$
on steps and \ethreeRjoule{} $[\ethreeRjoulelo, \ethreeRjoulehi]$ on joules,
against a null arm at \ethreeAAstep{}. The interval contains one, the
registered outcome (ii): at $B = 512$ the effect of reuse and the position on
the schedule are not separable at $n = \etwoseeds$, and every number of this
subsection at $B = 512$ carries that qualification; the abstract and
Section~\ref{sec:conclusions} carry the same sentence.

\paragraph{Provenance of the replication's seconds and joules.}
Every one of the \etwonruns{} runs reports
\texttt{gpu\_method = nvml\_total\_energy\_counter}, the hardware counter of
Section~\ref{sec:energy:instrument}, on an RTX~PRO~6000 with a power limit of
\blackwellcap{}~W; no run fell back to sampling, so the coverage clause was
never invoked. Each job was submitted for one card (\etwogres{}) and each run
header records a single visible device, so one job reads the counter of its
device. The other cards of the node can be held by other jobs, and the counter
reads the device, not the process: exclusivity is by allocation of the card,
and no per-run utilization certificate is in the archive. The registration
reads exclusivity from a per-run utilization record, and a run not certified
exclusive leaves the energy reading. Here the certificate is reconstructed
from the allocation, one card per job and one visible device, a deviation
from the plan's letter reported as one; the replication's energy claim is
read with that qualification. The second,
independent reading of the joule ratio at $B = 512$ that the registration
names is not recorded in the archive either; the joule ratios above rest on
the counter alone. Table~\ref{tab:e2abs} gives the absolute costs per arm on
this card, eight seeds per cell, never pooled with the T4 absolutes of
Section~\ref{sec:results:secondary}. Mean power sits at
\etwoAbsWbasec--\etwoAbsWpersc{}~W at $B = 512$, well under the cap, and
agrees across arms within a few watts in every cell: on this card too a step
costs what a step costs, and joules follow steps.

\begin{table}[t]
\centering
\footnotesize
\begin{tabular}{llrrr}
\toprule
$B$ & arm & seconds & kJ & mean W \\
\midrule
$32$  & baseline    & \etwoAbsSecbasea & \etwoAbsKJbasea & \etwoAbsWbasea \\
$32$  & persistency & \etwoAbsSecpersa & \etwoAbsKJpersa & \etwoAbsWpersa \\
$32$  & A/A         & \etwoAbsSecaaa   & \etwoAbsKJaaa   & \etwoAbsWaaa \\
$128$ & baseline    & \etwoAbsSecbaseb & \etwoAbsKJbaseb & \etwoAbsWbaseb \\
$128$ & persistency & \etwoAbsSecpersb & \etwoAbsKJpersb & \etwoAbsWpersb \\
$128$ & A/A         & \etwoAbsSecaab   & \etwoAbsKJaab   & \etwoAbsWaab \\
$512$ & baseline    & \etwoAbsSecbasec & \etwoAbsKJbasec & \etwoAbsWbasec \\
$512$ & persistency & \etwoAbsSecpersc & \etwoAbsKJpersc & \etwoAbsWpersc \\
$512$ & A/A         & \etwoAbsSecaac   & \etwoAbsKJaac   & \etwoAbsWaac \\
\bottomrule
\end{tabular}
\caption{Absolute costs to $\tau'(B)$ in the replication, on the RTX~PRO~6000
(power limit \blackwellcap{}~W, NVML total-energy counter): mean over
\etwoAbsNbasec{} seeds per cell of training seconds and GPU kilojoules at the
arm's cost-minimizing rate, and mean power as kilojoules over seconds. Never
pooled with the T4 absolutes of Section~\ref{sec:results:secondary}.}
\label{tab:e2abs}
\end{table}

\subsection{The schedule-position control at \texorpdfstring{$B = 512$}{B = 512}}
\label{sec:results:e3}

Extension~E3 was registered on 7~September 2026, after the primary, the epoch
contrast, the replication and the bridge had been read in full, as a
sensitivity analysis of the replication's estimate at $B = 512$: it adds
nothing to the family of confirmatory tests, changes no verdict, and is
reported beside $\hat g(512)$ wherever that number is quoted. Its runs were
executed the same evening on the replication's card; the registered text
wrote the ratio inverted, and an erratum tagged before any run was read
restates it as persistency over baseline, the plan's convention, in which the
reading rule below had been written.

\emph{Design.} The replication's three arms at $B = 512$ only, on one
schedule in steps for all three: \ethreesteps{} optimizer steps, the
persistency arm's planned length, with a warmup of \ethreewarmup{} steps and
the same cosine. The persistency arm is the replication's configuration with
its stopping rule restated in steps; the baseline and the null arm run
\ethreesteps{} steps, hence \ethreefresh{}~Mtok of fresh tokens, four times
the budget, and at every step the three arms sit on the same point of the same
cosine. Rates, seeds (\etwoseedrange{}), evaluation cadence, card, precision
and micro-batch are the replication's; \ethreenruns{} runs. The estimand is
$R_{\mathrm{sched}}(512)$, the geometric mean over seeds of the persistency
arm's steps to $\tau'(512)$ over the baseline's, with a 95\% $t$-interval on
the log, together with the same ratio on seconds and on joules and the null
arm's ratio as the floor. The fresh-token axis is not read, being four by
construction. A baseline that has not crossed $\tau'$ by step \ethreesteps{}
is censored and its pair dropped; with fewer than \nseedsfloor{} pairs the
extension reports inconclusive by censoring.

\emph{Reading rule, registered.} If the interval of $R_{\mathrm{sched}}(512)$
lies entirely below one, the sign of the replication's estimate at $B = 512$
is attributed to reuse and not to schedule position. If it contains one, the
paper states, in the abstract, here and in Section~\ref{sec:conclusions}, that
at $B = 512$ the effect of reuse and the position on the schedule are not
separable at $n = \etwoseeds{}$, and quotes the estimate with that
qualification. If it lies entirely above one, the sign at $B = 512$ is
attributed to schedule position and the compute and energy claim at $B = 512$
is withdrawn, the replication's numbers staying in
Section~\ref{sec:results:e2} with this reading beside them. In every case the
replication's numbers are reported unchanged.

\emph{Result.} All \ethreeRstepn{} pairs survive: no baseline is censored.
On the shared cosine the baseline crosses $\tau'(512)$ at steps
\ethreeXbaselo--\ethreeXbasehi{}, the persistency arm at
\ethreeXperslo--\ethreeXpershi{} and the null arm at
\ethreeXaalo--\ethreeXaahi{}, all between \ethreeXfracbaselo{} and
\ethreeXfracpershi\% of the schedule at \ethreeXlrperslo--\ethreeXlrbasehi{}
of the peak rate, and every arm at $\eta_1$ in \ethreeWinEtabase{} of
\ethreeRstepn{} seeds. $R_{\mathrm{sched}}(512) = \ethreeRstep{}$
$[\ethreeRsteplo, \ethreeRstephi]$ on steps, \ethreeRsec{} $[\ethreeRseclo,
\ethreeRsechi]$ on seconds and \ethreeRjoule{} $[\ethreeRjoulelo,
\ethreeRjoulehi]$ on joules; the null arm returns \ethreeAAstep{}
$[\ethreeAAsteplo, \ethreeAAstephi]$ on steps. The interval contains one:
this is the registered outcome (ii). At $B = 512$ the effect of reuse and
the position on the schedule are not separable at $n = \etwoseeds$, and the
replication's estimate there---\etwoRstepc{} on steps, \etwoRjoulec{} on
joules---is quoted with that qualification. Read at the same point of the
same schedule, four-fold reuse reaches $\tau'(512)$ in the same number of
optimizer steps as fresh data, within the interval, on a quarter of the
fresh tokens; the replication's \etwoRstepc{} was read with the baseline at
\etwoXfracbaseloc--\etwoXfracbasehic\% of its own anneal against a
treatment at \etwoXfracpersloc--\etwoXfracpershic\% of its, and the
control does not say which of the two readings a third schedule would
give. Only the peak rate is tuned here, as everywhere in the paper; the
longer cosine is a different schedule for the baseline, not a re-tuned one.

\emph{The rate grid on the extension's card (E3-lr).} The replication's
configuration at $B = 512$, seed \ethreeseed{}, all three arms, at all
\nlrcoarse{} rates of the primary's coarse grid at the full budget:
\ethreelrnruns{} runs, two of which per arm repeat the replication's own
seed-\ethreeseed{} runs. It reads, per arm, the steps to $\tau'(512)$ at each
rate and its argmin $\eta^\star(\text{arm})$, the ratio
$\rho'(512) = \eta^\star(K = 4)/\eta^\star(K = 1)$ under Amendment~2's
thresholds, and whether each argmin lies inside the inherited pair; one seed,
no test, no interval. The registered consequence has two branches. If either
argmin lies outside the pair, the replication's estimate at $B = 512$ is
reported as conditional on the rate transfer, and the seed-\ethreeseed{}
$R(512)$ at the argmin rates is printed beside it as an exploratory number.
If both lie inside, the paper states that the pair brackets the optimum on
this card as well. $\rho'(512)$
enters the last row of Table~\ref{tab:rho}.

\emph{Result.} On the extension's card both argmins lie inside the inherited
pair. The baseline reaches $\tau'(512)$ in \ethreelrStepsbaseB{} steps at
$\eta_1$ against \ethreelrStepsbaseA{} at the grid point below it, and does
not reach it at all at the two points above, where its best loss over the
full budget stays at \ethreelrBestbaseC{} and \ethreelrBestbaseD{} nats. The
persistency arm takes \ethreelrStepspersB{} steps at $\eta_1$ against
\ethreelrStepspersA{}, \ethreelrStepspersC{} and \ethreelrStepspersD{} at the
other three rates, and the null arm behaves as the baseline
(\ethreelrStepsaaB{} steps at $\eta_1$, no crossing above it). Hence
$\eta^\star = \eta_1$ for every arm and $\rho'(512) = \ethreelrRho$, in the
band Amendment~2 reads as not a re-parametrised rate, at one seed, on this
card, at this depth; it enters Table~\ref{tab:rho} beside the coarse value
and not in its place. The pair brackets the optimum on this card as well,
which is the second branch of the registered consequence. The baseline at
three times $\eta_1$, the leg Section~\ref{sec:results:rho} could cite only
from the coarse stage, here does not reach $\tau'(512)$ within its budget:
the direction opposite to the one a larger-rate reading of the $B = 512$
result would need.

\subsection{The four cost axes, read together}
\label{sec:results:axes}

Table~\ref{tab:wins} is the paper's own conjecture put in numbers, one cell
per axis and batch size, and it carries two readings that must not be merged.

On the \emph{data} axis, reuse wins at $B = 128$ and $B = 512$: there the
persistency arm reaches the target having consumed fewer than half the fresh
tokens of the tuned baseline, and the ratio falls as $B$ grows. At $B = 32$
the fresh-token interval covers one at the frozen target and lies above one at
the deeper one (Table~\ref{tab:e2axes}). Note that at $M = 1$ this is $R/K$
and hence the primary quantity re-expressed, not a second piece of evidence:
it is drawn here descriptively, the registered reading being the confirmatory
one of Section~\ref{sec:results:primary}, on the closed \nrunstotal{}-run set.
On the \emph{step}, \emph{compute} and \emph{joule} axes the ratio stays above
one at every batch size \emph{at the frozen targets}, and the three axes lie
nearly on top of each other. Observe that this coincidence is not a
coincidence at all: on a pre-tokenized text pipeline the accelerator is never
starved, so a step costs what a step costs and joules are very nearly
proportional to steps. Channel~2 of Section~\ref{sec:energy:channels} has
nothing to skip here by construction; channel~1 is bounded by measurement
rather than switched off: per step the two arms agree in seconds and in joules
within \primPSmaxdevpct\% on the point estimates, and the null arm shows a
residual of the same size, with intervals, in
Section~\ref{sec:results:secondary}. What remains is channel~3, and at the
frozen targets it does not pay.

We state the consequence in the plainest terms we can, because it is the result
a reader is most likely to want softened: \textbf{at the frozen targets
$\tau(B)$, and at those targets only, minibatch persistency costs more GPU
energy to reach the target than the tuned baseline at every batch size
measured}. At the deeper $\tau'(512)$ of the replication the joule ratio is
\etwoTjoulec{} $[\etwoTjouleloc,\,\etwoTjoulehic]$, below one; at $B = 32$
and $B = 128$ no depth the curves resolve brings it below one
(Table~\ref{tab:e2axes}, Appendix~\ref{app:rtau}). The reading at $\tau(B)$
carries two qualifications. At $B = 128$ and $B = 512$ the persistency arm is
already below the frozen target at its first evaluation, so its cost there is
a ceiling (Appendix~\ref{app:curves}); and at the re-adjusted targets of
Table~\ref{tab:targets} the same penalty is $\gcadjFINAL$ on steps, with the
joule interval at $B = 512$ covering one (Table~\ref{tab:adjaxes}). The
energy verdict is a statement about a target, not about a regime. The energy
case for reuse rests on the input-bound regime, where the pipeline work that
reuse skips is real work; this study bounds it rather than demonstrates it,
and Section~\ref{sec:conclusions} names the experiment that would settle it.

\begin{figure}[t]
\centering
\includegraphics[width=\textwidth]{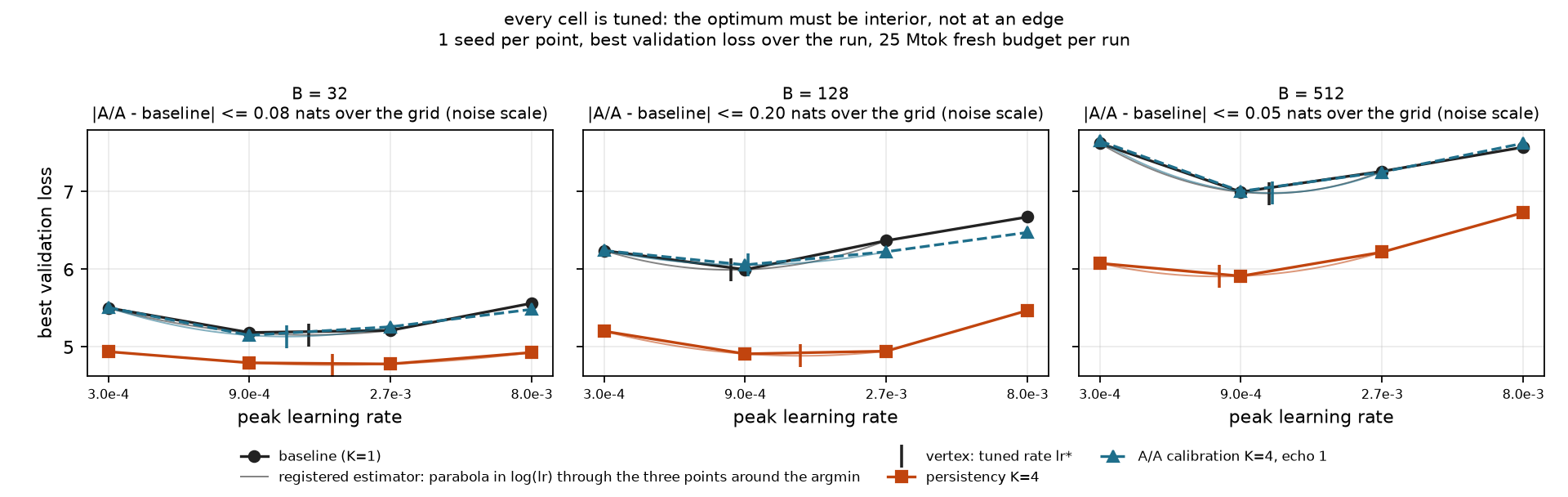}
\caption{The tuning evidence, on the coarse grid where the learning rate is
actually chosen: best validation loss against peak learning rate, one panel per
batch size. One seed per point, at the coarse stage's budget, best validation
loss over the run. Thin curves: the registered estimator of Amendment~2, the
parabola in $\log \eta$ through the three grid points around the argmin, its
vertex marked as the tuned rate. The largest gap between the A/A arm and the
baseline over the grid, \zfouraaspreada{}, \zfouraaspreadb{} and
\zfouraaspreadc{} nats at $B = 32$, $128$ and $512$, is the noise scale of a
single point. Every cell has its optimum in the interior of the swept range---the
sweep is wide enough to have found it, in all nine cells. Without this figure
the 2018 criticism repeats itself, which is why it is a main-text figure and not
an appendix one.}
\label{fig:lr}
\end{figure}

Figure~\ref{fig:lr} is the evidence behind Table~\ref{tab:rho} and behind the
claim that the comparison is between tuned arms. It is drawn on the coarse
stage, and deliberately so: the final stage carries forward only the best
\nlrfinal{} learning rates per cell, so its argmin is always at an edge of what
remains and the figure would answer a question about the surviving grid rather
than about the tuning. This is the same asymmetry that the edge-policy decision
of Section~\ref{sec:setup:prereg} addresses, and it is worth seeing drawn.

The learning curves themselves---validation loss against fresh tokens and
against joules, one figure per $B$---are collected in
Appendix~\ref{app:curves}. They show where the arms cross and how the frozen
targets sit on the evaluation grid, which no ratio at a single target can show.

\subsection{Secondary axes, absolute costs, and the matched arm}
\label{sec:results:secondary}

Everything in this subsection is descriptive. The primary study runs exactly
one confirmatory test, on the step axis, and spends no $\alpha$ anywhere else
(Section~\ref{sec:method:hypothesis}; the family of three tests is counted at
the head of this section); what follows is estimation with
intervals, read on the \nseedsnow{} seeds per batch size of the closed
stage, at the frozen targets of
Table~\ref{tab:targets}, each arm at its own best learning rate, paired by
seed. Unless stated otherwise a ratio is the geometric mean over seeds, with a
95\% $t$-interval computed on the logarithms and exponentiated back---the
estimator of the logarithmic branch of the primary test. The step column of
Table~\ref{tab:wins} is the same arithmetic as the primary endpoint, but it is
not the primary reading, which is the registered test of
Section~\ref{sec:results:primary}.

\paragraph{Fresh tokens: a check, not a second test.}
At $M = 1$ the fresh-token ratio is the step ratio divided by $K$
(Section~\ref{sec:method:hypothesis}), so the fresh-token axis carries no
information the step axis does not, and we verify the identity rather than test
it. The largest relative deviation of the measured fresh-token ratio from
$R/\Kpers$ over the seeds is \freshdeva\% at $B = 32$,
\freshdevb\% at $B = 128$ and \freshdevc\% at $B = 512$, against
a tolerance of $\freshdevtol\%$ written into the analysis code. The last value
trips the tolerance, and the excess has a mechanism: the deviation equals
$K - 1$ divided by \emph{that seed's} steps to target, to machine precision,
and is therefore identical across seeds at $B = 128$ and $B = 512$, where every
seed reaches the target at its first evaluation. This is the evaluation-grid
offset of Section~\ref{sec:setup:prereg}. A $K > 1$ arm is evaluated $K - 1$
steps into a pool that the counter has already charged in full, so at every
evaluation it has trained on fewer tokens than it has been charged for, and the
identity holds up to $K - 1$ steps (three, at $K = \Kpers$) out of the
\stepspersalo--\stepspersahi{} (a range over seeds), \stepspersb{}
and \stepspersc{} steps the arm needs to reach the target at the three
batch sizes---a quantization of the grid, largest where the run to target is
shortest, and not a fault of the counters. The null arm shows the same offset,
and we verify it rather than assert it: for the A/A arm the ratio
$R_{\text{fresh}}/R_{\text{steps}}$ equals $1 + (K - 1)/t$ seed by seed, to
machine precision, with $t$ that seed's steps to target
(\aastepscmin--\aastepscmax{} at $B = 512$), so that the offset on the
geometric means is \aaoffsetc{}. At $B = 512$ the A/A fresh-token ratio is
$\aafreshc$, with interval $[\aafreshclo,\,\aafreshchi]$, off its
construction value of one on this axis alone. The plan's clause~6.4---the
harness-integrity gate of Section~\ref{sec:results:aa}---does not name a cost
axis. We read it on the step ratio, the scale on which
Section~\ref{sec:method:hypothesis} defines $R$, and we record that reading
here: on steps the null covers one at every $B$ ($\aastepc$
$[\aastepclo,\,\aastepchi]$ at $B = 512$), while on fresh tokens it carries the
evaluation-grid offset just described. As a consequence, we claim nothing on
the fresh-token axis at $B = 512$ finer than that offset. The step axis was
chosen for the gate after this excursion had been seen, and we say so. The
gate on the other axes is in Table~\ref{tab:wins}: on seconds and joules the
null arm's interval covers one at every $B$, on fresh tokens it does not at
$B = 512$, for the reason given.

\paragraph{Seconds and joules.}
Of the \nrunsnow{} runs of the stage, \primaryontfour{} executed on Kaggle Tesla
T4 cards (\gpupowerlimit~W power limit) and \primaryonupscale{}---the whole
group $B = 512$, seed $7$---on the Blackwell of
Section~\ref{sec:results:bridge}; every joule came from the NVML total-energy
counter, one mechanism on both cards, and no $(B, \text{seed})$ group straddles
two. The per-machine rule of Amendment~3
(Section~\ref{sec:setup:hardware}) therefore censors no ratio, and the
sampling-coverage clause of the plan was never invoked, since the hardware
counter does not sample. Absolute costs are reported for the T4 and, per the
amendment, are never pooled across cards: over \nseeds{} seeds at $B = 32$ and
$B = 128$, and over the \nseedstfourc{} that ran there at $B = 512$. The one
primary group on the Blackwell, $B = 512$ seed~$7$, is a single seed: its
absolute seconds and joules stay in the run archive, and the Blackwell
absolutes of the replication, eight seeds per cell, are in
Table~\ref{tab:e2abs}. The tuned baseline reaches
the target in $\basesecsa \pm \basesecsasd$, $\basesecsb \pm \basesecsbsd$ and
$\basesecsc \pm \basesecscsd$ seconds of training at $B = 32$, $128$ and $512$,
for $\basekJa \pm \basekJasd$, $\basekJb \pm \basekJbsd$ and
$\basekJc \pm \basekJcsd$~kJ of GPU energy (mean $\pm$ standard deviation over
seeds); the persistency arm needs $\perssecsa \pm \perssecsasd$,
$\perssecsb \pm \perssecsbsd$ and $\perssecsc \pm \perssecscsd$ seconds and
$\perskJa \pm \perskJasd$, $\perskJb \pm \perskJbsd$ and
$\perskJc \pm \perskJcsd$~kJ. Paired within seed and card, the time ratios are
$\Rseca$ $[\Rsecalo,\,\Rsecahi]$, $\Rsecb$ $[\Rsecblo,\,\Rsecbhi]$ and
$\Rsecc$ $[\Rsecclo,\,\Rsecchi]$, and the energy ratios
$\Rjoulea$ $[\Rjoulealo,\,\Rjouleahi]$, $\Rjouleb$ $[\Rjouleblo,\,\Rjoulebhi]$
and $\Rjoulec$ $[\Rjouleclo,\,\Rjoulechi]$. Each lies within a few percent of
the step ratio at the same batch size ($\Rstepa$, $\Rstepb$ and $\Rstepc$),
and every one of these intervals lies entirely above one---at $B = 128$ and
$B = 512$ as ceilings, the persistency leg being the cost of its first
evaluation. At $B = 512$ the paired ratios pool the seven T4 pairs with the
Blackwell pair, each pair on its own card. On the seven T4 pairs alone the
step ratio is \loostepc{} $[\loosteploc,\,\loostephic]$, the time ratio
\loosecc{} $[\loosecloc,\,\loosechic]$, the energy ratio \loojoulec{}
$[\loojouleloc,\,\loojoulehic]$ and the fresh-token ratio \loofreshc{}
$[\loofreshloc,\,\loofreshhic]$, the null arm at \looAAsecc{}
$[\looAAsecloc,\,\looAAsechic]$ on seconds and \looAAjoulec{}
$[\looAAjouleloc,\,\looAAjoulehic]$ on joules. The Blackwell pair is the
outlier of the seconds cell: it widens the interval from
$[\loosecloc,\,\loosechic]$ to $[\Rsecclo,\,\Rsecchi]$, the widest of
Table~\ref{tab:wins}, and the null arm's to $[\aasecclo,\,\aasecchi]$, while
it leaves the joule cell where it was (\loojoulec{} against \Rjoulec{}). The
discrepancy is in wall-clock per step on a card \cardspeedratio{} times
faster---the baseline reaches $\tau(512)$ in \cardspeedTfourSec{} seconds on
the T4 against \cardspeedBlackwellSec{} on the Blackwell---not in energy per
step. On that pair the two arms' mean power parts by the same wall-clock
difference: mean power is joules per step over seconds per step, and only the
denominator moved.

Per step, the persistency arm's seconds relative to the baseline's are
\primPSpersseca{} $[\primPSperssecloa,\,\primPSperssechia]$, \primPSperssecb{}
$[\primPSpersseclob,\,\primPSperssechib]$ and \primPSperssecc{}
$[\primPSperssecloc,\,\primPSperssechic]$ at the three batch sizes, and its
joules per step \primPSpersjoulea{} $[\primPSpersjouleloa,\,\primPSpersjoulehia]$,
\primPSpersjouleb{} $[\primPSpersjoulelob,\,\primPSpersjoulehib]$ and
\primPSpersjoulec{} $[\primPSpersjouleloc,\,\primPSpersjoulehic]$. The null
arm, which reuses nothing, shows \primPSaaseca{}
$[\primPSaasecloa,\,\primPSaasechia]$, \primPSaasecb{}
$[\primPSaaseclob,\,\primPSaasechib]$ and \primPSaasecc{}
$[\primPSaasecloc,\,\primPSaasechic]$ on seconds and \primPSaajoulea{}
$[\primPSaajouleloa,\,\primPSaajoulehia]$, \primPSaajouleb{}
$[\primPSaajoulelob,\,\primPSaajoulehib]$ and \primPSaajoulec{}
$[\primPSaajouleloc,\,\primPSaajoulehic]$ on joules, the same sign and size.
Channel~1 of
Section~\ref{sec:energy:channels} is therefore bounded, not switched off: the
residual is a few percent, and an arm that reuses nothing shares it. The one
persistency residual above one is joules per step at $B = 512$, in the cell
with the Blackwell pair, where the null arm's is above one as well. In the replication the residuals are \etwoPSperssecc{}
$[\etwoPSperssecloc,\,\etwoPSperssechic]$ on seconds and \etwoPSpersjoulec{}
$[\etwoPSpersjouleloc,\,\etwoPSpersjoulehic]$ on joules at $B = 512$, the
null arm's \etwoPSaasecc{} $[\etwoPSaasecloc,\,\etwoPSaasechic]$ and
\etwoPSaajoulec{} $[\etwoPSaajouleloc,\,\etwoPSaajoulehic]$ there, and both
arms' within one percent at every $B$.

\paragraph{Outright wins and losses.}
Table~\ref{tab:wins} collects the ratios on every axis at the frozen targets,
together with the null arm; Table~\ref{tab:adjaxes} repeats the reading at the
re-adjusted targets of Table~\ref{tab:targets}, and Table~\ref{tab:e2axes} is
the replication's at $\tau'$. Following the plan, persistency \emph{wins
outright} on an axis when the interval lies entirely below one. The plan
defines no symmetric term; we call a cell an \emph{outright loss} when the
interval lies entirely above one and the cost in it is a measurement rather
than a ceiling. At the frozen targets the treatment wins outright on fresh
tokens at $B = 128$ and $B = 512$, and at $B = 32$ the fresh-token interval
covers one. It loses outright on steps, seconds and joules at $B = 32$. At
$B = 128$ and $B = 512$ those intervals lie above one too, but the persistency
cost in them is the cost of the arm's first evaluation, an upper bound
(Section~\ref{sec:results:primary}), and no outright loss is assigned there.
At the re-adjusted targets no cell is a ceiling: the loss on steps, seconds
and joules stands at $B = 32$ and $B = 128$, and at $B = 512$ every interval
covers one (\adjRstepc{} on steps, \adjRsecc{} on seconds, \adjRjoulec{} on
joules). The null arm is marked nowhere, except in the fresh-token cell at
$B = 512$ of the frozen targets, discussed above.

\begin{table}[t]
\centering
\footnotesize
\begin{tabular}{llcccc}
\toprule
arm & $B$ & fresh tokens & steps (= compute) & seconds & joules \\
\midrule
persistency & $32$
  & $\Rfresha$ & $\Rstepa^{\,\blacktriangle}$ & $\Rseca^{\,\blacktriangle}$ & $\Rjoulea^{\,\blacktriangle}$ \\
 & & {\scriptsize$[\Rfreshalo,\,\Rfreshahi]$} & {\scriptsize$[\Rstepalo,\,\Rstepahi]$}
   & {\scriptsize$[\Rsecalo,\,\Rsecahi]$} & {\scriptsize$[\Rjoulealo,\,\Rjouleahi]$} \\
persistency & $128$
  & $\Rfreshb^{\,\blacktriangledown}$ & $\Rstepb^{\,\triangle}$ & $\Rsecb^{\,\triangle}$ & $\Rjouleb^{\,\triangle}$ \\
 & & {\scriptsize$[\Rfreshblo,\,\Rfreshbhi]$} & {\scriptsize$[\Rstepblo,\,\Rstepbhi]$}
   & {\scriptsize$[\Rsecblo,\,\Rsecbhi]$} & {\scriptsize$[\Rjouleblo,\,\Rjoulebhi]$} \\
persistency & $512$
  & $\Rfreshc^{\,\blacktriangledown}$ & $\Rstepc^{\,\triangle}$ & $\Rsecc^{\,\triangle}$ & $\Rjoulec^{\,\triangle}$ \\
 & & {\scriptsize$[\Rfreshclo,\,\Rfreshchi]$} & {\scriptsize$[\Rstepclo,\,\Rstepchi]$}
   & {\scriptsize$[\Rsecclo,\,\Rsecchi]$} & {\scriptsize$[\Rjouleclo,\,\Rjoulechi]$} \\
\midrule
A/A null & $32$
  & $\aafresha$ & $\aastepa$ & $\aaseca$ & $\aajoulea$ \\
 & & {\scriptsize$[\aafreshalo,\,\aafreshahi]$} & {\scriptsize$[\aastepalo,\,\aastepahi]$}
   & {\scriptsize$[\aasecalo,\,\aasecahi]$} & {\scriptsize$[\aajoulealo,\,\aajouleahi]$} \\
A/A null & $128$
  & $\aafreshb$ & $\aastepb$ & $\aasecb$ & $\aajouleb$ \\
 & & {\scriptsize$[\aafreshblo,\,\aafreshbhi]$} & {\scriptsize$[\aastepblo,\,\aastepbhi]$}
   & {\scriptsize$[\aasecblo,\,\aasecbhi]$} & {\scriptsize$[\aajouleblo,\,\aajoulebhi]$} \\
A/A null & $512$
  & $\aafreshc^{\,\blacktriangle}$ & $\aastepc$ & $\aasecc$ & $\aajoulec$ \\
 & & {\scriptsize$[\aafreshclo,\,\aafreshchi]$} & {\scriptsize$[\aastepclo,\,\aastepchi]$}
   & {\scriptsize$[\aasecclo,\,\aasecchi]$} & {\scriptsize$[\aajouleclo,\,\aajoulechi]$} \\
\bottomrule
\end{tabular}
\caption{Cost-to-target ratios of the persistency arm and of the A/A
calibration arm to the tuned baseline, on every cost axis, at the frozen
targets: geometric mean over seeds, with the 95\% $t$-interval beneath. Below
one favors reuse. $\blacktriangledown$ marks an \emph{outright win} (interval
entirely below one) and $\blacktriangle$ an \emph{outright loss} (interval
entirely above one); no mark means the interval covers one. $\triangle$ marks
a cell whose interval lies above one but whose persistency cost is the cost
of the arm's first evaluation, a ceiling (Section~\ref{sec:results:primary}):
no outright loss is assigned there. The fresh-token cells at $B = 128$ and
$B = 512$ rest on the same ceiling, and their $\blacktriangledown$ is read a
fortiori: a smaller cost only widens
the win. Compute is a deterministic transform of
steps at fixed $B$ and shares its column. The table is descriptive in the
sense of the registered plan: the primary study runs one confirmatory test,
on the step axis, and no $\alpha$ is spent here. Every cell has \npairs{}
pairs, on the closed stage; at $B = 512$ one of them is on the Blackwell
(Section~\ref{sec:results:bridge}), and the seven-pair reading is in the
text.}
\label{tab:wins}
\end{table}

\begin{table}[t]
\centering
\footnotesize
\begin{tabular}{llcccc}
\toprule
arm & $B$ & fresh tokens & steps (= compute) & seconds & joules \\
\midrule
persistency & $32$
  & $\adjRfresha$ & $\adjRstepa^{\,\blacktriangle}$ & $\adjRseca^{\,\blacktriangle}$ & $\adjRjoulea^{\,\blacktriangle}$ \\
 & & {\scriptsize$[\adjRfreshloa,\,\adjRfreshhia]$} & {\scriptsize$[\adjRsteploa,\,\adjRstephia]$}
   & {\scriptsize$[\adjRsecloa,\,\adjRsechia]$} & {\scriptsize$[\adjRjouleloa,\,\adjRjoulehia]$} \\
persistency & $128$
  & $\adjRfreshb^{\,\blacktriangledown}$ & $\adjRstepb^{\,\blacktriangle}$ & $\adjRsecb^{\,\blacktriangle}$ & $\adjRjouleb^{\,\blacktriangle}$ \\
 & & {\scriptsize$[\adjRfreshlob,\,\adjRfreshhib]$} & {\scriptsize$[\adjRsteplob,\,\adjRstephib]$}
   & {\scriptsize$[\adjRseclob,\,\adjRsechib]$} & {\scriptsize$[\adjRjoulelob,\,\adjRjoulehib]$} \\
persistency & $512$
  & $\adjRfreshc^{\,\blacktriangledown}$ & $\adjRstepc$ & $\adjRsecc$ & $\adjRjoulec$ \\
 & & {\scriptsize$[\adjRfreshloc,\,\adjRfreshhic]$} & {\scriptsize$[\adjRsteploc,\,\adjRstephic]$}
   & {\scriptsize$[\adjRsecloc,\,\adjRsechic]$} & {\scriptsize$[\adjRjouleloc,\,\adjRjoulehic]$} \\
\midrule
A/A null & $32$
  & $\adjAAfresha$ & $\adjAAstepa$ & $\adjAAseca$ & $\adjAAjoulea$ \\
 & & {\scriptsize$[\adjAAfreshloa,\,\adjAAfreshhia]$} & {\scriptsize$[\adjAAsteploa,\,\adjAAstephia]$}
   & {\scriptsize$[\adjAAsecloa,\,\adjAAsechia]$} & {\scriptsize$[\adjAAjouleloa,\,\adjAAjoulehia]$} \\
A/A null & $128$
  & $\adjAAfreshb$ & $\adjAAstepb$ & $\adjAAsecb$ & $\adjAAjouleb$ \\
 & & {\scriptsize$[\adjAAfreshlob,\,\adjAAfreshhib]$} & {\scriptsize$[\adjAAsteplob,\,\adjAAstephib]$}
   & {\scriptsize$[\adjAAseclob,\,\adjAAsechib]$} & {\scriptsize$[\adjAAjoulelob,\,\adjAAjoulehib]$} \\
A/A null & $512$
  & $\adjAAfreshc$ & $\adjAAstepc$ & $\adjAAsecc$ & $\adjAAjoulec$ \\
 & & {\scriptsize$[\adjAAfreshloc,\,\adjAAfreshhic]$} & {\scriptsize$[\adjAAsteploc,\,\adjAAstephic]$}
   & {\scriptsize$[\adjAAsecloc,\,\adjAAsechic]$} & {\scriptsize$[\adjAAjouleloc,\,\adjAAjoulehic]$} \\
\bottomrule
\end{tabular}
\caption{Table~\ref{tab:wins} at the re-adjusted targets of
Table~\ref{tab:targets}, the targets the registered floor check prescribes
once the confirmatory stage is closed: the same estimator, pairs and marks on
the same \nrunstotal{} runs. No cell is a ceiling at these targets. The
re-adjusted targets at $B = 128$ and $B = 512$ are set by the persistency
arm's own first evaluations and are not exogenous to the treatment
(Section~\ref{sec:results:primary}); the frozen targets carry the registered
verdict.}
\label{tab:adjaxes}
\end{table}

\paragraph{The learning-rate-matched arm.}
Amendment~2 of the plan (Section~\ref{sec:setup:prereg}) runs the persistency
configuration at the baseline's tuned rate $\eta_1$, at full length, on
\lrmn{} seeds per batch size (\lrmnruns{} runs), so that the two legs of the
ratio differ by the reuse alone; the contrast
$\Delta = R_{\text{matched}} - R_{\text{tuned}}$ then measures what re-tuning
the learning rate buys the treatment. The plan declares the paired test on
$\Delta$ underpowered at $n = \lrmn{}$, and we repeat that here. It is the
dose--response reading of the design: the matched arm holds the step size fixed
and varies only the dose of reuse, and if it returns the ratio the tuned arm
returns, the effect is the dose and not the rate. On this data the arm is
degenerate, in a way that is itself the answer: $\eta_1 = \etaone$ at all three
batch sizes, and at all three the persistency arm's cost-minimizing rate among
the \nlrfinal{} final-stage rates is the same $\eta_1$, so the matched
configuration \emph{is} the tuned configuration. At $B = 128$ and $B = 512$ the
amendment lets the run be reused and $\Delta = 0$ by construction, the ratios
being $\lrmtunedstepb \pm \lrmtunedstepbsem$ and
$\lrmtunedstepc \pm \lrmtunedstepcsem$ on steps
($\lrmtunedfreshb \pm \lrmtunedfreshbsem$ and
$\lrmtunedfreshc \pm \lrmtunedfreshcsem$ on fresh tokens)---arithmetic
means $\pm$ standard error over the \lrmn{} seeds, as the amendment specifies,
and therefore not the geometric means of Table~\ref{tab:wins}. At $B = 32$ the
matched arm is an independent re-execution of the same configuration: it
reached the target in \lrmstepsmatcheda{} steps against
\lrmstepsfinala{} for the final-stage runs at seeds $0$, $1$ and $2$,
giving $\Delta = \lrmdeltastepa \pm \lrmdeltastepasem$ on steps
(tuned $\lrmtunedstepa \pm \lrmtunedstepasem$, matched
$\lrmmatchstepa \pm \lrmmatchstepasem$) and
$\Delta = \lrmdeltafresha \pm \lrmdeltafreshasem$ on fresh tokens---the
replication noise between two runs of one configuration. The tuned leg is
read seed by seed on the matched seeds only. Since
$\eta^\star(K)$ coincides with $\eta_1$ among the final-stage rates at every
$B$, no re-tuning took place: $\Delta$ is zero by construction at $B = 128$ and
$B = 512$ and replication noise at $B = 32$, so the matched arm adds no
evidence of its own beyond confirming that the treatment obtained its ratio
without a different step size. The evidence against the re-parametrization
reading remains $\rho(B)$ of Table~\ref{tab:rho}, which is indeterminate at
$B = 512$. Note that $\rho$ is read on the interpolated coarse optimum of best
validation loss, whereas the rate used here is the argmin of cost-to-target
over the \nlrfinal{} surviving rates; the two need not coincide. Observe,
finally, that the plan's reading of a negative $\Delta$ as a caveat on the
two-stage protocol presupposes that a re-tuning took place; none did, and the
caveat does not apply.

\paragraph{What this adds to the energy reading.}
The finer accounting of this subsection leaves the reading of
Section~\ref{sec:results:axes} where it was, and bounds it. On the T4, with a
pipeline that never starves the card, a persistency step costs what a baseline
step costs in seconds and in joules within the residuals above. So the arm
that needs $\Rstepc$ times the steps at $B = 512$ at the frozen target needs
$\Rjoulec$ times the joules there, both as ceilings: joules lose where steps
lose, on this workload. At the re-adjusted target the same cell gives
\adjRstepc{} on steps and \adjRjoulec{} on joules, both intervals covering
one, and at $\tau'$ on the replication's card \etwoTstepc{} and
\etwoTjoulec{}, both below one.

\subsection{Is immediate reuse just an early second epoch?}
\label{sec:results:e1}

A reader who accepts everything above may still hold that nothing here is about
reuse at all: a run that visits each example four times has simply taken four
epochs, and the interesting comparison is not against a baseline that sees more
data but against the same data seen the same number of times in the ordinary
way. The second registered extension, at commit \texttt{\preregvfivehash{}} of
the registration (prereg-v5), is that comparison, and it is built so that only
the \emph{spacing} of the repetitions differs. Each seed draws a
fixed stream of \eonebudget{} Mtok, and two arms run over it: one takes the
persistency policy in a single pass, with each pool feeding four consecutive
updates; the other takes four epochs of the same stream at $K = 1$, the
repetitions as far apart as the data allows. Both make \eonesteps{} optimizer
steps at $B = 32$, $128$ and $512$, both push \eonefwdtok{} Mtok through the
forward pass, and both draw the same \eoneitems{} distinct examples---not
merely the same count: the digest of the drawn indices is equal arm to arm,
seed by seed, at every batch size (Appendix~\ref{app:passcount}). The endpoint
is the final validation loss at the shared step budget, a quality at a fixed
budget, which Section~\ref{sec:method:cost} otherwise excludes. The
registration makes this the one exception and gives the reason: with
\eonebudget{} Mtok of distinct data neither arm reaches the frozen $\tau(B)$,
so cost to target would censor every pair. Within each (arm, cell, seed) the
loss is the minimum over the arm's two rates, the same order statistic on both
sides of $\Delta$; the selection carries a min-over-two bias common to the two
arms, not quantified here. The two rates per cell come from a coarse stage of
the extension's own on the Blackwell, \eonencoarse{} runs at one seed and a
quarter of the budget. Its argmin by best validation loss is
$\lrdispEoneCoarse$ in all six cells, both arms at every batch size, the
primary's $\eta_1$. The primary's T4 coarse argmins are $\lrdispCoarseBase$
in five cells and $\lrdispCoarsePersa$ for persistency at $B = 32$, the grid
column of Table~\ref{tab:rho}. That agreement is a hint that the tuned rate
transfers between the cards, no more: the workloads differ, an epoch contrast
at \eonebudget{} Mtok of distinct data against a single pass at four times
the data. The test is two-sided at $n = \nseeds{}$, because the
outcome of interest is as much a null as an effect, and the registration
spends its $\alpha$ at $B = 512$ only: the readings at $B = 32$ and $B = 128$
are descriptive, with the same estimate and interval. Both arms executed on
the Blackwell, on one card; the bridge of Section~\ref{sec:results:bridge} is
defined on steps to target and does not speak to a fixed-budget loss
endpoint.

At the two smaller batch sizes, spacing wins in every seed. Immediate reuse
ends $\eonedeltaa{} \pm \eonedeltaasem{}$ nats worse at $B = 32$ (mean $\pm$
its standard error; 95\% interval $[\eoneCIloa,\,\eoneCIhia]$,
$t = \eonetstata{}$, $p = \eonepa{}$, descriptive) and
$\eonedeltab{} \pm \eonedeltabsem{}$ worse at $B = 128$
($[\eoneCIlob,\,\eoneCIhib]$, $t = \eonetstatb{}$, $p = \eonepb{}$,
descriptive), losing in every one of the \nseeds{} seeds at both. At
$B = 512$, the registered test, the difference is
$\eonedeltac{} \pm \eonedeltacsem{}$ nats, 95\% interval
$[\eoneCIloc,\,\eoneCIhic]$, $t = \eonetstatc{}$, $p = \eonepc{}$; immediate
reuse is ahead in \eonewinsc{} seeds of \nseeds{}, a sign test at
$p = \eoneSignPc$. The interval contains zero and is not contained in the
pre-declared $\pm\delta^\star = \pm\eonedeltastar$ nats: on the registered
reading the two policies are \emph{indistinguishable at this budget},
published as such with the interval's width, and no equivalence is claimed.
Resolving $\delta^\star$ at power \eonepower{} with the observed dispersion of
\eoneSDc{} nats would take about \eoneNneededc{} seeds against the \nseeds{}
run. The absolute losses locate the three readings: single pass against four
epochs end at \eonepassa{} against \eoneepocha{} nats at $B = 32$,
\eonepassb{} against \eoneepochb{} at $B = 128$ and \eonepassc{} against
\eoneepochc{} at $B = 512$. The registered third reading is the bridge
baseline of Section~\ref{sec:results:bridge}: fresh data throughout, at
$\eta_1$, on the same card, ending at \eoneanchorBhundreda{},
\eoneanchorBhundredb{} and \eoneanchorBhundredc{} nats
($n = \eoneanchorBhundredna$ per batch size). Those runs stop within four
steps of $S(B)$, so their final loss stands in for the loss at $S(B)$; the
anchor is descriptive. The registered secondary reading of the same
contrast, the best validation loss over the evaluation grid instead of the
final one, gives the same three differences: \eonebestdeltaa{}
$[\eonebestloa, \eonebesthia]$, \eonebestdeltab{} $[\eonebestlob,
\eonebesthib]$ and \eonebestdeltac{} $[\eonebestloc, \eonebesthic]$ nats
(single pass minus four epochs, paired by seed, $n = \nseeds$); no batch
size changes sign between the two readings, so the discordance rule the
plan fixed for them does not fire. At $B = 512$ both arms stop at a depth the smaller
batches pass early in their budget, so the null there is a null at shallow
depth as much as at large batch.

The reading is the one the 2018 conjecture would have made, within its
limits. Repeating a minibatch immediately costs against spacing the
repetitions out at $B = 32$ and $B = 128$, in every seed, by amounts a stale
gradient could produce. That at large batch the gradient is estimated well
enough for four consecutive uses to cost nothing is a conjecture consistent
with the null at $B = 512$, not something this experiment measures: the
interval does not show it. At $B = 512$ and \eonebudget{} Mtok of distinct
data, on this evidence, persistency is not distinguishable from an epoch
schedule over the same data; at the two smaller batch sizes it is worse than
one.

\section{Conclusions and future work}
\label{sec:conclusions}

We have taken an eight-year-old idea and subjected it to the test that was
missing when it was proposed: a baseline tuned as carefully as the method, a
target-based metric, a pre-registered analysis plan, a calibrated noise floor,
and cost measured in joules as well as in steps. Of the two claims we kept
separable from Section~\ref{sec:intro} onward, the optimization claim comes out
as follows. On the closed stage the registered test returns \emph{conjecture
supported}: the point estimate of the step penalty of reuse at $B = 512$ is
below half of its value at $B = 32$---a fortiori, since $\hat g(512)$ is a
ceiling there---the reading is unchanged under every registered perturbation of
the target, and the replication of Section~\ref{sec:results:e2} returns the
same verdict on new seeds and a different card. Supported,
however, does not mean cheaper, and the distinction is the finding. In
optimizer steps and in accelerator compute, persistency reaches the frozen
targets at a higher cost than the tuned baseline at every batch size, while at
the deeper $\tau'$ of the replication it is cheaper at $B = 512$ on both axes.
In fresh tokens it reaches the frozen targets at a lower cost at $B = 128$ and
$B = 512$---fewer than half---while at $B = 32$ the two arms are
indistinguishable. Both halves are the finding, and neither is to be read
without the other. The fresh-token half is read against the epoch contrast of
Section~\ref{sec:results:e1}, at \eonebudget{} Mtok of data and equal steps:
immediate reuse ends $\eonedeltaa{}$ nats worse than four spaced epochs at
$B = 32$ and $\eonedeltab{}$ nats worse at $B = 128$, in every seed. At
$B = 512$ the difference is $\eonedeltac{}$ nats with a 95\% interval of
$[\eoneCIloc, \eoneCIhic]$: indistinguishable at that budget, which is not
equivalence. The energy claim is
narrower than the one we set out to make, and we state it in the plainest
terms: in the compute-bound regime measured here, and at the frozen
targets, reuse costs \emph{more} GPU energy to reach the target than the tuned
baseline, not less; the reading at $B = 512$ is a ceiling taken inside the
first evaluation interval (Appendix~\ref{app:curves}), and at the re-adjusted
target of Table~\ref{tab:targets} the step penalty at $B = 512$ is already
below zero. That qualification now has an answer of its own. On the registered replication of
Section~\ref{sec:results:e2}---new seeds, a grid in optimizer steps, targets on
full-length baselines---no cost is a ceiling, and at $B = 512$ reuse reaches
the deeper target $\tau'$ on $\etwoRjoulec{}$ of the baseline's joules,
$\etwoRsecc{}$ of its seconds and $\etwoRstepc{}$ of its compute, against an
A/A floor of $\etwoAAjoulec{}$ $[\etwoAAjouleloc, \etwoAAjoulehic]$ on joules.
That estimate is conditional on the inherited rate pair transferring to the
Blackwell, which the bridge of Section~\ref{sec:results:bridge} checks at one
fixed rate only; Extension~E3 reads the full grid there
(Section~\ref{sec:results:e3}): on that card both argmins fall on $\eta_1$,
inside the pair, with $\rho'(512) = \ethreelrRho$ at one seed, and the
baseline at three times $\eta_1$ does not reach the deeper target at all. The energy
statement above is therefore a statement about a target, not about a regime:
at the frozen $\tau(B)$ reuse costs more energy at every batch size, and at
$\tau'$ it costs less at $B = 512$ and there only, no resolved depth bringing
the step ratio below one at $B = 32$ or $B = 128$ (Appendix~\ref{app:rtau}).
Reuse buys data at $B = 128$ and $B = 512$ and not at $B = 32$---%
indistinguishable at the frozen target, \Rfresha{} $[\Rfreshalo, \Rfreshahi]$,
and \etwoRfresha{} times the baseline's fresh tokens at the deeper one---and
buys energy only at $B = 512$ and only past the deeper target. Beside that
reading stands the schedule-matched control of Section~\ref{sec:results:e3}:
on one cosine in steps for all three arms, reuse at $B = 512$ costs
\ethreeRstep{} $[\ethreeRsteplo, \ethreeRstephi]$ of the baseline's steps and
\ethreeRjoule{} $[\ethreeRjoulelo, \ethreeRjoulehi]$ of its joules, so the
effect of reuse and the position on the schedule are not separable at
$n = \nseeds{}$, and the energy statement at $B = 512$ is read with that
qualification. Data is what the settings named in Section~\ref{sec:intro}
pay for: a corpus that cannot grow, a pipeline that pays per sample, a stream
that cannot be rewound. For them the fresh-token reading is the result, at
large batch and at equal steps; for a pipeline with the data already in
memory it buys nothing, and where a second pass is possible spaced epochs do
at least as well (Section~\ref{sec:results:e1}). On a pre-tokenized text
pipeline the accelerator is
never starved, a step costs what a step costs, and joules follow steps.
Channel~2 of Section~\ref{sec:energy:channels} has nothing to skip by
construction; channel~1 is bounded by measurement, the per-step offsets of the
persistency arm and of the A/A arm being of the same order
(Section~\ref{sec:results:secondary}). Every joule that reuse saves here must
therefore come from channel~3: at the frozen targets channel~3 does not pay,
and at $\tau'$ it pays at $B = 512$ and only there. The larger energy case for
reuse rests on the input-bound regime, where the pipeline work it skips is
real work. This study bounds that case and does not
demonstrate it, and the fresh-token gain above must not be read as an energy
gain---it is a statement about data, and it becomes a statement about joules
only on a pipeline where data costs joules.

The methodological claim is the one we regard as settled, since each piece of
the apparatus bought something specific. The tuned baseline gives $\rho(B)$, a
registered descriptive diagnostic. At two of the three batch sizes it is
inconsistent with the larger-learning-rate reading, at the precision a
single-seed coarse sweep affords; at the third it is reported as indeterminate
rather than argued away (Table~\ref{tab:rho}). The
target metric makes every comparison but the epoch contrast a cost-to-target
comparison; it does not free the comparison from the learning-rate schedule,
whose shape is fixed and anchored to each arm's own horizon
(Section~\ref{sec:results:e3}). The A/A null covers one on steps, seconds and
joules at every batch size---on fresh tokens at $B = 512$ it carries the
evaluation-grid offset of Section~\ref{sec:results:secondary}---so the effect
is not apparatus, and its dispersion is the yardstick against which every ratio in
the paper is read. Finally, the pre-registration fixed the hypothesis, the
estimator and the decision rule before the confirmatory runs, so that the
verdict above is the one we committed to and not the one we preferred. Note
that this is what answering the tuned-baseline objection looks like, as opposed
to repeating it: the 2018 evidence could not tell whether persistency was
anything more than a re-parametrized learning rate, and the present evidence
can at $B = 32$ and $B = 128$, at the precision just stated, while reporting
the question as open at $B = 512$ rather than closed.

A lesson learned concerns measurement rather than optimization. Of the two
instruments available for reporting the energy of a training run, the one that
looks authoritative---the wall-plug reading of the node---answers a different
question than the one being asked, and does so by a factor that is large enough
to change conclusions (Section~\ref{sec:energy:ipmi}). The instrument that does
answer the question, the per-device counter, is absent from precisely the
hardware that academic groups own. Any future study that reports joules should
state which of the two it used, on which hardware, and with what share of the
machine allocated to it; we would go further and suggest that a joule without a
recorded provenance should not be accepted in a table.

Several questions are left open, and we name them as future work rather than
leave them implicit.

\begin{itemize}
  \item \textbf{Locality.} As stated in Section~\ref{sec:method:knobs}, our
    third arm is a calibration null and this study makes no claim about
    locality. The genuine experiment---reuse over a non-exchangeable stream,
    that is, sequential shard reads with no shuffling---remains to be run, and
    is the natural next use of the apparatus.
  \item \textbf{A genuinely input-bound arm.} The energy channel that reuse
    should exploit best is the one this study cannot see, since a pre-tokenized
    text pipeline has nothing to skip. A vision workload with online decoding
    and augmentation, with the host-to-device ratio deliberately throttled, is
    the setting where the joule claim can be made rather than bounded.
  \item \textbf{Decorated reuse.} We tested the strict variant---batch intact,
    no augmentation between repeats---which the literature suggests is the
    weakest member of its family. A credible practical recommendation would have
    to place it against reshuffled and augmented repeats.
  \item \textbf{Scale.} The study is budget-limited by construction
    (Section~\ref{sec:setup:workload}). The conjecture is about the large-batch
    limit, and larger batches with larger models is where it should eventually
    be tested; the design transfers unchanged, only the invoice grows.
  \item \textbf{The reuse factor.} $K = \Kpers{}$ is the one dimension of the
    design never varied. The fresh-token saving is $R/K$ by identity
    (Section~\ref{sec:method:hypothesis}), so every fresh-token ratio in this
    paper is a step ratio divided by an untested $K$. A second value of $K$ at
    $B = 128$ and $B = 512$, on the same seeds, is the run that would make the
    data claim a claim about reuse factors.
  \item \textbf{The bridge to the theory of repetition.} The 2024--2025 results
    on what repeated batches make learnable
    \citep{dandi2024, repetita2024, linwu2025} do not cite the applied
    literature on echoing, and the applied literature does not cite them.
    Connecting a provable change in learnability to a measured change in
    cost-to-target is, in our view, the most interesting thing that could be
    done next with this question.
\end{itemize}

\section{Code, data and registration availability}
\label{sec:availability}

The supplementary archive submitted with this paper for review contains,
anonymized, what the results rest on. It holds the training and measurement
harness---the \texttt{mbp} package with the test suite of
Section~\ref{sec:setup:hardware}---and the analysis scripts that produce every
number and table of this paper from the run archives. It holds the frozen
targets $\tau(B)$, the re-adjusted targets and the replication's $\tau'(B)$,
and the per-run archive of every study reported here, each energy reading with
its provenance. It holds the registration at versions v1 to v6 with its
amendments and extensions, the two decision documents that followed it, the
harness review the registration freezes with it, and the OpenTimestamps
receipts. The receipts anchor v4, v5, v6 and the erratum to v6, together with
the decision documents and the harness review; v1 to v3 carry no receipt of
their own and are listed as amendments in force by the stamped v4 manifest.
The archive is released
publicly on acceptance. The registration versions are the tags
\texttt{prereg-v1} to \texttt{prereg-v6} of the source repository, at commits
\texttt{\preregvonehash{}}, \texttt{\preregvtwohash{}},
\texttt{\preregvthreehash{}}, \texttt{\preregvfourhash{}},
\texttt{\preregvfivehash{}} and \texttt{\preregvsixhash{}}, with the erratum to
v6 at \texttt{\preregvsixerratumhash{}}. The SHA-256 of the registration
document at \texttt{prereg-v6} is
{\footnotesize\texttt{\preregvsixsha{}}}. The registration on the OSF
Registries is embargoed until \osfembargo{}; the hash lets a reader match the
supplementary copy to the embargoed one.

\makeatletter
\if@accepted
\subsubsection*{Acknowledgments}
Work partially supported by MiUR, Italy. The author thanks the system
administrators of the DEI blade cluster and of the Upscale HPC facility of the
University of Padova for access, for the exclusive-node energy measurements, and
for the fairshare reset that made the attribution study of
Section~\ref{sec:energy:ipmi} possible.
\fi
\makeatother

\bibliography{refs}

\begin{thebibliography}{18}
\providecommand{\natexlab}[1]{#1}
\providecommand{\url}[1]{\texttt{#1}}
\expandafter\ifx\csname urlstyle\endcsname\relax
  \providecommand{\doi}[1]{doi: #1}\else
  \providecommand{\doi}{doi: \begingroup \urlstyle{rm}\Url}\fi

\bibitem[Agarwal et~al.(2020)Agarwal, Anil, Koren, Talwar, and
  Zhang]{agarwal2020}
Naman Agarwal, Rohan Anil, Tomer Koren, Kunal Talwar, and Cyril Zhang.
\newblock Stochastic optimization with laggard data pipelines.
\newblock In \emph{Advances in Neural Information Processing Systems
  (NeurIPS)}, 2020.

\bibitem[Fischetti(2026)]{osf}
Fischetti, M.
\newblock Pre-registration of this study.
\newblock OSF Registries, embargoed until 2027-08-10, 2026.
\newblock SHA-256 and timestamp receipts in the supplementary archive.

\bibitem[Arnaboldi et~al.(2024)Arnaboldi, Dandi, Krzakala, Pesce, and
  Stephan]{repetita2024}
Luca Arnaboldi, Yatin Dandi, Florent Krzakala, Luca Pesce, and Ludovic Stephan.
\newblock Repetita iuvant: Data repetition allows {SGD} to learn
  high-dimensional multi-index functions.
\newblock \emph{arXiv preprint arXiv:2405.15459}, 2024.
\newblock URL \url{https://arxiv.org/abs/2405.15459}.

\bibitem[Audibert et~al.(2023)Audibert, Chen, Graur, Klimovic, \v{S}im\v{s}a,
  and Thekkath]{audibert2023tfdata}
Andrew Audibert, Yang Chen, Dan Graur, Ana Klimovic, Ji\v{r}\'{i}
  \v{S}im\v{s}a, and Chandramohan~A. Thekkath.
\newblock tf.data service: A case for disaggregating {ML} input data
  processing.
\newblock In \emph{Proceedings of the 2023 ACM Symposium on Cloud Computing
  (SoCC)}, pp.\  358--375, 2023.

\bibitem[Choi et~al.(2019)Choi, Passos, Shallue, and Dahl]{choi2019}
Dami Choi, Alexandre Passos, Christopher~J. Shallue, and George~E. Dahl.
\newblock Faster neural network training with data echoing.
\newblock \emph{arXiv preprint arXiv:1907.05550}, 2019.
\newblock URL \url{https://arxiv.org/abs/1907.05550}.

\bibitem[Chung et~al.(2024)Chung, Gu, Jang, Meng, Bansal, and
  Chowdhury]{chung2024perseus}
Jae-Won Chung, Yile Gu, Insu Jang, Luoxi Meng, Nikhil Bansal, and Mosharaf
  Chowdhury.
\newblock Reducing energy bloat in large model training.
\newblock In \emph{Proceedings of the ACM SIGOPS 30th Symposium on Operating
  Systems Principles (SOSP)}, 2024.
\newblock \doi{10.1145/3694715.3695970}.
\newblock The Perseus system.

\bibitem[Dahl et~al.(2023)Dahl, Schneider, et~al.]{dahl2023}
George~E. Dahl, Frank Schneider, et~al.
\newblock Benchmarking neural network training algorithms.
\newblock \emph{arXiv preprint arXiv:2306.07179}, 2023.
\newblock The AlgoPerf benchmark and its rolling leaderboard.

\bibitem[Dandi et~al.(2024)Dandi, Troiani, Arnaboldi, Pesce, Zdeborov\'{a}, and
  Krzakala]{dandi2024}
Yatin Dandi, Emanuele Troiani, Luca Arnaboldi, Luca Pesce, Lenka Zdeborov\'{a},
  and Florent Krzakala.
\newblock The benefits of reusing batches for gradient descent in two-layer
  networks: Breaking the curse of information and leap exponents.
\newblock In \emph{Proceedings of the 41st International Conference on Machine
  Learning (ICML)}, volume 235 of \emph{PMLR}, pp.\  9991--10016, 2024.
\newblock arXiv:2402.03220.

\bibitem[European Union()]{euaiact}
European Union.
\newblock Regulation ({EU}) 2024/1689 of the european parliament and of the
  council laying down harmonised rules on artificial intelligence ({AI} act).
\newblock Official Journal of the European Union, 2024.
\newblock Annex XI: documentation of training energy consumption for
  general-purpose AI models.

\bibitem[Fischetti et~al.(2018)Fischetti, Mandatelli, and
  Salvagnin]{fischetti2018}
Matteo Fischetti, Iacopo Mandatelli, and Domenico Salvagnin.
\newblock Faster {SGD} training by minibatch persistency.
\newblock \emph{arXiv preprint arXiv:1806.07353}, 2018.
\newblock URL \url{https://arxiv.org/abs/1806.07353}.

\bibitem[Kaddour et~al.(2023)Kaddour, Key, Nawrot, Minervini, and
  Kusner]{kaddour2023}
Jean Kaddour, Oscar Key, Piotr Nawrot, Pasquale Minervini, and Matt~J. Kusner.
\newblock No train no gain: Revisiting efficient training algorithms for
  transformer-based language models.
\newblock In \emph{Advances in Neural Information Processing Systems
  (NeurIPS)}, 2023.

\bibitem[Lin et~al.(2025)Lin, Wu, and Bartlett]{linwu2025}
Licong Lin, Jingfeng Wu, and Peter~L. Bartlett.
\newblock Improved scaling laws in linear regression via data reuse.
\newblock \emph{arXiv preprint arXiv:2506.08415}, 2025.
\newblock URL \url{https://arxiv.org/abs/2506.08415}.

\bibitem[Muennighoff et~al.(2023)Muennighoff, Rush, Barak, Le~Scao, Tazi,
  Piktus, Pyysalo, Wolf, and Raffel]{muennighoff2023}
Niklas Muennighoff, Alexander~M. Rush, Boaz Barak, Teven Le~Scao, Nouamane
  Tazi, Aleksandra Piktus, Sampo Pyysalo, Thomas Wolf, and Colin~A. Raffel.
\newblock Scaling data-constrained language models.
\newblock In \emph{Advances in Neural Information Processing Systems
  (NeurIPS)}, 2023.

\bibitem[Penedo et~al.(2024)Penedo, Kydl\'{i}\v{c}ek, Ben~Allal, Lozhkov,
  Mitchell, Raffel, Von~Werra, and Wolf]{penedo2024fineweb}
Guilherme Penedo, Hynek Kydl\'{i}\v{c}ek, Loubna Ben~Allal, Anton Lozhkov,
  Margaret Mitchell, Colin Raffel, Leandro Von~Werra, and Thomas Wolf.
\newblock The {FineWeb} datasets: Decanting the web for the finest text data at
  scale.
\newblock In \emph{Advances in Neural Information Processing Systems (NeurIPS),
  Datasets and Benchmarks Track}, 2024.

\bibitem[Ramezani et~al.(2020)Ramezani, Cong, Mahdavi, Sivasubramaniam, and
  Kandemir]{ramezani2020}
Morteza Ramezani, Weilin Cong, Mehrdad Mahdavi, Anand Sivasubramaniam, and
  Mahmut Kandemir.
\newblock {GCN} meets {GPU}: Decoupling ``when to sample'' from ``how to
  sample''.
\newblock In \emph{Advances in Neural Information Processing Systems
  (NeurIPS)}, 2020.
\newblock LazyGCN; reuses a sampled minibatch over several steps and cites
  minibatch persistency by name.

\bibitem[Shallue et~al.(2019)Shallue, Lee, Antognini, Sohl-Dickstein, Frostig,
  and Dahl]{shallue2019}
Christopher~J. Shallue, Jaehoon Lee, Joseph Antognini, Jascha Sohl-Dickstein,
  Roy Frostig, and George~E. Dahl.
\newblock Measuring the effects of data parallelism on neural network training.
\newblock \emph{Journal of Machine Learning Research}, 20\penalty0
  (112):\penalty0 1--49, 2019.

\bibitem[You et~al.(2023)You, Chung, and Chowdhury]{you2023zeus}
Jie You, Jae-Won Chung, and Mosharaf Chowdhury.
\newblock Zeus: Understanding and optimizing {GPU} energy consumption of {DNN}
  training.
\newblock In \emph{USENIX Symposium on Networked Systems Design and
  Implementation (NSDI)}, 2023.

\bibitem[Zhao et~al.(2022)Zhao, Agarwal, Basant, et~al.]{zhao2022ingestion}
Mark Zhao, Niket Agarwal, Aarti Basant, et~al.
\newblock Understanding data storage and ingestion for large-scale deep
  recommendation model training.
\newblock In \emph{Proceedings of the 49th Annual International Symposium on
  Computer Architecture (ISCA)}, 2022.
\newblock \doi{10.1145/3470496.3533044}.

\end{thebibliography}

\appendix \section{Learning curves}
\label{app:curves}

\begin{figure}[p]
\centering
\includegraphics[width=\textwidth]{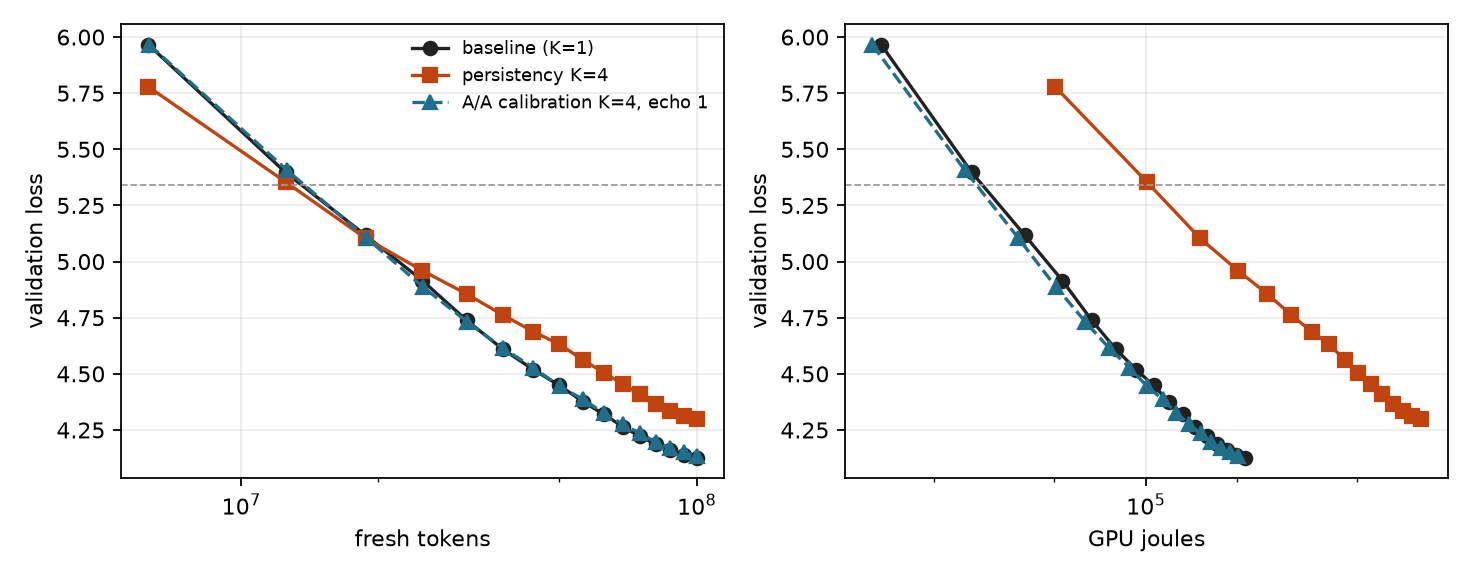}
\caption{Learning curves at $B = 32$: validation loss against fresh tokens
(left) and against GPU joules (right), one run per arm. The dashed line is
$\tau(32)$. The persistency arm (red) starts below the baseline (black) on
fresh tokens and is overtaken shortly after both have passed the target; on
joules it is displaced to the right by a factor that grows over the run. The A/A arm (blue,
dashed) tracks the baseline on both axes. One run per arm, unpaired: the run
with the lowest final validation loss in its cell over learning rates and
seeds, among the runs on the card that carries the cell.}
\label{fig:curves32}
\end{figure}

\begin{figure}[p]
\centering
\includegraphics[width=\textwidth]{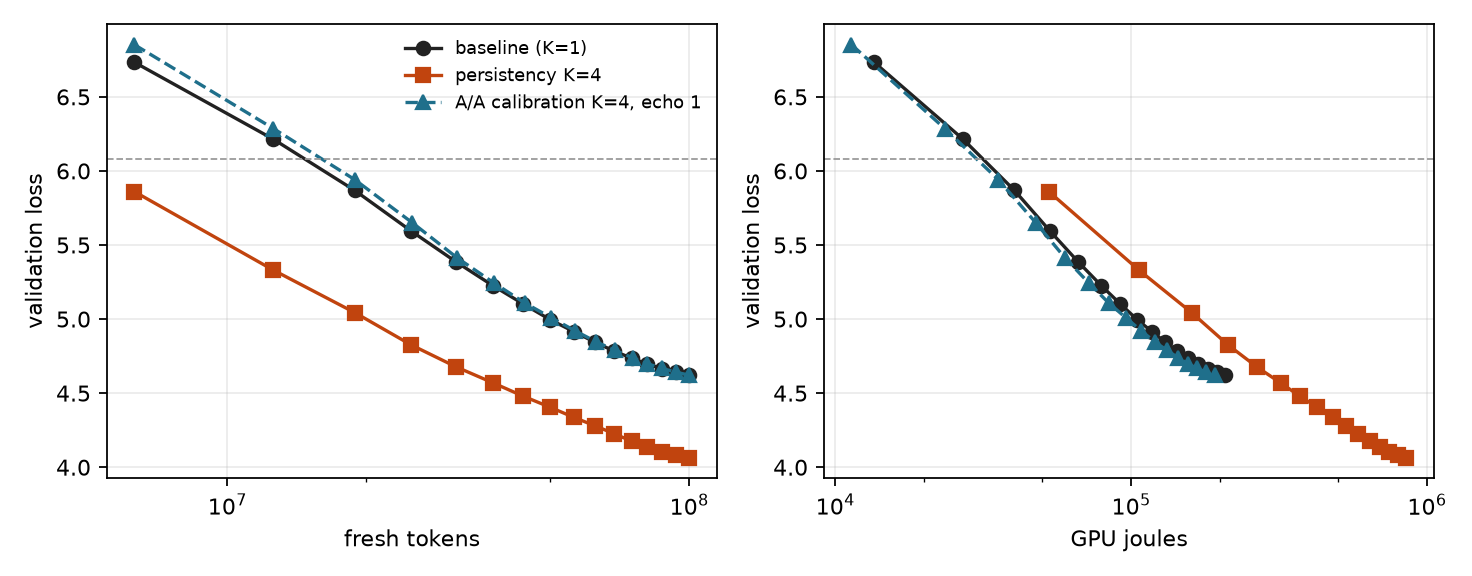}
\caption{Learning curves at $B = 128$, same layout as
Figure~\ref{fig:curves32}. The persistency arm is below the baseline on fresh
tokens at every evaluation, and below $\tau(128)$ already at its first one; on
joules it is displaced to the right of the baseline for the whole run. One run
per arm, unpaired, chosen by the lowest final validation loss over learning
rates and seeds among the runs on the card that carries the cell.}
\label{fig:curves128}
\end{figure}

\begin{figure}[p]
\centering
\includegraphics[width=\textwidth]{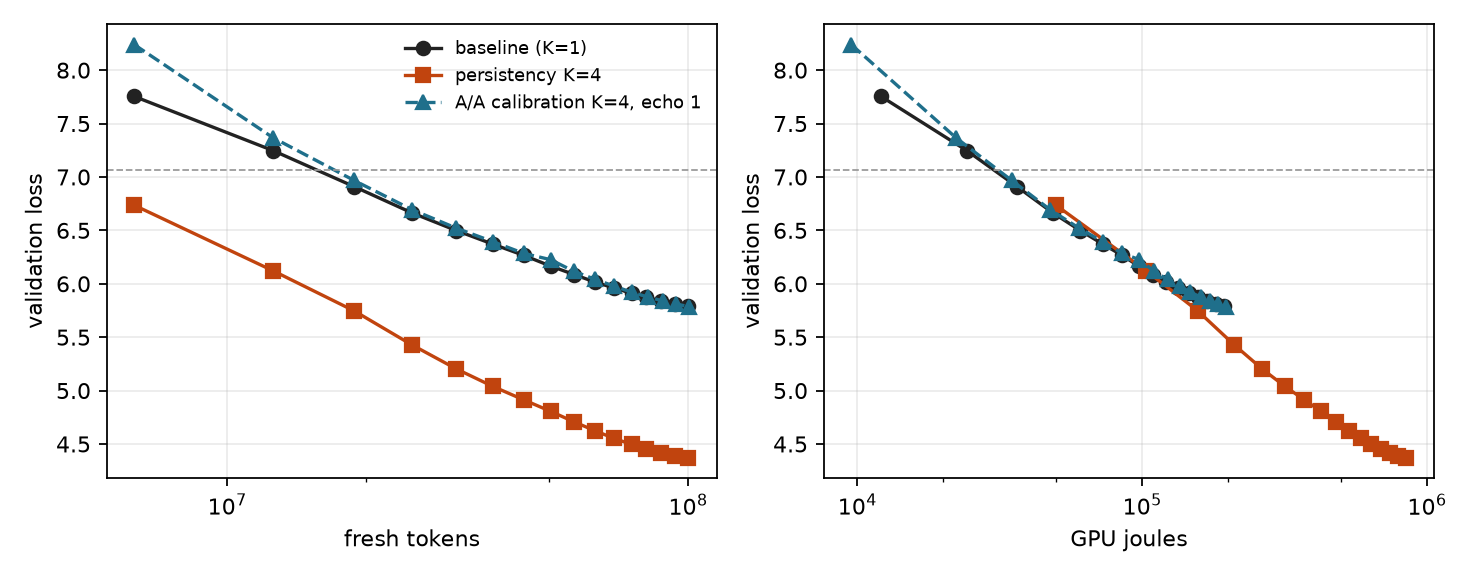}
\caption{Learning curves at $B = 512$, same layout as
Figure~\ref{fig:curves32}. On fresh tokens the persistency arm is below the
baseline at every evaluation and below $\tau(512)$ at its first one; on joules
the three curves lie on top of one another over the range where they overlap,
the persistency arm continuing further because the same fresh-token budget
lasts \Kpers{} times as many steps. One run per arm, unpaired, chosen by the
lowest final validation loss over learning rates and seeds among the runs on
the card that carries the cell; the $(512, 7)$ group of
Section~\ref{sec:results:bridge}, which ran on the Blackwell, is excluded.}
\label{fig:curves512}
\end{figure}

Figures~\ref{fig:curves32}--\ref{fig:curves512} show the learning curves
behind the ratios of Section~\ref{sec:results}, one figure per batch size. Each
figure has two panels: validation loss against fresh tokens consumed (left) and
against GPU joules (right), both on a logarithmic axis, with the frozen target
$\tau(B)$ of Table~\ref{tab:targets} as a dashed horizontal line. Three curves
are drawn per panel: the tuned baseline (black circles), the persistency arm
(red squares) and the A/A calibration arm (blue triangles, dashed). One run is
drawn per arm, namely, the run with the lowest final validation loss in its
cell over learning rates and seeds, among the runs on the card that carries the
cell, so that the joule panel compares the treatment and not the card (the
$(512, 7)$ group of Section~\ref{sec:results:bridge} ran on the Blackwell and
is not drawn). The arms drawn therefore need not share a seed, and the figures
are a picture of the curves, not of the paired estimator of
Section~\ref{sec:method:hypothesis}.

Three things are visible here that the ratios cannot show. The first is where
the arms cross. At $B = 32$ the persistency arm starts below the baseline on
fresh tokens and is overtaken between the third and the fourth evaluation, at
about \crosslossa{} nats and \crosstoka~Mtok; the target $\tau(32)$
sits just above the crossing, which is why the two arms reach it at nearly the
same fresh-token cost (the ratio \Rfresha{} of Table~\ref{tab:wins}) and
why the fresh-token verdict at $B = 32$ is neither a win nor a loss. At
$B = 128$ and $B = 512$ the persistency curve lies below the baseline on fresh
tokens at every evaluation and never crosses it; the drawn runs end at
\finalpersb{} against \finalbaseb{} nats and at \finalpersc{}
against \finalbasec{} nats, the same budget of fresh tokens having carried
the persistency arm through \Kpers{} times as many steps. The second is how the
frozen targets sit on the evaluation grid. At $B = 128$ and $B = 512$ the
persistency arm is already below $\tau(B)$ at its first evaluation
(\persfirstevalb{} and \persfirstevalc{} nats against targets of
\taub{} and \tauc{}); under the crossing rule its cost to target is then the
cost at that evaluation, with no interpolation, so at those two batch sizes the
frozen-target ratio is a ceiling on the treatment's cost, read at the
resolution of one evaluation interval. The ceiling works against the hypothesis
of Section~\ref{sec:method:hypothesis}, since it overstates $g(512)$. The
re-adjusted targets of Table~\ref{tab:targets} lie below that first
evaluation---this is what the re-adjustment exists for---and
Section~\ref{sec:results:primary} reports both. The third is the shape of the
joule curves. At $B = 32$ the persistency curve is displaced to the right of
the baseline by a factor near \Rjoulea{} at the target, and the
displacement widens over the rest of the run. At $B = 128$ it is near
\Rjouleb{} at the target and of the same order to the end. At $B = 512$
the persistency curve lies on top of the baseline over the whole range where the
two overlap: between the two drawn runs the ratio of joules at equal loss is
\jratiosixfivec{} at $\jlosshi$ nats and \jratiosixc{} at $\jlosslo$ nats, one
run per arm and unpaired; the paired reading over seeds is in
Appendix~\ref{app:rtau}. Below its first evaluation the persistency arm
therefore buys about the same loss per joule as the baseline, and the joule
ratio of \Rjoulec{} at $\tau(512)$ is a ceiling read inside the first
evaluation interval, which the curves do not resolve.

\section{The ratio as a function of the target}
\label{app:rtau}

\begin{figure}[p]
\centering
\includegraphics[width=\textwidth]{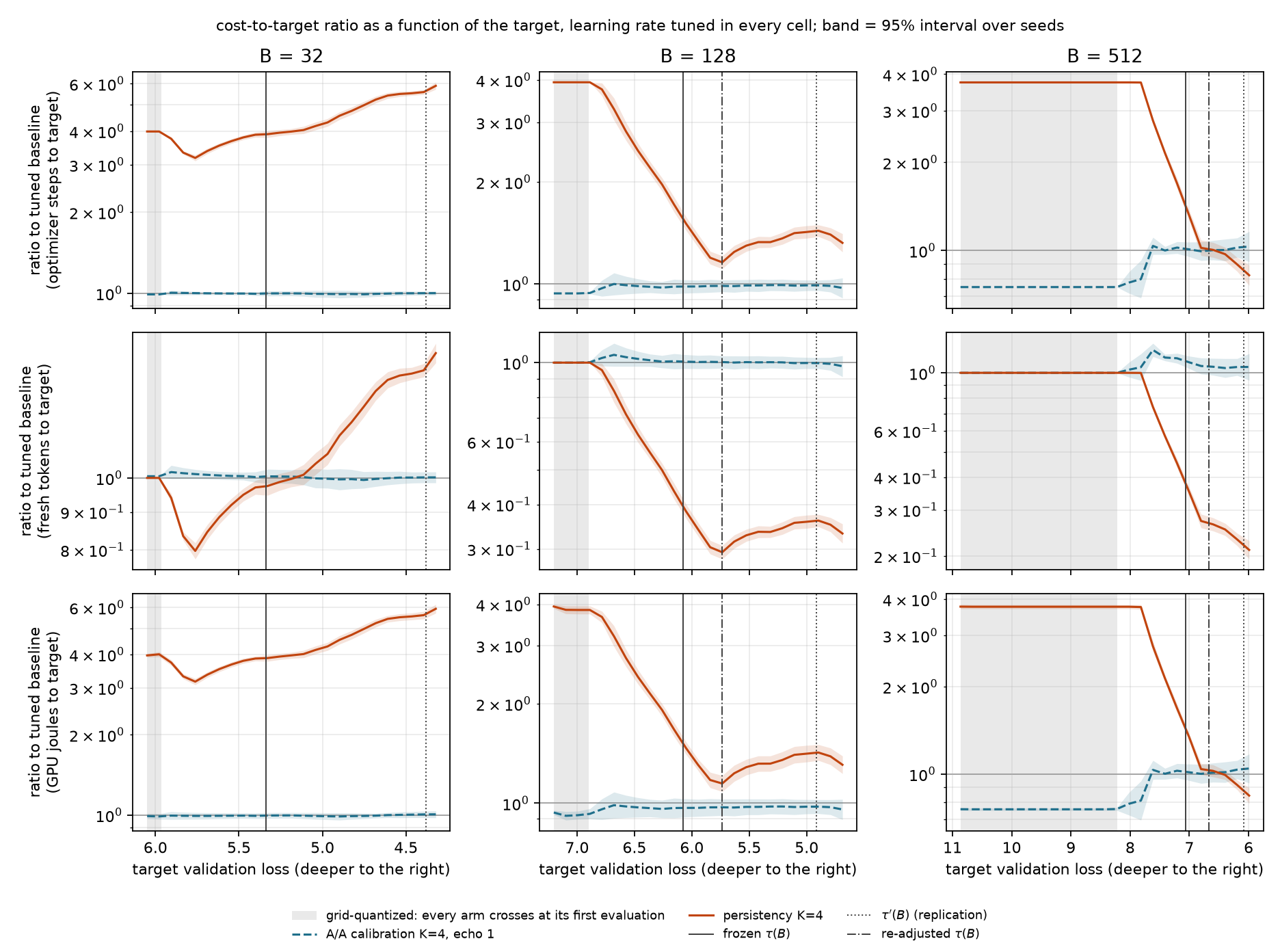}
\caption{Cost-to-target ratio of the persistency arm (red) and of the A/A arm
(blue, dashed) to the tuned baseline, as a function of the target validation
loss, deeper targets to the right; one column per batch size, one row per cost
axis; bands are 95\% $t$-intervals over seeds on the logarithm. Shaded: the
grid-quantized stretch, where every arm crosses at its first evaluation and
the ratio is fixed by the evaluation grid. Solid vertical line: the frozen
$\tau(B)$; dash-dotted: the re-adjusted $\tau(B)$ of Table~\ref{tab:targets};
dotted: the replication's $\tau'(B)$. Below one favors reuse. Drawn on the
\nrunsnow{} runs of the confirmatory stage.}
\label{fig:rtau}
\end{figure}

The frozen targets are one point on each learning curve, and
Section~\ref{sec:results:primary} explains why at $B = 128$ and $B = 512$
that point falls inside the persistency arm's first evaluation interval.
Figure~\ref{fig:rtau} removes the choice of point. It sweeps the target from
the highest first-evaluation loss in the cell down to the deepest loss that
every (arm, seed) pair reaches at some learning rate, below which pairs begin
to be censored. At each target it computes exactly the estimator of
Table~\ref{tab:wins}: the paired geometric mean over seeds of the
persistency-to-baseline cost at the cost-minimizing learning rate, with its
95\% interval, on steps, fresh tokens and joules. Not all of that range is
informative. Above the first evaluation of the slowest arm at the rate that
reaches the target first---\zfourquanta{}, \zfourquantb{} and \zfourquantc{}
nats at $B = 32$, $128$ and $512$---every arm crosses at its first evaluation
in every seed. There the ratio is a ratio of two grid positions (the
first-evaluation column of Table~\ref{tab:crossing}) and its band has zero
width. At $B = 512$ the A/A calibrator sits well below one there on steps and
joules, with a band that does not cover one: the plainest sign that the
stretch is not interpretable. That stretch is shaded in every panel. The
frozen $\tau(B)$ lies below it at every batch size and is drawn as a solid
vertical line. The re-adjusted targets of Table~\ref{tab:targets} are
dash-dotted. The targets $\tau'(B)$ of the pre-registered replication
(\tauEtwoa{}, \tauEtwob{} and \tauEtwoc{} nats, the running minimum of the
full-length baselines at \Etwofraction{} of the fresh budget) are dotted. The
A/A arm is drawn behind the treatment.

The figure is descriptive and post hoc: it was drawn after the frozen-target
results had been read, and no target on it other than $\tau(B)$ carries a
registered claim. What it shows is that the verdict at $B = 512$ on steps and
joules is a function of the target's depth. Read at $\tau(512)$ the ratios sit
above one; followed to deeper targets they fall, cross one, and at
$\tau'(512)$ lie below it. On joules the paired ratio is \zfourjpairedhi{}
[\zfourjpairedhilo{}, \zfourjpairedhihi{}] at \jlosshi{} nats and
\zfourjpairedlo{} [\zfourjpairedlolo{}, \zfourjpairedlohi{}] at \jlosslo{}
nats, on \nseeds{} pairs; the single-run, unpaired reading of
Appendix~\ref{app:curves} at the same two losses sits above it at both. At
$B = 128$ the ratio falls steeply
through $\tau(128)$ and then settles above one, where it stays to the deepest
target resolved. At $B = 32$ nothing of the kind happens: the step ratio never
comes near one at any depth---it dips briefly below $\Kpers$ and then grows
past it at the deepest targets---and the reuse buys nothing there at any
depth. This dependence on depth is the reason the replication
registers its targets on the full-length baselines and its evaluation grid in
optimizer steps, and it is the reason the paper quotes no headline penalty from
this figure. Drawn on the \nrunsnow{} runs of the confirmatory stage.

Table~\ref{tab:crossing} gives, per cell, the numbers behind the ceiling: how
many optimizer steps a run makes, how many evaluations it gets, at which step
the first of them falls, where on that grid the cost-minimizing run of each
seed crosses $\tau(B)$, and in how many seeds the crossing \emph{is} the first
evaluation.

\begin{table}[p]
\centering
{\footnotesize\setlength{\tabcolsep}{3pt}%
\begin{tabular}{llrrrrrc}
\toprule
$B$ & arm & \begin{tabular}[b]{@{}c@{}}steps\\per run\end{tabular} & \begin{tabular}[b]{@{}c@{}}evals\\per run\end{tabular} & \begin{tabular}[b]{@{}c@{}}first eval.\\(step)\end{tabular} & steps to $\tau(B)$ & \begin{tabular}[b]{@{}c@{}}seeds\\reached\end{tabular} & \begin{tabular}[b]{@{}c@{}}at first\\eval.\end{tabular} \\
\midrule
$32$ & baseline & 3052 & 16 & 191 & 408--435 & 8/8 & 0/8 \\
$32$ & persistency & 12205 & 16 & 761 & 1570--1689 & 8/8 & 0/8 \\
$32$ & A/A & 3049 & 16 & 189 & 414--433 & 8/8 & 0/8 \\
$128$ & baseline & 763 & 16 & 48 & 112--129 & 8/8 & 0/8 \\
$128$ & persistency & 3049 & 16 & 189 & 189 & 8/8 & 8/8 \\
$128$ & A/A & 761 & 16 & 45 & 112--127 & 8/8 & 0/8 \\
$512$ & baseline & 191 & 16 & 12 & 30--34 & 8/8 & 0/8 \\
$512$ & persistency & 761 & 16 & 45 & 45 & 8/8 & 8/8 \\
$512$ & A/A & 189 & 16 & 9 & 29--34 & 8/8 & 0/8 \\
\bottomrule
\end{tabular}}

\caption{Steps, evaluations and crossings per cell at the frozen targets, on
the \nrunsnow{} runs of the confirmatory stage. ``Steps per run'' and ``evals
per run'' are ranges over the runs of the cell (both learning rates, all
seeds); ``first eval.'' is the optimizer step of the first evaluation; ``steps
to $\tau(B)$'' is the range over seeds of the cost at the cost-minimizing
learning rate, the quantity every ratio in the paper is built from; ``seeds
reached'' counts the seeds that reached the target; ``at first eval.'' counts,
among those, the seeds whose crossing is the first evaluation, where the cost
is a ceiling read at the resolution of one evaluation interval. Generated by
\texttt{mbp.plots -{}-crossing-table}; the file records its provenance.}
\label{tab:crossing}
\end{table}

\section{The replication's grid, and the pass counts of the epoch contrast}
\label{app:passcount}

Table~\ref{tab:crossinge2} is Table~\ref{tab:crossing} recomputed on the
replication of Section~\ref{sec:results:e2}. Two things are worth reading off
it. The last column is zero in every cell: with an evaluation grid in optimizer
steps and targets set on full-length baselines, no seed reaches its target at
its first evaluation, and no cost in the table is a ceiling. And the two
readings of the crossing---over the registered horizon, and under the
unmodified two-evaluation rule of Section~\ref{sec:method}---agree on the range
of every cell, which is the table-level version of the statement that the
verdict does not depend on the confirmation rule.

Table~\ref{tab:rates} lists the two learning rates that the coarse stage
carried into the final stage in each cell, read on the run archive; every
final-stage, replication and E3 run of a cell uses one of the two. The coarse
stage ran on seed~\coarseseed{}; the primary and the bridge use seeds
0--\lastseed{}, the epoch contrast seeds \eoneseedrange{}, and the
replication seeds \etwoseedrange{}.

\begin{table}[t]
\centering
\small
\caption{The two learning rates per cell carried from the coarse stage
(Section~\ref{sec:setup:stages}) into the final stage, from the run archive.
The A/A arm's pair coincides with the persistency arm's at $B = 32$ and $128$
and with the baseline's at $B = 512$; which of the two wins is chosen per
(cell, seed), as the text states where the winner matters.}
\label{tab:rates}
\begin{tabular}{lccc}
\toprule
arm & $B = 32$ & $B = 128$ & $B = 512$ \\
\midrule
baseline $K = 1$ & \ratepairbaseLoa{}, \ratepairbaseHia{} & \ratepairbaseLob{}, \ratepairbaseHib{} & \ratepairbaseLoc{}, \ratepairbaseHic{} \\
persistency $K = 4$ & \ratepairpersLoa{}, \ratepairpersHia{} & \ratepairpersLob{}, \ratepairpersHib{} & \ratepairpersLoc{}, \ratepairpersHic{} \\
A/A & \ratepairaaLoa{}, \ratepairaaHia{} & \ratepairaaLob{}, \ratepairaaHib{} & \ratepairaaLoc{}, \ratepairaaHic{} \\
\bottomrule
\end{tabular}
\end{table}

The epoch contrast of Section~\ref{sec:results:e1} rests on a claim about data
rather than about statistics: that its two arms consume the same examples the
same number of times, so that only the spacing of the repetitions differs. The
claim is checked rather than assumed, and checked per seed, since each seed
draws its own stream. Each run records a digest of the distinct example indices
it drew, together with the number of those indices and the number of epochs it
made over them; the harness compares the digests of the two arms seed by seed,
which is stronger than comparing them as sets---two arms could hold the same
eight digests and hand them to different seeds. At every batch size the digests
agree in all \nseeds{} seeds, the epoch counts are one and four as registered,
the distinct-example counts are \eoneitems{}, and the two arms take the same
\eonesteps{} optimizer steps and push the same \eonefwdtok{} Mtok through the
forward pass. The check is part of \texttt{mbp.epochs} and runs with the
analysis, not beside it.

\begin{table}[p]
\centering
{\footnotesize\setlength{\tabcolsep}{3pt}%
\begin{tabular}{llrrrrrrc}
\toprule
$B$ & arm & \begin{tabular}[b]{@{}c@{}}steps\\per run\end{tabular} & \begin{tabular}[b]{@{}c@{}}evals\\per run\end{tabular} & \begin{tabular}[b]{@{}c@{}}first eval.\\(step)\end{tabular} & \begin{tabular}[b]{@{}c@{}}steps to $\tau'(B)$\\horizon rule\end{tabular} & \begin{tabular}[b]{@{}c@{}}steps to $\tau'(B)$\\2-eval.\ rule\end{tabular} & \begin{tabular}[b]{@{}c@{}}seeds\\reached\end{tabular} & \begin{tabular}[b]{@{}c@{}}at first\\eval.\end{tabular} \\
\midrule
$32$ & baseline & 3052 & 64 & 47 & 1722--1782 & 1722--1782 & 8/8 & 0/8 \\
$32$ & persistency & 12205 & 259 & 47 & 9600--10113 & 9600--10113 & 8/8 & 0/8 \\
$32$ & A/A & 3049 & 64 & 47 & 1703--1825 & 1703--1825 & 8/8 & 0/8 \\
$128$ & baseline & 763 & 69 & 11 & 430--475 & 430--475 & 8/8 & 0/8 \\
$128$ & persistency & 3049 & 277 & 11 & 608--673 & 608--673 & 8/8 & 0/8 \\
$128$ & A/A & 761 & 69 & 11 & 431--480 & 431--480 & 8/8 & 0/8 \\
$512$ & baseline & 191 & 95 & 2 & 104--138 & 104--138 & 8/8 & 0/8 \\
$512$ & persistency & 761 & 380 & 2 & 101--109 & 101--109 & 8/8 & 0/8 \\
$512$ & A/A & 189 & 94 & 2 & 104--126 & 104--126 & 8/8 & 0/8 \\
\bottomrule
\end{tabular}}

\caption{Steps, evaluations and crossings per cell in the replication of
Section~\ref{sec:results:e2}, at its targets $\tau'(B)$, on the replication's
own \etwonruns{} runs (seeds \zfourEtwoseedfirst{}--\zfourEtwoseedlast{}).
Columns as in Table~\ref{tab:crossing}, with the crossing read twice: over
the registered horizon of \etwohorizon{} steps (``horizon rule'') and under
the unmodified two-evaluation rule of Section~\ref{sec:method} (``2-eval.\
rule''). Note the last column, zero throughout, against every seed at
$B = 128$ and $B = 512$ in Table~\ref{tab:crossing}. Generated by
\texttt{mbp.plots -{}-crossing-table -{}-confirm-horizon}; the file records its
provenance.}
\label{tab:crossinge2}
\end{table}

\end{document}